\documentclass{article}

\usepackage{PRIMEarxiv}

\usepackage[utf8]{inputenc}
\usepackage[T1]{fontenc}
\usepackage{amsfonts}
\usepackage{microtype}
\usepackage{fancyhdr}
\usepackage{amsmath}
\usepackage{amssymb}
\usepackage{graphicx}
\usepackage{hyperref}

\newtheorem{theorem}{Theorem}
\newtheorem{lemma}{Lemma}
\newtheorem{proposition}{Proposition}

\title{BEACONS: Bounded-Error, Algebraically-Composable Neural Solvers for Partial Differential Equations}

\author{Jonathan Gorard\\
Princeton University\\
Princeton, NJ 08544, USA\\
\texttt{gorard@princeton.edu}\\
\And
Ammar Hakim, James Juno\\
Princeton Plasma Physics Laboratory\\
Princeton, NJ 08540, USA\\
\texttt{\{ahakim, jjuno\}@pppl.gov}}

\begin{document}

\maketitle

\begin{abstract}
The traditional limitations of neural network architectures in reliably generalizing substantively beyond the convex hulls of their training data present a significant problem for computational physics, in which one often wishes to solve partial differential equations (PDEs) in regimes far beyond anything which can be experimentally or analytically validated. In this paper, we demonstrate how it is possible to circumvent these limitations by constructing \textit{formally-verified} neural network architectures for the solution of PDEs, with rigorous convergence, stability, and conservation properties, whose correctness can therefore be guaranteed even in extrapolatory regimes. By using the \textit{method of characteristics} to predict the analytical properties of PDE solutions a priori (even in regions arbitrarily far from the training domain), we show how it is possible to construct rigorous extrapolatory bounds on the worst-case ${L^{\infty}}$ errors of shallow neural network approximations. Then, by decomposing PDE solutions into compositions of simpler functions, we show how it is possible to compose these shallow neural network approximations together to form deep architectures, building on ideas from \textit{compositional deep learning}, in which the large ${L^{\infty}}$ errors in the approximations of discontinuous functions have been suppressed by composing them with smoother ones (thus generalizing the theory of \textit{flux limiters} in traditional numerical solvers). The resulting framework, called \textit{BEACONS} (Bounded-Error, Algebraically-COmposable Neural Solvers), comprises both an automatic code-generator for the neural solvers themselves, as well as a bespoke automated theorem-proving system for producing machine-checkable \textit{certificates of correctness}. We apply the BEACONS framework to a variety of linear and non-linear PDEs, including the linear advection and inviscid Burgers' equations, as well as the full compressible Euler equations, in both 1D and 2D, and illustrate how BEACONS architectures are able to extrapolate solutions far beyond the training data in a reliable and bounded way, as compared to conventional neural network architectures of equivalent size. Various advantages of the BEACONS approach over the classical \text{PINN} (Physics-Informed Neural Network) approach are discussed.
\end{abstract}

\section{Introduction}

Conventional machine learning wisdom suggests that neural networks (NNs) are highly effective tools for \textit{interpolating} between training data within a given domain or statistical distribution, but frequently struggle at \textit{extrapolating} to Out-of-Domain or Out-of-Distribution (OOD) cases\cite{devore_nonlinear_1998}\cite{bishop_pattern_2016}. This is often formulated as the statement that neural networks perform well on cases that lie near or within the \textit{convex hull} of their training sets, but fail when tested on cases that lie far outside it. For example, most analyses of \textit{physics-informed neural network} (PINN) architectures have focused on evaluating their ability to interpolate partial differential equation solutions within the temporal ranges on which they have been trained\cite{raissi_physics-informed_2019}, and have shown that the predictions of PINN architectures tend to deviate significantly from the ground truth solution when made to extrapolate beyond such ranges\cite{kim_dpm:_2021}\cite{wang_long-time_2023}. Modern \textit{foundation model} (FM) architectures effectively circumvent this conventional extrapolatory limitation by pretraining on a sufficiently large and diverse corpus of data that the convex hull of the pretraining set becomes arbitrarily expansive and high-dimensional\cite{bommasani2022opportunitiesrisksfoundationmodels}, in particular such that a very wide range of seemingly unrelated tasks lie strictly within (or otherwise near) it, and therefore the model is able to exhibit an impressive capacity for generalization even by ``mere interpolation'' of the pretraining set. In some trivial sense, accurate extrapolation far outside the convex hull of the training set is a fundamental mathematical impossibility: for any given subdomain, there exist infinitely many functions whose values all agree on that subdomain (but which may differ arbitrarily outside of it), and there is no guarantee that a neural network which has been trained only on that subdomain will fit the ``correct'' (or even ``approximately correct'') one. In fact, this is deeply related to a variety of instabilities that occur in classical numerical interpolation, such as the Gibbs and Runge phenomena in polynomial and Fourier interpolation, and more broadly to the idea that an interpolation which is accurate over a fixed interval may diverge arbitrarily rapidly outside of it. We will return to this theme later in the paper. It has been proposed that deep neural network architectures exhibit an implicit bias towards fitting functions with low algorithmic/Kolmogorov complexity (i.e. the \textit{Occam's razor hypothesis}\cite{mingard_deep_2025}), which may help to restrict the search space of possible function extrapolations somewhat. However, even if the Occam's razor hypothesis were a strict mathematical theorem or provable guarantee, there still exist many possible and relevant criteria for ``correctness'' of a function extrapolation beyond minimization of algorithmic complexity.

In traditional computational physics, the situation is rather different. Large, complex simulation codes are frequently pushed into regimes well beyond anything that can be validated against either experiments or analytical theory, and in many cases the results are nevertheless deemed trustworthy. In certain cases, this degree of trustworthiness can be even explicitly quantified. This is in large part due to a range of foundational results in classical numerical analysis, which make it possible to construct finite volume, finite element, finite difference, discontinuous Galerkin, etc. schemes with \textit{rigorous} mathematical and physical correctness properties, such as enforcement of conservation laws to machine precision\cite{wesseling_principles_2001}, or guaranteed ${L^2}$ stability\cite{courant_partial_1967}. Indeed, in previous work\cite{gorard2025shockconfidenceformalproofs}, the authors demonstrated that it was possible to construct a bespoke automated theorem-proving framework which, when coupled with an automatic code-generator, was capable of synthesizing fully \textit{formally-verified} numerical solvers for hyperbolic partial differential equations, with machine-checkable certificates of correctness (covering properties such as thermodynamic consistency, strict hyperbolicity preservation, and flux continuity) up to machine precision. One of the principal contentions of the present paper is that neural networks, properly conceived, simply represent another class of fundamental numerical method, and can therefore be rigorously analyzed using essentially the same mathematical techniques. More accurately, neural networks represent a \textit{vast generalization} of all classical numerical methods: all numerical methods ultimately work by approximating an infinite-dimensional function space by a finite-dimensional subspace, yet only in neural networks are the bases for these finite-dimensional subspaces treated as fully dynamic, flexible, and adaptive to the specific problem. By this token, if we regard neural networks as a generalization of classical numerical techniques, we intend to address the following question: could we imagine designing \textit{formally-verified} neural network architectures with rigorous robustness, convergence, stability, and correctness properties, just as we have demonstrated previously in the classical case?

To this end, this paper introduces the framework of \textit{BEACONS}: Bounded-Error, Algebraically-COmposable Neural Solvers for partial differential equations (PDEs). The ``bounded-error'' aspect has its origins in some of the classical theorems of Mhaskar\cite{mhaskar_neural_1996}\cite{mhaskar_neural_1995}, Pinkus\cite{pinkus_approximation_1999}, and others on the approximation of ${C^n}$-smooth functions by simple, feed-forward neural networks (especially multi-layer perceptrons with a single hidden layer) with smooth activation functions, in which rigorous bounds on the ${L^{\infty}}$ errors of such approximations can be proven, dependent only upon the dimensionality of the function, the number of continuous derivatives, and the width of the neural network. Here, we focus specifically on the case of neural network approximations for solutions to \textit{hyperbolic} PDEs. Unlike in the case of arbitrary function interpolation, in this case the number of continuous derivatives of the function can be known \textit{a priori}, at every point in the domain (even at points arbitrarily far removed from the subdomain on which the neural network has been trained), by virtue of the \textit{method of characteristics}\cite{roe_characteristic-based_1986}. Crucially, when combined with an appropriate choice of PDE residual-based loss function, this technique enables one to prove fully \textit{extrapolatory} error bounds on the neural network approximation, rather than merely the conventional \textit{interpolatory} ones. For shallow neural network architectures (with only a single hidden layer), such bounds may nevertheless be infeasibly large for any reasonable number of neurons, which in turn motivates the ``algebraically-composable'' aspect of the BEACONS framework. Building upon intuitions from applied category theory\cite{fong_invitation_2019}\cite{gorard2024appliedcategorytheorywolfram} and compositional deep learning\cite{gavranović2019compositionaldeeplearning}\cite{gavranović2024positioncategoricaldeeplearning}, the idea behind algebraic composability is to assemble deeper, more expressive neural network architectures by composing together shallower, more tractable ones\cite{gorard2022functorialperspectivemulticomputationalirreducibility}, in such a way that the error bounds still remain tightly controlled. In particular, we demonstrate how it is possible to improve an otherwise unfavorable ${L^{\infty}}$ bound on the neural network approximation of a highly discontinuous function (e.g. a shock wave, in the context of hyperbolic PDEs) by composing it with several favorable neural network approximations of smooth, slowly-varying functions. Such an approach effectively generalizes the use of \textit{flux limiters}\cite{leveque_finite_2011} and \textit{total variation diminishing} schemes\cite{harten_high_1997} in conventional finite volume numerical methods.

At this point, it is worth stressing several crucial distinctions between the BEACONS approach and the traditional PINN approach\footnote{For further details on these critiques of PINN architectures, we refer the reader to the systematic investigations of McGreivy and Hakim\cite{mcgreivy_weak_2024}.}. In a traditional PINN architecture, the latent space of the neural network is heavily constrained (for instance through specialized choices of loss functions, penalty functions, and so on) to prevent the neural network from entering an ``invalid'' region of the latent space that violates certain physical correctness properties, such as conservation laws, thermodynamic principles, etc. This approach works well for simple toy problems (especially ones involving linear or almost-linear differential operators), where the latent space structure is convex and/or geometrically simple. However, for more sophisticated problems involving complex, tightly coupled, highly non-linear systems of PDEs, the latent space structure becomes highly non-convex and geometrically complicated, and one quickly reaches the situation where the shortest paths between two ``valid'' regions of latent space (and therefore the paths preferentially traversed by a gradient descent algorithm) may be forced to intersect an otherwise ``invalid'' region. For example, the shortest path between two regions of latent space corresponding to thermodynamically-consistent solutions (with convex entropy functionals) may traverse through a ``ravine'' consisting of solutions in which the entropy functional is non-convex, and therefore in which the laws of thermodynamics are violated. In such a scenario, a traditional PINN architecture will simply fail to converge, or will converge to a physically incorrect but ``valid'' result on the wrong side of the ``ravine''. On the other hand, in the BEACONS approach, the latent space is entirely unconstrained.: the provable error bounds on the approximations of PDE solutions arise from the compositional structure of the architecture itself, rather than through any a priori restriction of the training process. During the gradient descent process for a BEACONS architecture, the neural network is fully permitted to traverse physically ``invalid'' regions of the latent space (e.g. corresponding to entropy-violating solutions, non-energy-conserving solutions, etc.) as necessary, without penalty, since there is nevertheless a provable guarantee by virtue of its mathematical structure that the architecture will \textit{eventually} converge to within some bounded distance of a ``valid'' region, in the limit of infinite epochs (and infinite training data). Formal certifications for a posteriori error bounds on PINN inferences have previously been studied by Ernst et al.\cite{ernst2025posterioricertificationneuralnetwork}, Eiras et al.\cite{eiras2024efficienterrorcertificationphysicsinformed}, and others.

The BEACONS software framework itself comprises: a Racket-based\cite{felleisen_et_al} domain-specific language (DSL) for specifying both hyperbolic PDE systems and compositional neural network architectures for approximating them; an automated theorem-proving framework that is able to prove both rigorous \textit{analytical} bounds on the ${L^{\infty}}$ errors of individual layers in the neural network, as well as robust \textit{algebraic} bounds on how those errors interact when the individual layers are composed into a deeper architecture; and an automatic code-generator that is able to synthesize optimized C code for generating the training data, constructing the neural network architecture, performing the training, validating the results, and inferring new solutions. This framework is designed to integrate seamlessly into the existing formal verification pipeline that the authors previously developed for hyperbolic PDE solvers, enabling the training data itself to be generated by fully formally-verified numerical solvers, thereby facilitating the construction of fully \textit{end-to-end} formally-verified neural network solvers with \textit{machine-checkable} certificates of correctness, across every stage of the training, validation, and inference process.

We begin in Section \ref{sec:Section1} by introducing the overarching mathematical theory of the BEACONS framework. In Section \ref{sec:Section2}, we start by formulating both classical numerical methods (e.g. finite element and finite volume methods) and neural network-based methods (e.g. PINN-style methods using PDE residual-based loss functions) for solving PDEs within a common mathematical framework, namely the construction of finite-dimensional approximations to infinite-dimensional function spaces. In Section \ref{sec:Section3}, we proceed to introduce the classical results of Mhaskar\cite{mhaskar_neural_1996}, as summarized by Mhaskar and Poggio\cite{mhaskar_deep_2016}, on the approximation of ${C^n}$-smooth functions by multi-layer perceptrons with a single hidden layer, and combine these with the \textit{method of characteristics} for hyperbolic PDEs\cite{roe_characteristic-based_1986} (which can be used to predict the smoothness of PDE solutions a priori, purely as a function of the smoothness of their initial data, as well as the analytical properties of their \textit{flux functions} and/or the eigenstructure of their \textit{flux Jacobians}). Using this combination, we prove one of the two central theorems underlying the BEACONS framework: a rigorous \textit{extrapolatory} bound on the worst-case ${L^{\infty}}$ error of a shallow neural network approximation to a hyperbolic PDE solution. Finally, in Section \ref{sec:Section4}, we prove the second central theorem of the BEACONS framework, illustrating that the worst-case ${L^{\infty}}$ error for a shallow neural network approximation to a discontinuous PDE solution (e.g. a shock wave) may be improved by decomposing the solution into a composition of a smooth and slowly-varying function (with a small Lipschitz constant $L$) and a discontinuous one. Each function is approximated by its own shallow neural network, such that the arbitrarily large ${L^{\infty}}$ error in the neural network approximation of the discontinuous function is suppressed in the composition by the arbitrarily small ${L^{\infty}}$ error in the neural network approximation of the smooth one. This theorem furnishes us with an \textit{algebraic composability} rule for shallow neural network architectures, and therefore a straightforward algorithm for assembling deep BEACONS architectures for approximating complex PDE solutions with favorable ${L^{\infty}}$ bounds by composing together shallower neural network architectures for approximating simpler, smoother functions.

We move on in Section \ref{sec:Section5} to the details of the software implementation of the BEACONS framework itself. We begin by summarizing the design of the Racket-based DSL and its constituent data structures for representing hyperbolic PDE systems, numerical simulations, and their associated neural solvers. Then, we move on to introducing the design of the automatic code-generator for synthesizing optimized C code for generating numerical training data, setting up and training the BEACONS architectures themselves, and validating the resulting neural solvers. Finally, we conclude by discussing the design of the automated theorem-proving algorithm, including explanations for how fully executable (i.e. machine-checkable) proofs of worst-case ${L^{\infty}}$ error bounds on the BEACONS architectures are generated, by means of a Racket-based symbolic rewriting system that implements fully \textit{correctness-preserving} algebraic transformations of symbolic expressions, always respecting the IEEE-754 standard for floating-point arithmetic. Section \ref{sec:Section6} presents various numerical results obtained by actually running the BEACONS framework end-to-end, for three illustrative equations (or equation systems), in both one and two dimensions. Section \ref{sec:Section7} shows results for the 1D and 2D (scalar) linear advection equation, while Section \ref{sec:Section8} shows results for the 1D and 2D (scalar) inviscid Burgers' equation, with all training data generated by fully formally-verified numerical solvers. Section \ref{sec:Section9} shows results for the 1D and 2D (coupled, vector) compressible Euler equations, with training data now generated by an \textit{unverified} but otherwise high-resolution numerical solver (and therefore with all BEACONS error bounds proved \textit{conditionally}, subject to the assumption that the solver is mathematically correct). In each case, we compare BEACONS architectures with a variety of sizes against ordinary feed-forward neural network architectures of equivalent width and depth, and perform ${L^2}$ and ${L^{\infty}}$ error analysis, as well as a rigorous conservation analysis, on the results. These error and conservation analyses are performed both pointwise, i.e. for the final simulation frame only, as well as integrated across time. In all cases, we find that the BEACONS architectures exhibit significantly lower ${L^2}$ and ${L^{\infty}}$ errors, significantly lower conservation errors, more favorable convergence and stability properties, and greater capability for capturing, preserving, and extrapolating qualitative features (especially in higher dimensions), than their ordinary neural network counterparts of equivalent size. We also find, as expected, that the true ${L^{\infty}}$ errors exhibited by the BEACONS architectures fall well within the worst-case ${L^{\infty}}$ bounds proven by the formal verification pipeline. We end in Section \ref{sec:Section10} with some concluding remarks and possible directions for future research.

The entire BEACONS framework is open source, and has been seamlessly integrated into the existing formal verification pipeline for hyperbolic PDE solvers described in \cite{gorard2025shockconfidenceformalproofs}, which is itself part of the \texttt{gkylcas} project\footnote{\url{https://github.com/ammarhakim/gkylcas}}. Although the resulting numerical solvers can be executed as purely standalone pieces of C code, they can also be integrated and run as part of the larger \textsc{Gkeyll} computational multi-physics simulation framework\footnote{\url{https://github.com/ammarhakim/gkeyll}}, which then provides the requisite (non-formally-verified) infrastructure for running and training on larger-scale simulations, such as parallelism, grid generation, data input/output, etc.

\section{Mathematical Theory}
\label{sec:Section1}

\subsection{Neural Network Approximations for Partial Differential Equations}
\label{sec:Section2}

Let ${\Omega \subset \mathbb{R}^d}$ denote a $d$-dimensional \textit{computational domain}, and let:

\begin{equation}
u : \Omega \to \mathbb{R},
\end{equation}
represent an unknown scalar field defined over this domain. This scalar field is assumed to represent a solution to the scalar partial differential equation (PDE):

\begin{equation}
\forall \mathbf{x} \in \Omega, \qquad \mathcal{L} \left[ u \right] \left( \mathbf{x} \right) = s \left( \mathbf{x} \right),
\end{equation}
obeying the boundary condition:

\begin{equation}
\forall \mathbf{x} \in \partial \Omega, \qquad \mathcal{B} \left[ u \right] \left( \mathbf{x} \right) = g \left( \mathbf{x} \right),
\end{equation}
where ${\mathcal{L}}$ is an arbitrary (potentially non-linear) differential operator, ${\mathcal{B}}$ is a boundary operator, ${s : \Omega \to \mathbb{R}}$ is an arbitrary source term (or forcing function), and ${g : \partial \Omega \to \mathbb{R}}$ is a boundary function. For instance, for a first-order hyperbolic PDE in conservation law form, evolving in a single spatial dimension, one has ${d = 2}$ (since our computational domain here spans both space and time, so ${\mathbf{x} = \left( t, x \right) \in \Omega \subset \mathbb{R}^2}$), with differential operator ${\mathcal{L}}$ given by:

\begin{equation}
\mathcal{L} \left[ u \right] = \frac{\partial u}{\partial t} + \frac{\partial f \left( u \right)}{\partial x},
\end{equation}
where ${f : \mathbb{R} \to \mathbb{R}}$ is a scalar \textit{flux function} and ${u : \Omega \to \mathbb{R}}$ now represents a conserved quantity. Since the space and time coordinates are treated here on the same footing, no distinction is made between \textit{initial conditions} and (spatial) boundary conditions: both are treated as special cases of \textit{space-time} boundary conditions. For example, to represent Dirichlet initial or boundary conditions, one simply has ${\mathcal{B} \left[ u \right] = u}$, while to represent Neumann initial or boundary conditions, one has:

\begin{equation}
\mathcal{B} \left[ u \right] = \frac{\partial u}{\partial t}, \qquad \text{ or } \qquad \mathcal{B} \left[ u \right] = \frac{\partial u}{\partial x},
\end{equation}
respectively.

In a classical numerical method, one is typically trying to approximate an infinite-dimensional function space defined over the computational domain ${\Omega}$ (such as the Sobolev space ${H^1 \left( \Omega \right)}$) by a finite-dimensional one:

\begin{equation}
\mathcal{V}_h = \mathrm{span} \left\lbrace \varphi_1, \varphi_2, \dots, \varphi_N \right\rbrace,
\end{equation}
where each ${\varphi_i : \Omega \to \mathbb{R}}$ is a scalar-valued \textit{basis function}, such that our PDE solution ${u : \Omega \to \mathbb{R}}$ is approximated by the linear combination:

\begin{equation}
u_h \left( \mathbf{x} \right) = \sum_{i = 1}^{N} U_i \varphi_i \left( \mathbf{x} \right),
\end{equation}
where ${\left( u_h : \Omega \to \mathbb{R} \right) \in \mathcal{V}_h}$, $N$ is the total number of \textit{degrees of freedom}, and the coefficients ${U_i \in \mathbb{R}}$ are arbitrary. For instance, in a finite element or discontinuous Galerkin method\cite{cockburn_local_1998}\cite{cockburn_rungekutta_1998}, the ${\varphi_i : \Omega \to \mathbb{R}}$ represent nodal or modal basis functions, while in a finite volume method\cite{toro_riemann_2009}, the ${\varphi_i : \Omega \to \mathbb{R}}$ represent piecewise functions of compact support (where the region of support in each case is localized to a single cell), etc. One may regard neural networks as being a grand generalization of this same idea, wherein the basis functions themselves are permitted to vary. Consider the case of a multi-layer perceptron (MLP) consisting of a single hidden layer, itself comprising $N$ hidden neurons. Each hidden neuron effectively defines a ``basis function'' ${\psi_i}$:

\begin{equation}
\forall i \in \left\lbrace 1, \dots, N \right\rbrace, \qquad \psi_i \left( \mathbf{x} ; W^{\left( 1 \right)}, \mathbf{b}^{\left( 1 \right)} \right) = \sigma \left( \sum_{j = 1}^{d} W_{i j}^{\left( 1 \right)} x_j + b_{i}^{\left( 1 \right)} \right),
\end{equation}
where ${\mathbf{x} \in \Omega}$ is an input vector, ${W^{\left( 1 \right)} \in \mathbb{R}^{N \times d}}$ is a matrix of hidden layer weights, ${\mathbf{b}^{\left( 1 \right)} \in \mathbb{R}^N}$ is a vector of hidden layer biases, and ${\sigma : \mathbb{R} \to \mathbb{R}}$ is a non-linear activation function. Then, the output layer assembles an approximate solution ${u_{\boldsymbol\theta} : \Omega \to \mathbb{R}}$ as a linear combination of these ``basis functions'' (plus an optional scalar bias):

\begin{equation}
u_{\boldsymbol\theta} \left( \mathbf{x} \right) = b^{\left( 2 \right)} + \sum_{i = 1}^{N} W_{1 i}^{\left( 2 \right)} \psi_i \left( \mathbf{x}; W^{\left( 1 \right)}, \mathbf{b}^{\left( 1 \right)} \right),
\end{equation}
where ${\mathbf{W}^{\left( 2 \right)} \in \mathbb{R}^{1 \times N}}$ is a vector of output layer weights, ${b^{\left( 2 \right)} \in \mathbb{R}}$ is an output layer bias, and ${\boldsymbol\theta \in \mathbb{R}^{1 + \left( d + 2 \right) N}}$ is a vector representing all trainable parameters, i.e. the concatenation of all entries of ${W^{\left( 1 \right)}}$, ${\mathbf{b}^{\left( 1 \right)}}$, ${\mathbf{W}^{\left( 2 \right)}}$, and ${b^{\left( 2 \right)}}$. If we hold the hidden layer weights ${W^{\left( 1 \right)}}$ and biases ${\mathbf{b}^{\left( 1 \right)}}$ fixed, and set the output layer bias ${b^{\left( 2 \right)} = 0}$, then we recover exactly the same form of approximation as in a classical numerical method (with the output layer weights ${\mathbf{W}^{\left( 2 \right)}}$ playing the role of the coefficients ${U_i \in \mathbb{R}}$ in the basis expansion). For this reason, the finite-dimensional function space spanned by possible neural network approximations:

\begin{equation}
\mathcal{V}_{\text{NN}} = \left\lbrace u_{\boldsymbol\theta} \mid \boldsymbol\theta \in \mathbb{R}^P \right\rbrace,
\end{equation}
where $P$ is the total number of trainable parameters (i.e. ${P = 1 + \left( d + 2 \right) N}$ for a multi-layer perceptron with a single hidden layer), is strictly higher-dimensional than the function space ${\mathcal{V}_h}$ spanned by classical numerical approximations, with the former being $N$-dimensional and the latter being (at least) ${\left( 1 + \left( d + 2 \right) N \right)}$-dimensional.

In order to construct this neural network approximation ${\left( u_{\boldsymbol\theta} : \Omega \to \mathbb{R} \right) \in \mathcal{V}_{\text{NN}}}$ to the PDE solution ${\left( u : \Omega \to \mathbb{R} \right) \in H^1 \left( \Omega \right)}$, we begin by sampling ${N_{\text{int}}}$ points ${\mathbf{x}_{\text{int}}^{\left( i \right)}}$ from the interior of the computational domain ${\Omega \setminus \partial \Omega}$, along with ${N_{\text{bound}}}$ points ${\mathbf{x}_{\text{bound}}^{\left( i \right)}}$ from its boundary ${\partial \Omega}$:

\begin{equation}
\left\lbrace \mathbf{x}_{\text{int}}^{\left( i \right)} \right\rbrace_{i = 1}^{N_{\text{int}}} \subset \Omega \setminus \partial \Omega, \qquad \text{ and } \qquad \left\lbrace \mathbf{x}_{\text{bound}}^{\left( i \right)} \right\rbrace_{i = 1}^{N_{\text{bound}}} \subset \partial \Omega.
\end{equation}
Next, we compute the scalar PDE residual ${r_{\text{PDE}} \left( \mathbf{x}_{\text{int}}^{\left( i \right)} ; \boldsymbol\theta \right)}$ for each sampled point on the interior:

\begin{equation}
\forall i \in \left\lbrace 1, \dots, N_{\text{int}} \right\rbrace, \qquad r_{\text{PDE}} \left( \mathbf{x}_{\text{int}}^{\left( i \right)} ; \boldsymbol\theta \right) = \mathcal{L} \left[ u_{\boldsymbol\theta} \right] \left( \mathbf{x}_{\text{int}}^{\left( i \right)} \right) - s \left( \mathbf{x}_{\text{int}}^{\left( i \right)} \right),
\end{equation}
along with the scalar boundary residual ${r_{\text{BC}} \left( \mathbf{x}_{\text{bound}}^{\left( i \right)} ; \boldsymbol\theta \right)}$ for each sampled point on the boundary:

\begin{equation}
\forall i \in \left\lbrace 1, \dots, N_{\text{bound}} \right\rbrace, \qquad r_{\text{BC}} \left( \mathbf{x}_{\text{bound}}^{\left( i \right)} ; \boldsymbol\theta \right) = \mathcal{B} \left[ u_{\boldsymbol\theta} \right] \left( \mathbf{x}_{\text{bound}}^{\left( i \right)} \right) - g \left( \mathbf{x}_{\text{bound}}^{\left( i \right)} \right),
\end{equation}
from which we can calculate the PDE and boundary scalar loss functions ${J_{\text{PDE}} \left( \boldsymbol\theta \right)}$ and ${J_{\text{BC}} \left( \boldsymbol\theta \right)}$, by means of a least squares approach, namely:

\begin{equation}
J_{\text{PDE}} \left( \boldsymbol\theta \right) = \frac{1}{N_{\text{int}}} \sum_{i = 1}^{N_{\text{int}}} \left( r_{\text{PDE}} \left( \mathbf{x}_{\text{int}}^{\left( i \right)} ; \boldsymbol\theta \right) \right)^2, \qquad \text{ and } \qquad J_{\text{BC}} \left( \boldsymbol\theta \right) = \frac{1}{N_{\text{bound}}} \sum_{i = 1}^{N_{\text{bound}}} \left( r_{\text{BC}} \left( \mathbf{x}_{\text{bound}}^{\left( i \right)} ; \boldsymbol\theta \right) \right)^2,
\end{equation}
respectively\cite{brevis_machine-learning_2021}. The overall scalar loss function ${J \left( \boldsymbol\theta \right)}$ is then assembled as a linear combination of the two:

\begin{equation}
J \left( \boldsymbol\theta \right) = \lambda_{\text{PDE}} J_{\text{PDE}} \left( \boldsymbol\theta \right) + \lambda_{\text{BC}} J_{\text{BC}} \left( \boldsymbol\theta \right),
\end{equation}
where the coefficients ${\lambda_{\text{PDE}}, \lambda_{\text{BC}} \in \mathbb{R}}$ are arbitrary (with ${\lambda_{\text{PDE}}}$ effectively characterizing how much to penalize violations of the PDE itself, and ${\lambda_{\text{BC}}}$ effectively characterizing how much to penalize violations of the initial and boundary conditions). The purpose of the neural network training process is now to compute a set of parameters ${\boldsymbol\theta^{*}}$ that (at least approximately) minimizes this combined loss function:

\begin{equation}
\boldsymbol\theta^{*} = \mathrm{arg} \min_{\boldsymbol\theta \in \mathbb{R}^P} \left\lbrace J \left( \boldsymbol\theta \right) \right\rbrace,
\end{equation}
typically by means of a gradient descent algorithm, for example by setting the initial parameter vector ${\boldsymbol\theta_0 = \mathbf{0}}$ and then iterating:

\begin{equation}
\boldsymbol\theta_{n + 1} = \boldsymbol\theta_n - \eta_n \nabla_{\boldsymbol\theta} J \left( \boldsymbol\theta_n \right),
\end{equation}
until the desired convergence is achieved (i.e. until ${\left\lVert \boldsymbol\theta_{n + 1} - \boldsymbol\theta_n \right\rVert_{\infty} < \varepsilon}$ for some ${\varepsilon > 0}$), where ${\eta_n \in \mathbb{R}}$ represents a scalar learning rate (analogous to a time-step ${\Delta t}$ in a classical explicit numerical method). By linearity, we can write out the $P$ components of the gradient vector:

\begin{equation}
\nabla_{\boldsymbol\theta} J \left( \boldsymbol\theta \right) = \begin{bmatrix}
\frac{\partial J \left( \boldsymbol\theta \right)}{\partial \theta_1} & \frac{\partial J \left( \boldsymbol\theta \right)}{\partial \theta_2} & \cdots & \frac{\partial J \left( \boldsymbol\theta \right)}{\partial \theta_P}
\end{bmatrix}^{\intercal} \in \mathbb{R}^P,
\end{equation}
explicitly as:

\begin{equation}
\forall p \in \left\lbrace 1, \dots, P \right\rbrace, \qquad \frac{\partial J \left( \boldsymbol\theta \right)}{\partial \theta_p} = \lambda_{\text{PDE}} \left( \frac{\partial J_{\text{PDE}} \left( \boldsymbol\theta \right)}{\partial \theta_p} \right) + \lambda_{\text{BC}} \left( \frac{\partial J_{\text{BC}} \left( \boldsymbol\theta \right)}{\partial \theta_p} \right),
\end{equation}
which we can then expand via the chain rule in terms of the PDE and boundary residuals ${r_{\text{PDE}} \left( \mathbf{x}_{\text{int}} ; \boldsymbol\theta \right)}$ and ${r_{\text{BC}} \left( \mathbf{x}_{\text{bound}} ; \boldsymbol\theta \right)}$, along with their respective derivatives, as:

\begin{multline}
\frac{\partial J \left( \boldsymbol\theta \right)}{\partial \theta_p} = \lambda_{\text{PDE}} \left( \frac{2}{N_{\text{int}}} \sum_{i = 1}^{N_{\text{int}}} r_{\text{PDE}} \left( \mathbf{x}_{\text{int}}^{\left( i \right)} ; \boldsymbol\theta \right) \left( \frac{\partial r_{\text{PDE}} \left( \mathbf{x}_{\text{int}}^{\left( i \right)} ; \boldsymbol\theta \right)}{\partial \theta_p} \right) \right)\\
+ \lambda_{\text{BC}} \left( \frac{2}{N_{\text{bound}}} \sum_{i = 1}^{N_{\text{bound}}} r_{\text{BC}} \left( \mathbf{x}_{\text{bound}}^{\left( i \right)} ; \boldsymbol\theta \right) \left( \frac{\partial r_{\text{BC}} \left( \mathbf{x}_{\text{bound}}^{\left( i \right)} ; \boldsymbol\theta \right)}{\partial \theta_p} \right) \right).
\end{multline}

Generalizing the above analysis to the case of coupled \textit{systems} of PDEs is relatively straightforward. Specifically, suppose that our unknown quantity is now a vector field:

\begin{equation}
\mathbf{U} : \Omega \to \mathbb{R}^m,
\end{equation}
where ${m \geq 2}$ represents the total number of unknown variables, and that this vector field is assumed to represent a solution to the coupled PDE system:

\begin{equation}
\forall \mathbf{x} \in \Omega, \qquad \boldsymbol{\mathcal{L}} \left[ \mathbf{U} \right] \left( \mathbf{x} \right) = \mathbf{S} \left( \mathbf{x} \right),
\end{equation}
obeying the boundary condition:

\begin{equation}
\forall \mathbf{x} \in \partial \Omega, \qquad \boldsymbol{\mathcal{B}} \left[ \mathbf{U} \right] \left( \mathbf{x} \right) = \mathbf{G} \left( \mathbf{x} \right),
\end{equation}
where ${\boldsymbol{\mathcal{L}}}$ is now an arbitrary and potentially non-linear \textit{vector} of differential operators (such that ${\boldsymbol{\mathcal{L}} \left[ \mathbf{U} \right] \left( \mathbf{x} \right) \in \mathbb{R}^m}$), ${\boldsymbol{\mathcal{B}}}$ is now a \textit{vector} of boundary operators (such that ${\boldsymbol{\mathcal{B}} \left[ \mathbf{U} \right] \left( \mathbf{x} \right) \in \mathbb{R}^m}$), ${\mathbf{S} : \Omega \to \mathbb{R}^m}$ is now an arbitrary source (or forcing) vector field, and ${\mathbf{G} : \partial \Omega \to \mathbb{R}^m}$ is now a boundary vector field. For instance, for a first-order system of hyperbolic PDEs in a single spatial dimension, in conservation law form, the vector of differential operators ${\boldsymbol{\mathcal{L}}}$ is now given by:

\begin{equation}
\boldsymbol{\mathcal{L}} \left[ \mathbf{U} \right] = \frac{\partial \mathbf{U}}{\partial t} + \frac{\partial \mathbf{F} \left( \mathbf{U} \right)}{\partial x},
\end{equation}
with flux vector field ${\mathbf{F} : \mathbb{R}^m \to \mathbb{R}^m}$, and with ${\mathbf{U} : \Omega \to \mathbb{R}^m}$ now representing a vector field of conserved quantities. Our neural network approximation ${\mathbf{U}_{\boldsymbol\theta} : \Omega \to \mathbb{R}^m}$ to the conserved vector field ${\mathbf{U} : \Omega \to \mathbb{R}^m}$ can now be assembled componentwise by taking linear combinations of the same ``basis functions'' ${\psi_i : \Omega \to \mathbb{R}}$ as in the scalar case:

\begin{equation}
\forall i \in \left\lbrace 1, \dots, m \right\rbrace, \qquad U_{\boldsymbol\theta, i} \left( \mathbf{x} \right) = b_{i}^{\left( 2 \right)} + \sum_{j = 1}^{N} W_{i j}^{\left( 2 \right)} \psi_j \left( \mathbf{x} ; W^{\left( 1 \right)} , \mathbf{b}^{\left( 1 \right)} \right),
\end{equation}
where now ${W^{\left( 2 \right)} \in \mathbb{R}^{m \times N}}$ has been promoted to a matrix of output layer weights, and ${\mathbf{b}^{\left( 2 \right)} \in \mathbb{R}^m}$ has been promoted to a vector of output layer biases. The structure of the hidden layer remains unchanged from before, while the total number of trainable parameters is now ${P = m + \left( m + d + 1 \right) N}$. Now the PDE residual ${\mathbf{r}_{\text{PDE}} \left( \mathbf{x}_{\text{int}}^{\left( i \right)} ; \boldsymbol\theta \right)}$ is a vector for each sampled point on the interior of the domain:

\begin{equation}
\forall i \in \left\lbrace 1, \dots, N_{\text{int}} \right\rbrace, \qquad \mathbf{r}_{\text{PDE}} \left( \mathbf{x}_{\text{int}}^{\left( i \right)}; \boldsymbol\theta \right) = \boldsymbol{\mathcal{L}} \left[ \mathbf{U}_{\boldsymbol\theta} \right] \left( \mathbf{x}_{\text{int}}^{\left( i \right)} \right) - \mathbf{S} \left( x_{\text{int}}^{\left( i \right)} \right),
\end{equation}
and likewise for the boundary residual ${r_{\text{BC}} \left( \mathbf{x}_{\text{bound}}^{\left( i \right)} ; \boldsymbol\theta \right)}$ for each sampled point on the boundary of the domain:

\begin{equation}
\forall i \in \left\lbrace 1, \dots, N_{\text{bound}} \right\rbrace, \qquad \mathbf{r}_{\text{BC}} \left( \mathbf{x}_{\text{bound}}^{\left( i \right)} ; \boldsymbol\theta \right) = \boldsymbol{\mathcal{B}} \left[ \mathbf{U}_{\theta} \right] \left( \mathbf{x}_{\text{bound}}^{\left( i \right)} \right) - \mathbf{G} \left( \mathbf{x}_{\text{bound}}^{\left( i \right)} \right),
\end{equation}
To convert these vector-valued PDE and boundary residuals into forms that are compatible with a scalar loss function, as needed for gradient descent, we compute the weighted Euclidean norms, with weight matrix ${W \in \mathbb{R}^{m \times m}}$ (assumed to be symmetric and positive semi-definite), yielding:

\begin{equation}
\left\lVert \mathbf{r}_{\text{PDE}} \left( \mathbf{x}_{\text{int}}^{\left( i \right)} ; \boldsymbol\theta \right) \right\rVert_{W} = \sqrt{\sum_{\alpha = 1}^{m} \sum_{\beta = 1}^{m} r_{\text{PDE}, \alpha} \left( \mathbf{x}_{\text{int}}^{\left( i \right)} ; \boldsymbol\theta \right) \, W_{\alpha \beta} \, r_{\text{PDE}, \beta} \left( \mathbf{x}_{\text{int}}^{\left( i \right)} ; \boldsymbol\theta \right)},
\end{equation}
and:

\begin{equation}
\left\lVert \mathbf{r}_{\text{BC}} \left( \mathbf{x}_{\text{bound}}^{\left( i \right)} ; \boldsymbol\theta \right) \right\rVert_{W} = \sqrt{\sum_{\alpha = 1}^{m} \sum_{\beta = 1}^{m} r_{\text{BC}, \alpha} \left( \mathbf{x}_{\text{bound}}^{\left( i \right)} ; \boldsymbol\theta \right) \, W_{\alpha \beta} \, r_{\text{BC}, \beta} \left( \mathbf{x}_{\text{bound}}^{\left( i \right)} ; \boldsymbol\theta \right)},
\end{equation}
respectively, with the diagonal components of $W$ effectively characterizing how much to penalize conservation errors in each of the conserved quantities separately, and the off-diagonal components of $W$ effectively characterizing how much to penalize conservation errors in particular \textit{combinations} of the conserved quantities. Therefore, the PDE and boundary scalar loss functions ${J_{\text{PDE}} \left( \boldsymbol\theta \right)}$ and ${J_{\text{BC}} \left( \boldsymbol\theta \right)}$ now become:

\begin{equation}
J_{\text{PDE}} \left( \boldsymbol\theta \right) = \frac{1}{N_{\text{int}}} \sum_{i = 1}^{N_{\text{int}}} \sum_{\alpha = 1}^{m} \sum_{\beta = 1}^{m} r_{\text{PDE}, \alpha} \left( \mathbf{x}_{\text{int}}^{\left( i \right)} ; \boldsymbol\theta \right) \, W_{\alpha \beta} \, r_{\text{PDE}, \beta} \left( \mathbf{x}_{\text{int}}^{\left( i \right)} ; \boldsymbol\theta \right),
\end{equation}
and:

\begin{equation}
J_{\text{BC}} \left( \boldsymbol\theta \right) = \frac{1}{N_{\text{bound}}} \sum_{i = 1}^{N_{\text{bound}}} \sum_{\alpha = 1}^{m} \sum_{\beta = 1}^{m} r_{\text{BC}, \alpha} \left( \mathbf{x}_{\text{bound}}^{\left( i \right)} ; \boldsymbol\theta \right) \, W_{\alpha \beta} \, r_{\text{BC}, \beta} \left( \mathbf{x}_{\text{bound}}^{\left( i \right)} ; \boldsymbol\theta \right),
\end{equation}
respectively, and so each of the $P$ components of the gradient vector ${\nabla_{\boldsymbol\theta} J \left( \boldsymbol\theta \right)}$ become:

\begin{multline}
\frac{\partial J \left( \boldsymbol\theta \right)}{\partial \theta_p} = \lambda_{\text{PDE}} \left( \frac{2}{N_{\text{int}}} \sum_{i = 1}^{N_{\text{int}}} \sum_{\alpha = 1}^{m} \sum_{\beta = 1}^{m} r_{\text{PDE}, \alpha} \left( \mathbf{x}_{\text{int}}^{\left( i \right)} ; \boldsymbol\theta \right) \, W_{\alpha \beta} \left( \frac{\partial r_{\text{PDE}, \beta} \left( \mathbf{x}_{\text{int}}^{\left( i \right)} ; \boldsymbol\theta \right)}{\partial \theta_p} \right) \right)\\
+ \lambda_{\text{BC}} \left( \frac{2}{N_{\text{bound}}} \sum_{i = 1}^{N_{\text{bound}}} \sum_{\alpha = 1}^{m} \sum_{\beta = 1}^{m} r_{\text{BC}, \alpha} \left( \mathbf{x}_{\text{bound}}^{\left( i \right)} ; \boldsymbol\theta \right) \, W_{\alpha \beta} \left( \frac{\partial r_{\text{BC}, \beta} \left( \mathbf{x}_{\text{bound}}^{\left( i \right)} ; \boldsymbol\theta \right)}{\partial \theta_p} \right) \right).
\end{multline}
All other aspects of the gradient descent algorithm remain unchanged from the scalar case.

\subsection{Bounding Errors on Neural Network Approximations for Hyperbolic PDEs}
\label{sec:Section3}

Here, we follow the notational conventions of Mhaskar and Poggio\cite{mhaskar_deep_2016}, and consider our computational domain ${\Omega}$ to be the compact $d$-dimensional region ${\Omega = I^d = \left[ -1, 1 \right]^d}$, whose space ${\mathbb{X} = C \left( \Omega \right) = C \left( I^d \right)}$ of continuous functions ${f : I^d \to \mathbb{R}}$ is equipped with the standard ${L^{\infty}}$-norm:

\begin{equation}
\forall f \in \mathbb{X}, \qquad \left\lVert f \right\rVert_{\infty} = \max_{\mathbf{x} \in I^d} \left\lbrace \left\lvert f \left( \mathbf{x} \right) \right\rvert \right\rbrace.
\end{equation}
If we fix the activation function ${\sigma : \mathbb{R} \to \mathbb{R}}$ to be smooth (i.e. ${C^{\infty}}$, infinitely differentiable), then the set ${\mathcal{V}_{\text{NN}}}$ of possible neural network approximations ${u_{\boldsymbol\theta} : \mathbb{R}^d \to \mathbb{R}}$ forms, with appropriate normalization of the coefficients, a subspace of ${\mathbb{X}}$. Let ${W_{r, d}}$ denote the Sobolev space of functions ${f : \mathbb{R}^d \to \mathbb{R}}$ for which all partial derivatives up to order ${r \geq 1}$ exist and are continuous, in which all functions have been normalized such that:

\begin{equation}
\forall f \in W_{r, d}, \qquad \left\lVert f \right\rVert_{\infty} + \sum_{1 \leq \left\lVert \mathbf{k} \right\rVert_1 \leq r} \left\lVert \mathcal{D}^{\mathbf{k}} f \right\rVert_{\infty} \leq 1,
\end{equation}
where ${\left\lVert \mathbf{k} \right\rVert_1 = \sum\limits_{i = 1}^{d} k_i}$ is the standard ${L^1}$-norm on ${\mathbb{R}^d}$, and ${k_i \in \mathbb{Z}}$, such that the sum is taken over the set of all partitions of integers between 1 and $r$, with the operator ${\mathcal{D}^{\mathbf{k}}}$ denoting the composition of ${k_1}$ partial derivatives with respect to ${x_1}$, ${k_2}$ partial derivatives with respect to ${x_2}$, etc. Then, we have (by Theorem 2.1 of Mhaskar and Poggio\cite{mhaskar_deep_2016}):

\begin{theorem}
Suppose that:

\begin{equation}
\mathcal{V}_{\text{NN}} = \left\lbrace u_{\boldsymbol\theta} \mid \boldsymbol\theta \in \mathbb{R}^P \right\rbrace,
\end{equation}
represents the space of possible functions ${u_{\boldsymbol\theta} : \Omega \to \mathbb{R}}$ that can be computed by a multi-layer perceptron with $P$ trainable parameters, comprising a single hidden layer consisting of $N$ hidden neurons. Suppose, moreover, that the activation function ${\sigma : \mathbb{R} \to \mathbb{R}}$ for such a perceptron is ${C^{\infty}}$-smooth, and not given by a polynomial on any subinterval of ${\mathbb{R}}$. Then, one has:

\begin{equation}
\forall f \in W_{r, d}, \qquad \inf_{u_{\boldsymbol\theta} \in \mathcal{V}_{\text{NN}}} \left\lbrace \left\lVert f - u_{\boldsymbol\theta} \right\rVert_{\infty} \right\rbrace = \mathcal{O} \left( N^{-\frac{r}{d}} \right).
\end{equation}
\end{theorem}
Thus, with appropriate normalization, a $d$-dimensional ${C^n}$-smooth function ${u : \Omega \to \mathbb{R}}$ can be approximated by a neural network ${u_{\boldsymbol\theta} : \mathbb{R}^d \to \mathbb{R}}$ with a single hidden layer (containing $N$ hidden neurons), with a worst-case ${L^{\infty}}$ error on the order of ${N^{-\frac{n}{d}}}$. Henceforth, we will therefore move from using $r$ to using $n$ to denote the number of continuous derivatives of a function $f$ (as in ``$f$ is a ${C^n}$-smooth function'').

Remarkably, in the case where ${u : \Omega \to \mathbb{R}}$ corresponds to a solution to a hyperbolic PDE (especially a first-order hyperbolic PDE in conservation law form), the \textit{method of characteristics} enables one to predict a priori how many continuous partial derivatives of ${u : \Omega \to \mathbb{R}}$ must exist, at any point within the space-time domain of the PDE, based purely on the smoothness of the initial data and the analytical properties of the flux function ${f : \mathbb{R} \to \mathbb{R}}$\cite{courant_methods_2009}. Since these predictions on the smoothness of ${u : \Omega \to \mathbb{R}}$ can still be made in regimes that may be arbitrarily far from the neural network's training data, this enables one to prove \textit{extrapolatory} error bounds on the correctness of the neural network solution ${u_{\boldsymbol\theta} : \mathbb{R}^d \to \mathbb{R}}$, as we outline below. Suppose that we have a first-order hyperbolic PDE in conservation law form, evolving in one spatial dimension, with no source terms:

\begin{equation}
\frac{\partial u}{\partial t} + \frac{\partial f \left( u \right)}{\partial x} = 0,
\end{equation}
obeying the initial condition (at ${t = 0}$):

\begin{equation}
u \left( 0, x \right) = u_0 \left( x \right),
\end{equation}
where ${u_0 : \mathbb{R} \to \mathbb{R}}$ is (for now) a function of indeterminate smoothness. Assuming that the flux function ${f : \mathbb{R} \to \mathbb{R}}$ is at least ${C^1}$-smooth, this is equivalent (via the chain rule) to:

\begin{equation}
\frac{\partial u}{\partial t} + f^{\prime} \left( u \right) \left( \frac{\partial u}{\partial x} \right) = 0,
\end{equation}
which, by expanding out the ordinary derivatives in terms of partial derivatives, we see is equivalent to the ordinary differential equation:

\begin{equation}
\frac{d u}{d t} = 0, \qquad \text{ along curves satisfying } \qquad \frac{d x}{d t} = f^{\prime} \left( u \right).
\end{equation}
Since $u$ is constant in time (i.e. $u$ is a conserved quantity), ${f^{\prime} \left( u \right)}$ is also constant, and therefore these ``curves'' are really straight lines in the ${x - t}$ plane (though in the presence of non-vanishing source terms, ${\frac{d u}{d t} \neq 0}$ and so this simplification no longer applies). By integrating the equation ${\frac{d x}{d t} = f^{\prime} \left( u \right)}$ in time, we obtain:

\begin{align}
x - x_0 &= f^{\prime} \left( u \right) \left( t - t_0 \right)\\
&= f^{\prime} \left( u_0 \left( x_0 \right) \right) t,
\end{align}
since ${t_0 = 0}$, and ${u \left( t, x_0 \right) = u_0 \left( x_0 \right)}$ for all ${t \geq 0}$ since $u$ is constant in time along any such curve. Thus, ${x = x_0 + f^{\prime} \left( u_0 \left( x_0 \right) \right) t}$ is the equation for a straight \textit{characteristic line} emanating from point ${x_0}$, and we have:

\begin{equation}
\frac{\partial x}{\partial x_0} = 1 + f^{\prime\prime} \left( u_0 \left( x_0 \right) \right) u_{0}^{\prime} \left( x_0 \right) t,
\end{equation}
by the chain rule, assuming now that the flux function ${f : \mathbb{R} \to \mathbb{R}}$ is at least ${C^2}$-smooth, and that the initial data ${u_0 : \mathbb{R} \to \mathbb{R}}$ is at least ${C^1}$-smooth.

When two or more of these characteristic lines intersect, the function ${u : \Omega \to \mathbb{R}}$ is forced to be many-valued (and hence the solution becomes discontinuous)\cite{debnath_nonlinear_2012}, which is equivalent to the derivative ${\frac{\partial u}{\partial x}}$ becoming singular. Again using the fact that ${u \left( t, x \right) = u_0 \left( x_0 \right)}$ for all ${t \geq 0}$ along characteristic lines, we can evaluate this derivative via the chain rule as:

\begin{align}
\frac{\partial u}{\partial x} &= \frac{\partial u_0 \left( x_0 \right)}{\partial x} = u_{0}^{\prime} \left( x_0 \right) \left( \frac{\partial x_0}{\partial x} \right) = u_{0}^{\prime} \left( x_0 \right) \left( \frac{\partial x}{\partial x_0} \right)^{-1}\\
&= \frac{u_{0}^{\prime} \left( x_0 \right)}{1 + f^{\prime\prime} \left( u_0 \left( x_0 \right) \right) u_{0}^{\prime} \left( x_0 \right) t},
\end{align}
which becomes singular whenever:

\begin{equation}
t = \frac{-1}{f^{\prime\prime} \left( u_0 \left( x_0 \right) \right) u_{0}^{\prime} \left( x_0 \right)}.
\end{equation}
Therefore, the earliest time ${t_{\infty}}$ at which blowup occurs (and hence the solution ${u : \Omega \to \mathbb{R}}$ fails to be even ${C^1}$-smooth) is given by:

\begin{equation}
t_{\infty} = \left( \sup_{x_0 \in \mathbb{R}} \left\lbrace - f^{\prime\prime} \left( u_0 \left( x_0 \right) \right) u_{0}^{\prime} \left( x_0 \right) \right\rbrace \right)^{-1},
\end{equation}
subject to the convention that ${\frac{1}{0} = + \infty}$. Once a finite-time blowup has occurred, the physically relevant solution (i.e. the \textit{entropy solution}\cite{osher_riemann_1984}) is, at best, only piecewise ${C^n}$. In summary, we have:

\begin{lemma}
\label{lemma1}
Suppose that ${u : \Omega \to \mathbb{R}}$ is a solution to a (homogeneous) first-order hyperbolic PDE in conservation law form:

\begin{equation}
\frac{\partial u}{\partial t} + \frac{\partial f \left( u \right)}{\partial x} = 0,
\end{equation}
where the initial data:

\begin{equation}
u \left( 0, x \right) = u_0 \left( x \right),
\end{equation}
is given by a ${C^n}$-smooth function ${u_0 : \mathbb{R} \to \mathbb{R}}$, and the flux is given by a ${C^{n + 1}}$-smooth function ${f : \mathbb{R} \to \mathbb{R}}$, where ${n \geq 1}$. Then:

\begin{enumerate}
\item
If the flux is linear (i.e. ${f^{\prime\prime} = 0}$), or the flux is convex and the initial data is monotonically increasing (i.e. ${f^{\prime\prime} \geq 0}$ and ${u_{0}^{\prime} \geq 0}$), or the flux is concave and the initial data is monotonically decreasing (i.e. ${f^{\prime\prime} \leq 0}$ and ${u_{0}^{\prime} \leq 0}$), then the solution ${u : \Omega \to \mathbb{R}}$ remains ${C^n}$-smooth for all ${t \geq 0}$.

\item
Otherwise, the solution ${u : \Omega \to \mathbb{R}}$ remains ${C^n}$-smooth for all ${0 \leq t < t_{\infty}}$, where:

\begin{equation}
t_{\infty} = \left( \sup_{x_0 \in \mathbb{R}} \left\lbrace - f^{\prime\prime} \left( u_0 \left( x_0 \right) \right) u_{0}^{\prime} \left( x_0 \right) \right\rbrace \right)^{-1},
\end{equation}
and is discontinuous (though still piecewise ${C^n}$) for all ${t \geq t_{\infty}}$.
\end{enumerate}
\end{lemma}

Note that, although we have presented these results in terms of scalar PDEs, it is relatively straightforward to extend all of the above analysis to the case of coupled (vector) \textit{systems} of hyperbolic PDEs too. In particular, if ${\mathbf{U} : \Omega \to \mathbb{R}^m}$ is now a solution to a (homogeneous) system of first-order hyperbolic PDEs in conservation law form:

\begin{equation}
\frac{\partial \mathbf{U}}{\partial t} + \frac{\partial \mathbf{F} \left( \mathbf{U} \right)}{\partial x} = \mathbf{0},
\end{equation}
where the initial data:

\begin{equation}
\mathbf{U} \left( 0, x \right) = \mathbf{U}_0 \left( x \right),
\end{equation}
is given by a ${C^n}$-smooth vector field ${\mathbf{U}_0 : \mathbb{R} \to \mathbb{R}^m}$, and the flux is given by a ${C^{n + 1}}$-smooth vector field ${\mathbf{F} : \mathbb{R}^m \to \mathbb{R}^m}$, where ${n \geq 1}$, then we can apply a higher-dimensional generalization of the method of characteristics in much the same way. Assuming that system is \textit{strictly hyperbolic}, i.e. that the \textit{flux Jacobian} ${J_{\mathbf{F}}}$, defined by:

\begin{equation}
J_{\mathbf{F}} = \nabla_{\mathbf{U}} \mathbf{F} \left( \mathbf{U} \right),
\end{equation}
is diagonalizable, with all of its eigenvalues ${\lambda_i}$ being both real and distinct, we can construct the higher-dimensional analog of the characteristic equation, namely the following (fully decoupled) system of \textit{Riccati-type} equations:

\begin{equation}
\forall i \in \left\lbrace 1, \dots, m \right\rbrace, \qquad \frac{d \omega_i \left( \mathbf{U} \right)}{d t} + \alpha_i \left( \mathbf{U} \right) \omega_{i}^{2} \left( \mathbf{U} \right) + \sum_{j \neq i} \beta_{i j} \left( \mathbf{U} \right) \, \omega_i \left( \mathbf{U} \right) \, \omega_j \left( \mathbf{U} \right) = 0,
\end{equation}
with each ${\omega_i}$ representing a different \textit{wave} or \textit{mode} of the system. In the above, we have introduced:

\begin{equation}
\omega_i \left( \mathbf{U} \right) = \mathbf{l}_i \left( \mathbf{U} \right) \cdot \left( \frac{\partial \mathbf{U}}{\partial x} \right), \qquad \text{ and } \qquad \alpha_i \left( \mathbf{U} \right) = \left( \nabla_{\mathbf{U}} \lambda_i \left( \mathbf{U} \right) \right) \cdot \mathbf{r}_i \left( \mathbf{U} \right),
\end{equation}
where ${\mathbf{l}_i \left( \mathbf{U} \right)}$ and ${\mathbf{r}_i \left( \mathbf{U} \right)}$ denote the left and right eigenvectors of the flux Jacobian ${J_{\mathbf{F}}}$, respectively (the elements of the matrix ${\beta_{i j} \left( \mathbf{U} \right)}$ are smooth coefficients built from the eigenstructure of ${J_{\mathbf{F}}}$, the details of which are not relevant here). The salient point is that, when ${\alpha_i = 0}$, this is the analog of the linear flux (i.e. ${f^{\prime\prime} = 0}$) case for scalar PDEs: namely, the corresponding wave ${\omega_i}$ does not steepen into a shock, and the corresponding part of the solution remains ${C^n}$-smooth. When ${\alpha_i \neq 0}$, then the corresponding wave ${\omega_i}$ \textit{does} indeed steepen to induce a finite-time blowup of ${\left\lVert \frac{\partial \mathbf{U}}{\partial x} \right\rVert_{\infty}}$ at:

\begin{equation}
t_{\infty, i} = \left( \inf_{x_0 \in \mathbb{R}} \left\lbrace - \alpha_i \left( \mathbf{U}_0 \left( x_0 \right) \right) \, \omega_i \left( \mathbf{U}_0 \left( x_0 \right) \right) \right\rbrace \right)^{-1},
\end{equation}
again subject to the convention that ${\frac{1}{0} = + \infty }$, whereupon the corresponding part of the solution fails to be even ${C^1}$-smooth, and the physically relevant (entropy) solution from that point onwards is, at best, only piecewise ${C^n}$.

Combining Lemma \ref{lemma1} with Theorem 2.1 of Mhaskar and Poggio\cite{mhaskar_deep_2016}, we obtain the main result of this section regarding error bounds for shallow neural network approximations for hyperbolic PDEs.

\begin{theorem}
Suppose that ${u : \Omega \to \mathbb{R}}$ is a solution to a (homogeneous) first-order hyperbolic PDE in conservation law form:

\begin{equation}
\frac{\partial u}{\partial t} + \frac{\partial f \left( u \right)}{\partial x} = 0,
\end{equation}
where the initial data:

\begin{equation}
u \left( 0, x \right) = u_0 \left( x \right),
\end{equation}
is given by a ${C^n}$-smooth function ${u_0 : \mathbb{R} \to \mathbb{R}}$, and the flux is given by a ${C^{n + 1}}$-smooth function ${f : \mathbb{R} \to \mathbb{R}}$, where ${n \geq 1}$. Suppose, moreover, that:

\begin{equation}
\mathcal{V}_{\text{NN}} = \left\lbrace u_{\boldsymbol\theta} \mid \boldsymbol\theta \in \mathbb{R}^P \right\rbrace,
\end{equation}
represents the space of possible functions ${u_{\boldsymbol\theta} : \Omega \to \mathbb{R}}$ that can be computed by a multi-layer perceptron with $P$ trainable parameters, comprising a single hidden layer consisting of $N$ hidden neurons, with the activation function given by a ${C^{\infty}}$-smooth function ${\sigma : \mathbb{R} \to \mathbb{R}}$ that is not a polynomial on any subinterval of ${\mathbb{R}}$. Then:

\begin{enumerate}
\item
If the flux is linear (i.e. ${f^{\prime\prime} = 0}$), or the flux is convex and the initial data is monotonically increasing (i.e. ${f^{\prime\prime} \geq 0}$ and ${u_{0}^{\prime} \geq 0}$), or the flux is concave and the initial data is monotonically decreasing (i.e. ${f^{\prime\prime} \leq 0}$ and ${u_{0}^{\prime} \leq 0}$), then there exists a neural network ${u_{\boldsymbol\theta} \in \mathcal{V}_{\text{NN}}}$ such that:

\begin{equation}
\forall t \geq 0, \qquad \left\lVert u - u_{\boldsymbol\theta} \right\rVert_{\infty} = \mathcal{O} \left( N^{-\frac{n}{d}} \right).
\end{equation}

\item
Otherwise, there exists a neural network ${u_{\boldsymbol\theta} \in \mathcal{V}_{\text{NN}}}$ such that:

\begin{equation}
\forall 0 \leq t < t_{\infty}, \qquad \left\lVert u - u_{\boldsymbol\theta} \right\rVert_{\infty} = \mathcal{O} \left( N^{-\frac{n}{d}} \right),
\end{equation}
where:

\begin{equation}
t_{\infty} = \left( \sup_{x_0 \in \mathbb{R}} \left\lbrace -f^{\prime\prime} \left( u_0 \left( x_0 \right) \right) u_{0}^{\prime} \left( x_0 \right) \right\rbrace \right)^{-1},
\end{equation}
and:

\begin{equation}
\forall t \geq t_{\infty}, \qquad \left\lVert u - u_{\boldsymbol\theta} \right\rVert_{\infty} = \mathcal{O} \left( 1 \right).
\end{equation}
In this case, there nevertheless exists a function:

\begin{equation}
\widetilde{u} \left( t, x \right) = \begin{cases}
u_{\boldsymbol\theta_1}^{\left( 1 \right)}, \qquad x \leq x_1 \left( t \right),\\
u_{\boldsymbol\theta_2}^{\left( 2 \right)}, \qquad x_1 \left( t \right) < x \leq x_2 \left( t \right),\\
\cdots,
\end{cases}
\end{equation}
where each ${u_{\boldsymbol\theta_i}^{\left( i \right)} \in \mathcal{V}_{\text{NN}}}$ is a neural network, and each ${x_i : \mathbb{R} \to \mathbb{R}}$ is a function tracking the position of a single discontinuity, such that:

\begin{equation}
\forall t \geq 0, \qquad \left\lVert u - \widetilde{u} \right\rVert_{\infty} = \mathcal{O} \left( N^{-\frac{n}{d}} \right).
\end{equation}
\end{enumerate}
\end{theorem}

\subsection{Algebraic Composability: Assembling Deep Neural Networks with Bounded Errors}
\label{sec:Section4}

The results of the preceding section indicate that shallow neural network architectures (i.e. multi-layer perceptrons consisting of a single hidden layer) can approximate ${C^n}$-functions with vanishingly small ${L^{\infty}}$ errors (i.e. ${\mathcal{O} \left( 0 \right)}$) in the smooth limit as ${n \to \infty}$, but that these approximations become arbitrarily poor (i.e. ${\mathcal{O} \left( 1 \right)}$) in the discontinuous limit as ${n \to 0}$. In particular, to obtain any reasonable bound on the ${L^{\infty}}$ error for a neural network approximation to a hyperbolic PDE permitting shocks and other weak/discontinuous solutions, even when considering piecewise approximations, the number of neurons $N$ in the hidden layer must be made impractically large, and the accuracy is still fundamentally restricted by the smoothness of the initial data. Our objective now is to ascertain whether these error bounds can be improved by considering deeper neural network architectures instead\cite{mhaskar2016learningfunctionsdeepbetter}, obtained as compositions of multiple shallower ones. For instance, suppose that, instead of attempting to approximate a PDE solution ${u \left( t, x \right)}$ directly using a shallow neural network, we instead assemble the solution as a composition:

\begin{equation}
u \left( t, x \right) = f \left( g \left( t, x \right), h \left( t, x \right) \right),
\end{equation}
of three other functions $f$, $g$, and $h$, each of which can be approximated by shallow neural networks sharing the same number of inputs and outputs. On the surface, this does not appear to represent an improvement, since if the functions $f$, $g$, and $h$ are ${C^{\alpha}}$-, ${C^{\beta}}$-, and ${C^{\gamma}}$-smooth, respectively, then their composition ${f \left( g \left( t, x \right), h \left( t, x \right) \right)}$ is ${C^{\min \left\lbrace \alpha, \beta, \gamma \right\rbrace}}$-smooth by the chain rule. Thus, at least one of the three functions in the composition must be at least as discontinuous as the function ${u \left( t, x \right)}$ itself, and therefore the bound on the ${L^{\infty}}$ error of its neural network approximation must be at least as large. However, by making a judicious choice of functions $f$, $g$, and $h$ to appear in the composition, it is possible to suppress this relatively large ${L^{\infty}}$ error in such a way that the bound on the ${L^{\infty}}$ error of the composition ${u \left( t, x \right)}$ is nevertheless smaller overall, as we shall show below. The argument generalizes inductively to an arbitrary number of functions with arbitrary arities, composed in an arbitrary manner, and therefore can be applied to construct arbitrarily deep neural network architectures with favorable ${L^{\infty}}$ bounds. We refer to this property as \textit{algebraic composability}, taking inspiration from prior work on \textit{compositional} approaches to deep learning\cite{gavranović2019compositionaldeeplearning}\cite{gavranović2024positioncategoricaldeeplearning}.

Let ${f : \mathbb{R} \to \mathbb{R}}$ and ${g : \mathbb{R} \to \mathbb{R}}$ be arbitrary functions, without any assumptions of differentiability, and let ${\widetilde{f} : \mathbb{R} \to \mathbb{R}}$ and ${\widetilde{g} : \mathbb{R} \to \mathbb{R}}$ be (shallow) neural network approximations to them with ${L^{\infty}}$ errors of ${e_f}$ and ${e_g}$, respectively, i.e:

\begin{equation}
\left\lVert f - \widetilde{f} \right\rVert_{\infty} = e_f, \qquad \left\lVert g - \widetilde{g} \right\rVert_{\infty} = e_g.
\end{equation}
We now wish to determine a bound on the ${L^{\infty}}$ error of the (deep) neural network approximation ${\widetilde{f} \circ \widetilde{g}}$ to the composition ${f \circ g}$. By applying the triangle inequality, we obtain:

\begin{align}
\left\lVert f \circ g - \widetilde{f} \circ \widetilde{g} \right\rVert_{\infty} &= \left\lVert f \circ g - f \circ \widetilde{g} + f \circ \widetilde{g} - \widetilde{f} \circ \widetilde{g} \right\rVert_{\infty}\\
&\leq \left\lVert f \circ g - f \circ \widetilde{g} \right\rVert_{\infty} + \left\lVert f \circ \widetilde{g} - \widetilde{f} \circ \widetilde{g} \right\rVert_{\infty}.
\end{align}
The first term ${\left\lVert f \circ g - f \circ \widetilde{g} \right\rVert_{\infty}}$ is effectively quantifying how much the value of $f$ varies in response to a change in its input from $g$ to ${\widetilde{g}}$ (a change which is bounded by ${\left\lVert g - \widetilde{g} \right\rVert_{\infty}}$), while the second term ${\left\lVert f \circ \widetilde{g} - \widetilde{f} \circ \widetilde{g} \right\rVert_{\infty}}$ is effectively quantifying how different the values of $f$ and ${\widetilde{f}}$ can be for a given fixed input of ${\widetilde{g}}$ (a difference which is bounded by ${\left\lVert f - \widetilde{f} \right\rVert_{\infty}}$). Introducing the modulus of continuity ${\omega_f : \mathbb{R} \to \mathbb{R}}$ of $f$ as:

\begin{equation}
\omega_f \left( \delta \right) = \sup_{\left\lvert u - v \right\rvert \leq \delta} \left\lbrace \left\lvert f \left( u \right) - f \left( v \right) \right\rvert \right\rbrace,
\end{equation}
we therefore have the bound:

\begin{align}
\left\lVert f \circ g - f \circ \widetilde{g} \right\rVert_{\infty} + \left\lVert f \circ \widetilde{g} - \widetilde{f} \circ \widetilde{g} \right\rVert_{\infty} &\leq \omega_f \left( \left\lVert g - \widetilde{g} \right\rVert_{\infty} \right) + \left\lVert f - \widetilde{f} \right\rVert_{\infty}\\
&= e_f + \omega_f \left( e_g \right).
\end{align}
If ${f : \mathbb{R} \to \mathbb{R}}$ is $L$-Lipschitz on the interval between $g$ and ${\widetilde{g}}$, then we can further bound:

\begin{equation}
\omega_f \left( e_g \right) \leq L e_g.
\end{equation}
Assembling this all together, we obtain the main result of this section regarding the behavior of ${L^{\infty}}$ errors of neural network approximations under composition:

\begin{proposition}
Suppose that ${f : \mathbb{R} \to \mathbb{R}}$ and ${g : \mathbb{R} \to \mathbb{R}}$ are arbitrary functions, subject to the hypothesis that $f$ is $L$-Lipschitz everywhere. Suppose that ${\widetilde{f} : \mathbb{R} \to \mathbb{R}}$ and ${\widetilde{g} : \mathbb{R} \to \mathbb{R}}$ are (shallow) neural network approximations to $f$ and $g$ with ${L^{\infty}}$ errors of ${e_f}$ and ${e_g}$ respectively:

\begin{equation}
\left\lVert f - \widetilde{f} \right\rVert_{\infty} = e_f, \qquad \left\lVert g - \widetilde{g} \right\rVert_{\infty} = e_g.
\end{equation}
Then the composition of (shallow) neural network approximations ${\widetilde{f} \circ \widetilde{g}}$ is a (deep) neural network approximation to the composite function ${f \circ g}$ with ${L^{\infty}}$ error bounded by:

\begin{equation}
\left\lVert f \circ g - \widetilde{f} \circ \widetilde{g} \right\rVert_{\infty} \leq e_f + L e_g.
\end{equation}
\end{proposition}

In other words, if we can decompose a discontinuous function ${u : \mathbb{R} \to \mathbb{R}}$ into a composition ${\left( f \circ g \right) : \mathbb{R} \to \mathbb{R}}$ of a smooth, slowly-varying function ${f : \mathbb{R} \to \mathbb{R}}$, and a discontinuous function ${g : \mathbb{R} \to \mathbb{R}}$, then we expect the error ${e_f}$ in the approximation of $f$ to be small, and the error ${e_g}$ in the approximation of $g$ to be large. However, if $f$ is $L$-Lipschitz everywhere, then we can suppress the arbitrarily large error in the approximation of $g$ by making the Lipschitz constant $L$ of $f$ as small as possible (in practice, this is restricted only by the range of $u$\footnote{Since the range of $f$ must be a non-strict superset of the range of $u$, and a lower bound on the Lipschitz constant $L$ of $f$ is determined by its range.}), such that the product ${L e_g}$ remains as small as possible, and the overall error in the approximation of ${u = f \circ g}$ becomes dominated, if possible, by the relatively small error ${e_f}$ in the approximation of $f$. This effectively gives us a prescription for how to construct deep neural network architectures for the approximation of discontinuous functions (e.g. solutions to hyperbolic PDEs permitting shocks and other discontinuous solutions) with favorable ${L^{\infty}}$ bounds: decompose the discontinuous function into a composition of a large number of smooth, slowly-varying functions, with all of the discontinuities ``packaged'' into a single, highly discontinuous function appearing right at the start of the chain of compositions. Then, suppress the large errors in the approximation of this first, discontinuous function to the greatest extent possible, by forcing the Lipschitz constants of the slowly-varying functions to be as small as possible. This procedure is precisely the prescription that BEACONS uses for assembling deep neural network approximations to PDE solutions with sharply bounded ${L^{\infty}}$ errors. Note that this basic idea, namely that one can improve the accuracy and stability of a numerical approximation in the vicinity of a shock (or other discontinuity or region of sharp gradient) by composing the approximate solution with a function that artificially reduces its gradient, is also the key insight behind the theory of \textit{flux limiters}\cite{leveque_finite_2011} and high-order \textit{total variation diminishing} (TVD) schemes\cite{harten_high_1997} in the classical numerical analysis of hyperbolic PDEs. In this respect, our results on the algebraic compositionality of deep BEACONS networks may be regarded as a generalized (and iterated) version of the classical circumvention of Godunov's theorem by means of a non-linear flux limiter.

As a minimal example of how such a decomposition ${u \left( x \right) = f \left( g \left( x \right) \right)}$, of a discontinuous function ${u : \mathbb{R} \to \mathbb{R}}$ into a smooth, slowly-varying ($L$-Lipschitz) function ${f : \mathbb{R} \to \mathbb{R}}$ and a discontinuous function ${g : \mathbb{R} \to \mathbb{R}}$, could be constructed, consider the trivial case:

\begin{equation}
f \left( x \right) = \frac{x}{C}, \qquad \text{ and } \qquad g \left( x \right) = C u \left( x \right),
\end{equation}
where ${C > 0}$ is some positive constant, such that $f$ is ${C^{\infty}}$-smooth, and the Lipschitz constant of $f$ is given by ${L = \frac{1}{C}}$. Clearly this trivial case does not confer any actual advantage, since although we can make the Lipschitz constant $L$ of $f$ arbitrarily small by increasing the value of $C$, this also has the effect of increasing the error ${e_g}$ in the approximation of $g$ by exactly the same amount. However, within (the failure of) this minimal example lies the key insight to the general algorithm: select a ${C^{\infty}}$-smooth function $f$ whose Lipschitz constant $L$ can be precisely controlled, whose range is a non-strict superset of the range of $u$:

\begin{equation}
\mathrm{range} \left( u \right) \subseteq \mathrm{range} \left( f \right),
\end{equation}
and which is either strictly-increasing or strictly-decreasing, such that its inverse ${f^{-1} : \mathbb{R} \to \mathbb{R}}$ is unique, and therefore:

\begin{equation}
g \left( x \right) = f^{-1} \left( u \left( x \right) \right),
\end{equation}
can be constructed unambiguously, such that the error ${e_g}$ in the approximation of $g$ grows as less than ${\frac{1}{L}}$ as $L$ becomes small. For example, a good general candidate is:

\begin{equation}
f \left( x \right) = \frac{\mathrm{arcsinh} \left( x \right)}{C}, \qquad \text{ and } \qquad g \left( x \right) = \sinh \left( C u \left( x \right) \right),
\end{equation}
where again ${C > 0}$ is some positive constant such that the Lipschitz constant of $f$ is given by ${L = \frac{1}{C}}$. The operative distinction between this case and the previous (trivial) case lies in the non-linearity of ${\mathrm{arcsinh} \left( x \right)}$; again, this is analogous to the theory of high-order TVD schemes for hyperbolic PDEs, wherein the flux limiter must necessarily be non-linear in order to circumvent the axioms of Godunov's theorem successfully. The BEACONS framework automatically trials several such candidate functions (primarily involving simple trigonometric and hyperbolic functions, and their inverses) until it finds one which minimizes the growth rate in the error ${e_g}$ in the approximation of $g$, as a function of ${\frac{1}{L}}$. Each layer of the BEACONS network is then trained separately, with its own specialized loss function, via the standard backpropagation algorithm for multi-layer perceptrons.

\section{Software Implementation}
\label{sec:Section5}

In the authors' previous work\cite{gorard2025shockconfidenceformalproofs} on formally-verified numerical solvers for hyperbolic PDEs, a domain-specific language (DSL) for representing hyperbolic PDE systems in Racket\cite{felleisen_et_al} was introduced, along with an automatic code-generator that was capable of converting these high-level Racket representations into optimized C code for a variety of different numerical algorithms, and an automated theorem-prover that was able to prove fully executable (and therefore fully machine-checkable) correctness theorems regarding the resulting solvers. The Racket-based DSL was designed to be sufficiently general as to encode both individual scalar PDEs and coupled (vector) systems of PDEs, in any number of dimensions. The code-generator was then able to generate standalone finite volume numerical solvers for these equations using either Lax-Friedrichs (low-resolution) fluxes\cite{lax_weak_1954} or Roe (high-resolution) fluxes\cite{roe_approximate_1997}, and using either a fully unsplit method (in 1D), or a second-order Strang split method (in higher dimensions)\cite{strang_construction_1968}. Although naively these solvers are all first-order accurate in space, the code-generator was also able to extend the spatial accuracy to second-order in all smooth regions by replacing the piecewise-constant representation of the solution with a piecewise-linear reconstruction obtained using a given choice of flux limiter (e.g. minmod, monotonized-centered, van Leer, or super-bee). Finally, the theorem-proving framework was then able to prove various correctness properties of the resulting numerical schemes, such as (strict) hyperbolicity preservation, CFL stability, convexity/local Lipschitz continuity, and flux continuity, in the case of the Lax-Friedrichs and Roe solvers, as well as symmetry and second-order total variation diminishing (TVD), in the case of the flux limiters for second-order flux extrapolation. The generated proofs were all, themselves, standalone pieces of symbolic Racket code which could be executed (with this execution constituting an automatic check of the validity of the proof), and which could therefore be regarded as executable \textit{certificates of correctness}. One of the key design principles underlying the theorem-prover was that only symbolic transformations respecting the underlying algebraic structure of the IEEE-754 standard for floating-point arithmetic should be permitted, since only these transformations would guarantee preservation of correctness of the corresponding generated C code (we thus referred to such transformations as \textit{strictly correctness-preserving} algebraic transformations). To this end, we constructed bespoke algorithms for automatic symbolic simplification (\texttt{symbolic-simp}), automatic symbolic differentiation (\texttt{symbolic-diff}), and automatic symbolic evaluation of limits (\texttt{evaluate-limit}) of arbitrary Racket expressions.

\begin{figure}[ht]
\centering
\fbox{\includegraphics[width=0.42\textwidth]{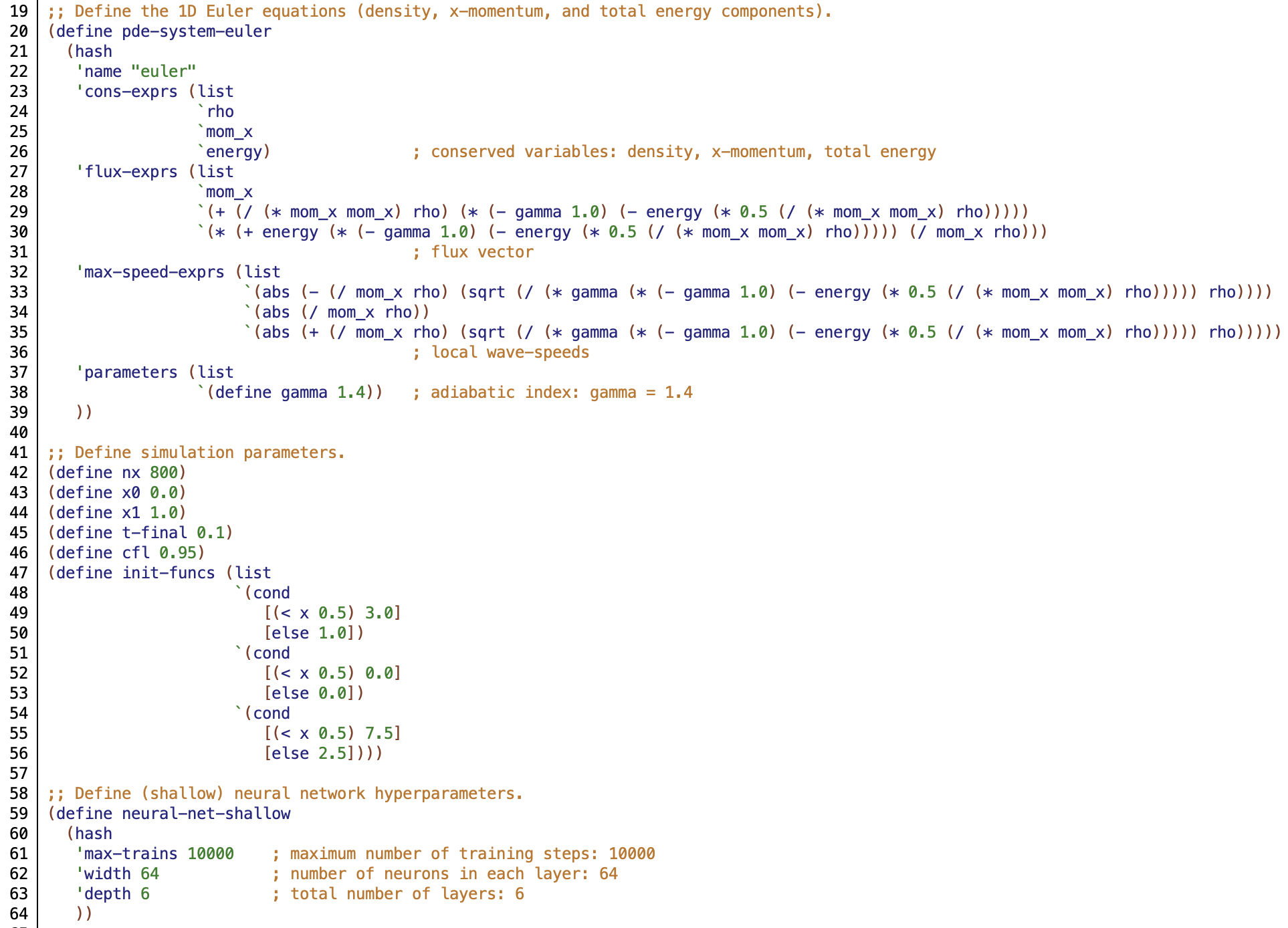}}
\fbox{\includegraphics[width=0.47\textwidth]{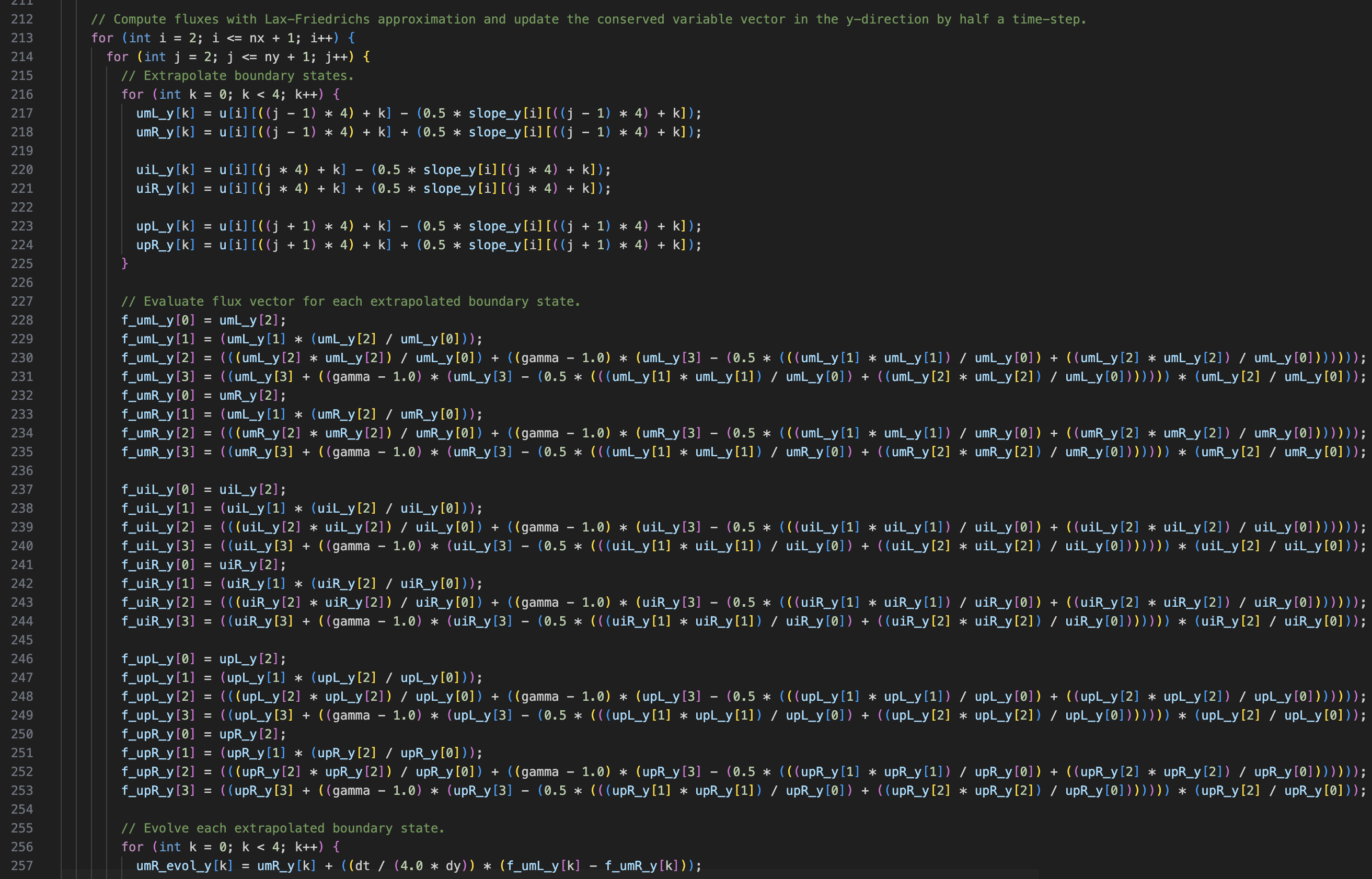}}
\caption{On the left, an example of the extended Racket data structure for representing a system of hyperbolic PDEs (in this case, the 1D compressible Euler equations), a collection of simulation parameters (in this case, representing a 1D Sod-type shock tube problem), and a collection of hyperparameters for a shallow BEACONS architecture (6 layers, 64 neurons per layer, 10,000 maximum training steps). On the right, the resulting optimized C code output by the automatic code-generator for running the specified simulation (using a Lax-Friedrichs solver with second-order flux extrapolation), generating the necessary training data, and training the BEACONS architecture accordingly.}
\label{fig:Figure7}
\end{figure}

\begin{figure}[ht]
\centering
\fbox{\includegraphics[width=0.445\textwidth]{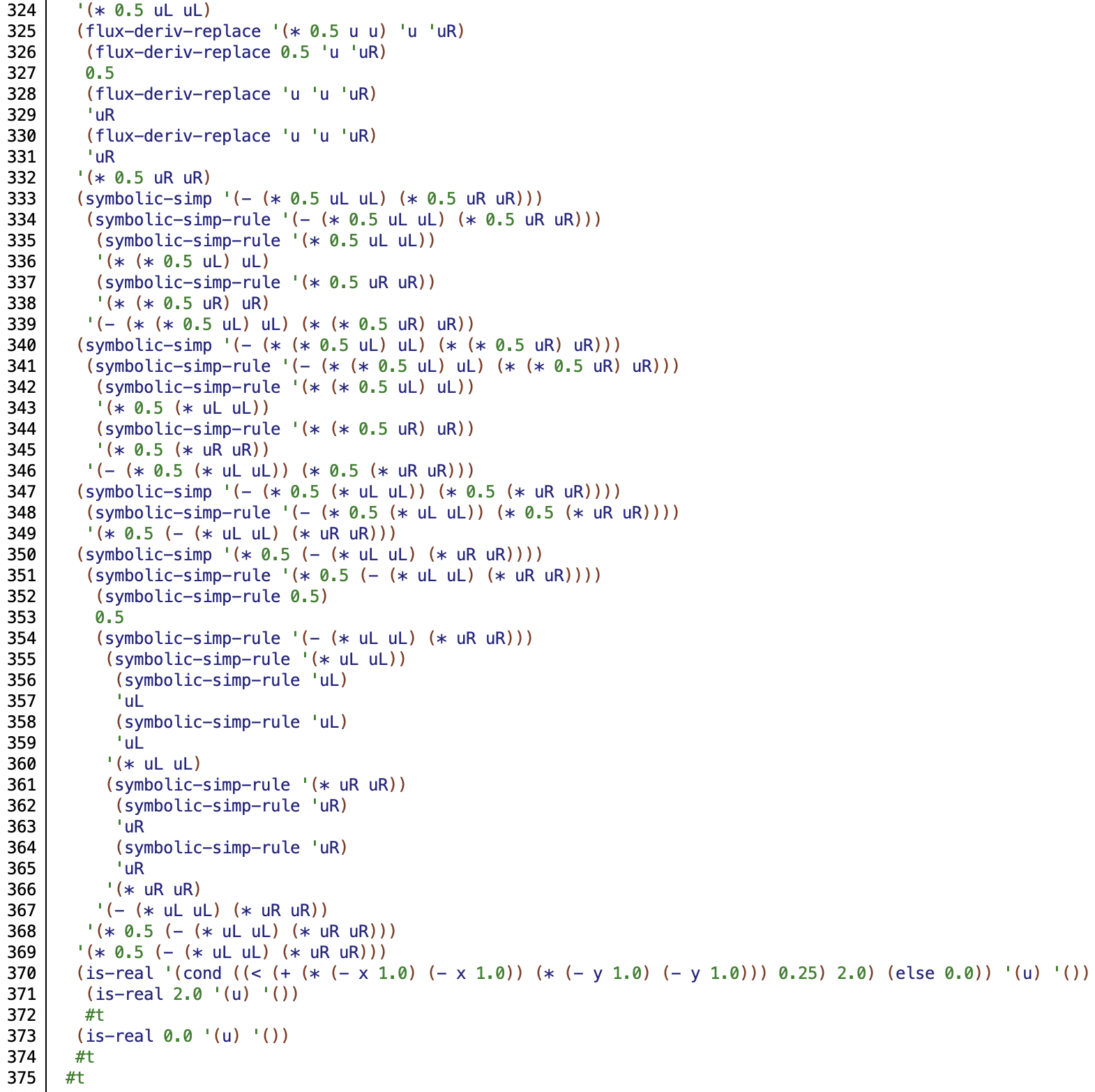}}
\fbox{\includegraphics[width=0.3\textwidth]{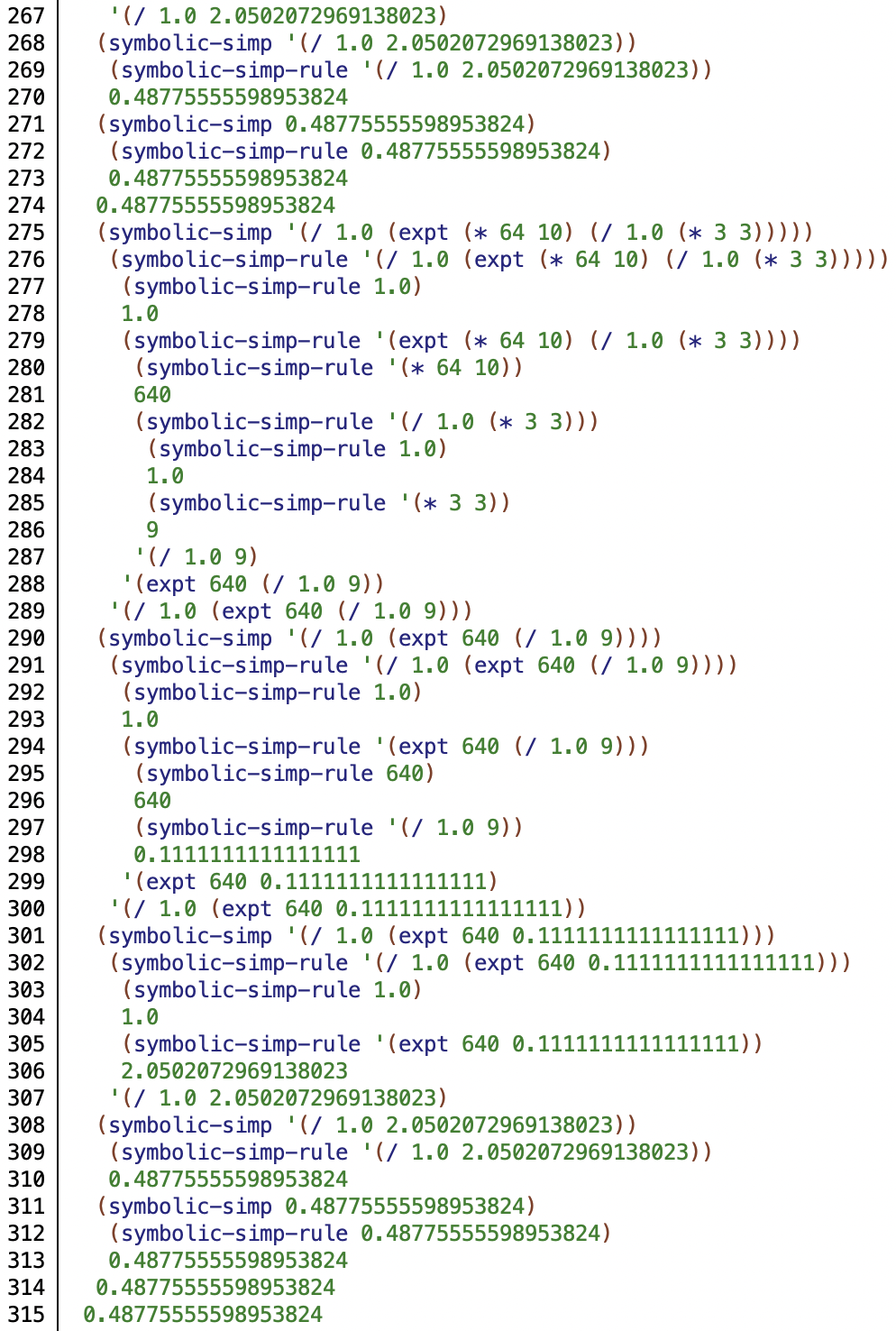}}
\caption{On the left, the final few steps of an executable Racket proof of flux continuity for a Roe solver for the 1D inviscid Burgers' equation. On the right, the final few steps of an executable Racket proof of a bound on the worst-case ${L^{\infty}}$ error (for the non-smooth parts of the solution) for a shallow BEACONS architecture for solving the 1D inviscid Burgers' equation.}
\label{fig:Figure8}
\end{figure}

In the present work, we have extended the Racket-based DSL to support the specification of certain hyperparameters for BEACONS architectures (namely width, depth, and maximum number of training steps). We have also extended the code-generator to support the automatic synthesis of optimized C code for both training and validating BEACONS networks for each equation system, with the training data being entirely supplied by formally-verified numerical solvers. Although the solvers themselves are fully dependency-free, the neural network training and validation is performed using the minimalist \texttt{kann}\footnote{\url{https://github.com/attractivechaos/kann}} neural network library. Finally, we have also extended the theorem-proving framework to facilitate the generation of automatic proofs of worst-case ${L^{\infty}}$ error bounds for the BEACONS approximations, based on the analytical properties of the flux function (or the eigensystem of its Jacobian), the analytical properties of the initial data, and the BEACONS hyperparameters. As described in the previous section, this theorem-proving algorithm begins by using the smoothness of the initial data and the analytical properties of the flux to prove a worst-case ${L^{\infty}}$ bound on the approximation of the solution by a shallow neural network architecture (i.e. a multi-layer perceptron with a single hidden layer), and then proceeds to trial possible decompositions of the solution into compositions of various smooth candidate functions combined with a single discontinuous function. It continues this process until it finds a decomposition which minimizes the worst-case ${L^{\infty}}$ error in the approximation of the overall composition by means of a deep neural network architecture (subject to the constraint that this architecture is consistent with the desired BEACONS hyperparameters). In Figure \ref{fig:Figure7}, we show an example of the 1D compressible Euler equations being represented using a data structure within our Racket-based DSL, plus additional data structures for representing the initial conditions of a 1D Sod-type shock tube problem, along with the hyperparameters for a shallow BEACONS network consisting of 6 layers, with 64 neurons per layer, and 10,000 maximum training steps. We then also show part of the output from the automatic code-generator for this particular example, illustrating how a Lax-Friedrichs solver with second-order flux extrapolation can be used to generate arbitrary amounts of training data, which is then automatically used to train (and subsequently validate) the desired BEACONS architecture. In Figure \ref{fig:Figure8}, we show two examples of output from the automated theorem-proving framework, corresponding to fully executable Racket proofs of flux continuity for a 1D Roe solver for the inviscid Burgers' equation (i.e. a correctness property of the underlying numerical solver), and of a bound on the worst-case ${L^{\infty}}$ error for the non-smooth parts of a shallow BEACONS architecture for approximating the output of this solver (i.e. a correctness property of the BEACONS architecture trained on the output of the solver).

As previously, the theorem-proving algorithm is implemented in a purely \textit{equational} fashion, by constructing a \textit{globally confluent} and \textit{strongly normalizing} symbolic rewriting system\cite{baader_term_1999}\cite{robinson_handbook_2001}, and then proceeding to apply these rewriting rules to an arbitrary symbolic Racket expression until the rewriting sequence terminates at a normal form; if two Racket expressions terminate at the same normal form, then this rewriting sequence is interpreted as a proof that the original expressions were symbolically equivalent. It is worth emphasizing that the proofs of correctness for the underlying numerical solvers and the proofs of correctness for the BEACONS architectures trained on the output of these solvers are conceptually distinct, and logically independent of one another. The former class of proofs effectively guarantee that the function represented by the training data is mathematically correct (i.e. agrees with the true analytical solution to the underlying PDE system in the appropriate limit), even if this function cannot itself be accurately approximated by a given BEACONS architecture. On the other hand, the latter class of proofs effectively guarantee that the BEACONS architecture is correctly computing an approximation (with bounded error) to the function represented by the training data, even if this function is not itself mathematically correct with respect to the underlying PDE system. Of course, it is preferable to have both guarantees wherever possible, but there may be occasions where it is not possible to produce a fully formally-verified underlying numerical solver, yet it may nevertheless be desirable to synthesize a formally-verified BEACONS architecture using that solver as a source of training data \textit{subject to the unproven assumption} that the solver is correct. We will encounter precisely this scenario later on in the paper, in the context of the compressible Euler equations; in this case, no formally-verified solver currently exists, yet a BEACONS architecture with bounded error can nevertheless be synthesized subject to the assumption that the generated solver is correct.

\section{Numerical Results}
\label{sec:Section6}

Note that, for all examples presented within this section, both the BEACONS architectures and the non-BEACONS (fully-connected) neural network architectures use the hyperbolic tangent loss function ${\sigma \left( x \right) = \tanh \left( x \right)}$, since it satisfies the requisite ${C^{\infty}}$-smoothness hypothesis for Theorem 2.1 of Mhaskar and Poggio\cite{mhaskar_deep_2016}. The learning rate in all cases is set to ${10^{-4}}$, the maximum number of epochs to 50, and the minimum number of epochs to 10. Fully-connected neural networks are trained in a single shot, on the entire PDE solution $u$, using standard full-batch\footnote{We opt to use full-batch gradient descent, as opposed to stochastic gradient descent with mini-batching, in order to make the results fully deterministic and hence more easily reproducible. The problems described here are sufficiently small that the computational cost of doing this is not prohibitive. Extension of these methods to stochastic gradient descent with mini-batching is straightforward.} gradient descent with backpropagation. BEACONS architectures are trained layer-by-layer, with each layer trained on a single constituent function $f$ appearing in the composition of the overall solution $u$ (with its own specialized loss function), again using standard full-batch gradient descent with backpropagation, as described in Section \ref{sec:Section5}.

\subsection{Linear Advection Equation}
\label{sec:Section7}

\begin{figure}[ht]
\centering
\includegraphics[width=0.495\textwidth]{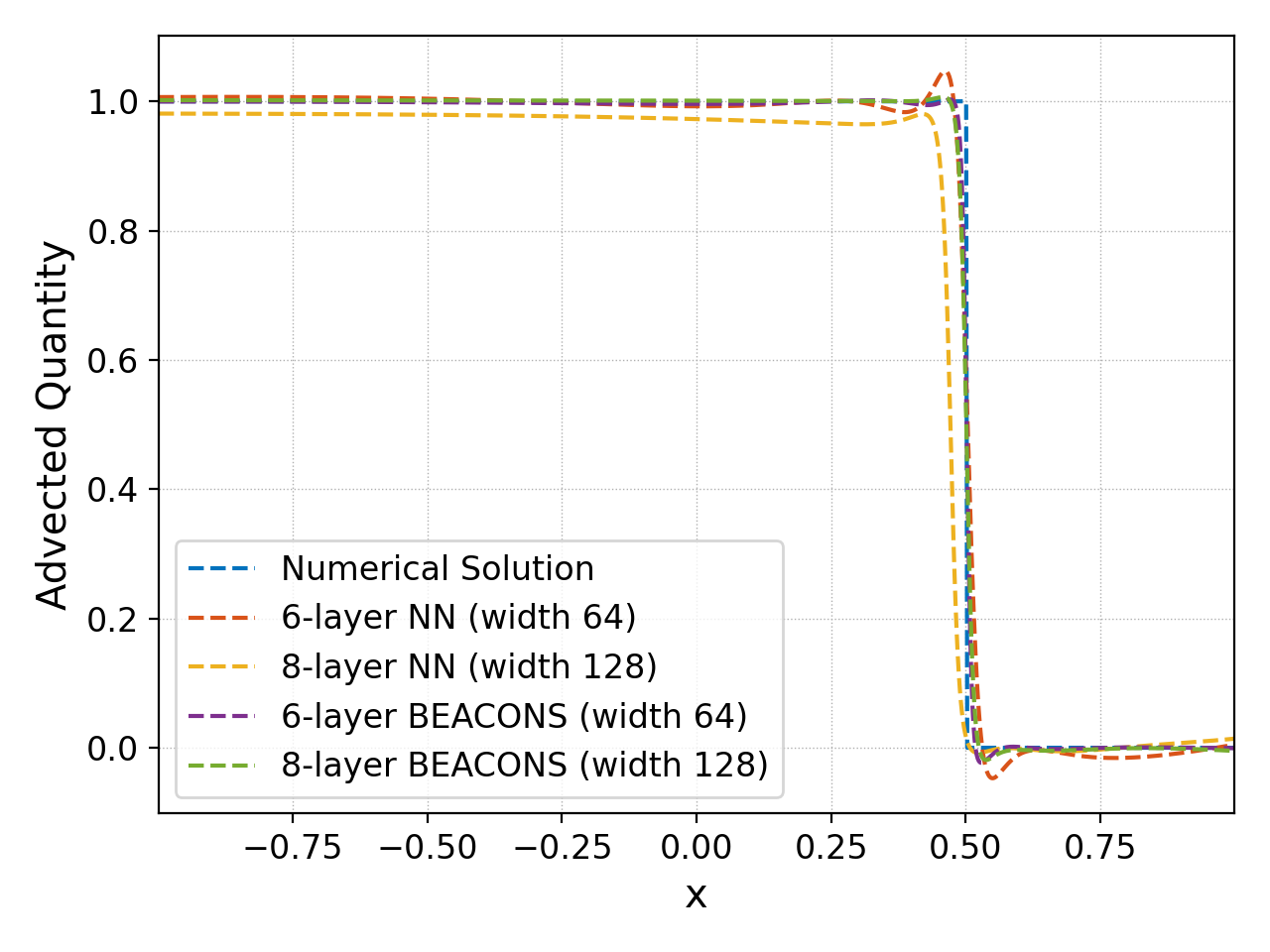}
\includegraphics[width=0.495\textwidth]{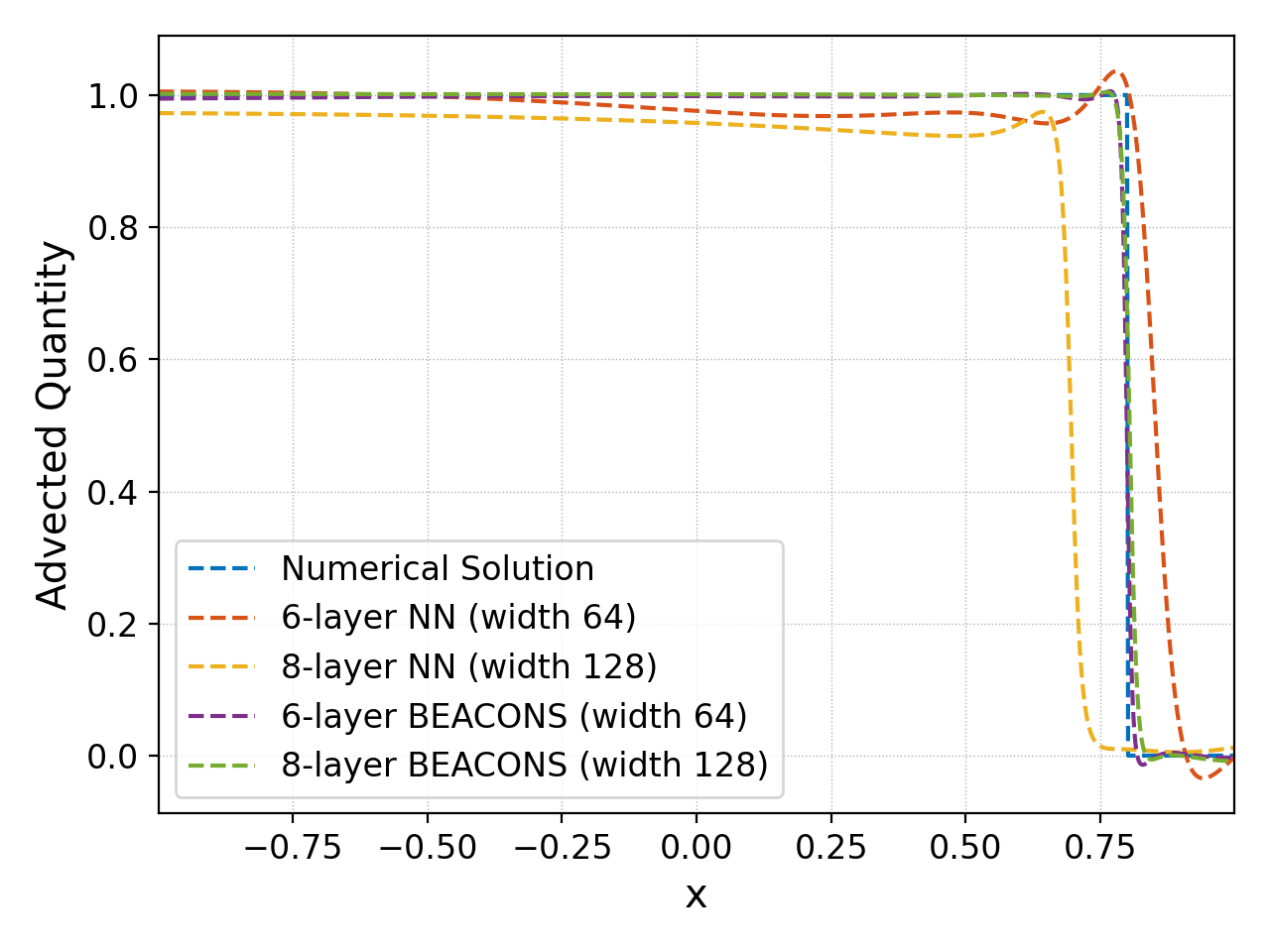}
\caption{Results for the 1D linear advection Riemann problem at times ${t = 0.5}$ (left) and ${t = 0.8}$ (right), obtained using a formally-verified numerical solver (blue), a 6-layer neural network (red), an 8-layer neural network (yellow), a 6-layer BEACONS architecture (purple), and an 8-layer BEACONS architecture (green). The 6-layer neural network slightly overestimates the advection speed, and the 8-layer neural network significantly underestimates it; both neural network solutions exhibit substantial conservation errors. Both BEACONS solutions track the numerical solution more-or-less perfectly.}
\label{fig:Figure1}
\end{figure}

Our first numerical test case will be to solve the scalar \textit{linear advection} equation:

\begin{equation}
\frac{\partial u}{\partial t} + \frac{\partial \left( a u \right)}{\partial x} = 0,
\end{equation}
i.e. a scalar conservation equation with linear flux ${f \left( u \right) = a u}$, where ${a \in \mathbb{R}}$ is an arbitrary (constant) advection speed. Our automated theorem-proving framework is able to produce full proofs of correctness for both the Lax-Friedrichs and Roe-type finite volume solvers for this equation; hyperbolicity-preservation, CFL stability, and local Lipschitz continuity of the Lax-Friedrichs solver require a total of 38, 50, and 44 proof steps to establish, respectively, while hyperbolicity-preservation and flux conservation of the Roe-type solver require a total of 60 and 97 proof steps to establish, respectively. The first test problem will be a one-dimensional Riemann problem over the spatial domain ${\left[ -1.0, 1.0 \right]}$, with advection speed ${a = 1.0}$, and initial data given by:

\begin{equation}
u_0 \left( x \right) = \begin{cases}
1.0, \qquad & \text{ for } x \leq 0.0,\\
0.0, \qquad & \text{ for } x > 0.0.
\end{cases}
\end{equation}
We shall solve this problem numerically using the formally-verified \textit{Roe-type} approximate Riemann solver (equivalent to an exact Riemann solver due to the linearity of the problem), with a CFL coefficient ${C_{\text{CFL}} = 1.0}$, and with a spatial resolution of 1024 cells. We evolve the problem numerically up until a final time of ${t = 1.0}$, producing 100 frames of output in the process. We then train four different neural network architectures on a subset of this output, namely the first 33 frames (up until time ${t = 0.33}$), and then ask each of them to predict the remainder of the simulation based on the learned solution function ${u \left( t, x \right)}$. The chosen neural network architectures are: a 6-layer BEACONS architecture with 64 neurons per layer, an 8-layer BEACONS architecture with 128 neurons per layer, and, for means of comparison, a 6-layer fully-connected (non-BEACONS) neural network with 64 neurons per layer, and an 8-layer fully-connected (non-BEACONS) neural network with 128 neurons per layer. For this problem, our automated theorem-proving framework proves a worst-case ${L^{\infty}}$ error of 0.903602 (across both the smooth and non-smooth parts of the solution) for the 6-layer BEACONS architecture, and 0.707106 (across both the smooth and non-smooth parts of the solution) for the 8-layer BEACONS architecture. In each case, the theorem-prover requires a total of 99 proof steps to establish the bound. The (inferred) solutions from the formally-verified numerical solver, the 6-layer neural network, the 8-layer neural network, the 6-layer BEACONS architecture, and the 8-layer BEACONS architecture, at times ${t = 0.5}$ and ${t = 0.8}$, are shown in Figure \ref{fig:Figure1}.

We see that the 6-layer neural network solution exhibits large overshoots and undershoots in the region surrounding the propagating discontinuity, and slightly overestimates the overall advection speed. These overshoots and undershoots are reduced in the 8-layer neural network solution, which is broadly more stable, but the advection speed is now significantly underestimated. Both neural network solutions visibly fail to conserve the advected quantity to any significant extent, with the conservation errors in the 8-layer case being especially severe. Both of the BEACONS solutions remain very close to the numerical solution, with the advection speed estimated more-or-less perfectly. The 6-layer BEACONS solution exhibits some small overshoots and undershoots in the region surrounding the propagating discontinuity (similar to the 6-layer neural network solution). These overshoots and undershoots are slightly reduced in the 8-layer BEACONS solution, which appears marginally more diffusive in some respects, and both BEACONS solutions appear to conserve the advected quantity at least approximately. Table \ref{tab:Table1} shows the normalized ${L^2}$ and ${L^{\infty}}$ errors (as compared against the formally-verified numerical solution) across the four different neural network architectures, both for the final frame only, and across all frames (using a per-frame average for the ${L^2}$ error, and normalizing across all frames for the ${L^{\infty}}$ error). We see overall that both the final-frame and all-frame ${L^2}$ and ${L^{\infty}}$ errors are significantly lower for the two BEACONS architectures than for the non-BEACONS architectures. In both the BEACONS and non-BEACONS cases, we see that increasing the number of layers has the effect of decreasing the ${L^{\infty}}$ errors at the expense of increasing the ${L^2}$ errors (in the final-frame case for the non-BEACONS architectures, this increase is particularly severe). Table \ref{tab:Table2} shows the normalized conservation errors (as obtained by integrating over the conserved quantity, and comparing against the integral of the formally-verified numerical solution) across the four different neural network architectures, again both for the final frame only, and integrated across all frames. We see overall that both the final-frame and all-frame conservation errors are significantly lower for the two BEACONS architectures than for the non-BEACONS architectures. In the non-BEACONS case, we see that increasing the number of layers has the effect of dramatically increasing both the final-frame and all-frame conservation errors. In the BEACONS case, we observe the opposite effect, with a dramatic decrease in all conservation errors for the 8-layer BEACONS architecture over the 6-layer BEACONS architecture. Note that the largest ${L^{\infty}}$ errors observed for the two BEACONS architectures (i.e. 0.782192 and 0.633319, respectively) remain comfortably below the worst-case bounds (i.e. 0.903602 and 0.707106, respectively).

\begin{table}[ht]
\centering
\begin{tabular}{|c||c|c|c|c|}
\hline
Architecture & ${L^{\infty}}$ Error (Final) & ${L^2}$ Error (Final) & ${L^{\infty}}$ Error (All) & ${L^2}$ Error (All)\\
\hline\hline
6-layer NN & 1.033641 & 2.536744 & 1.076130 & 3.002461\\
\hline
8-layer NN & 0.976984 & 9.030859 & 1.002587 & 4.807408\\
\hline
6-layer BEACONS & 0.612160 & 1.132093 & 0.782192 & 1.061877\\
\hline
8-layer BEACONS & 0.605036 & 1.211529 & 0.633319 & 1.189765\\
\hline
\end{tabular}
\caption{${L^2}$ and ${L^{\infty}}$ error analysis for the 1D linear advection Riemann problem, comparing the 6-layer neural network, 8-layer neural network, 6-layer BEACONS, and 8-layer BEACONS solutions against the formally-verified numerical solution, both for the final predicted frame, and across all predicted frames.}
\label{tab:Table1}
\end{table}

\begin{table}[ht]
\centering
\begin{tabular}{|c||c|c|}
\hline
Architecture & Conservation Error (Final) & Conservation Error (Total)\\
\hline\hline
6-layer NN & -21.169941 & 498.706193\\
\hline
8-layer NN & -130.422471 & -4391.640256\\
\hline
6-layer BEACONS & -3.770073 & -155.321387\\
\hline
8-layer BEACONS & 1.881948 & 49.570335\\
\hline
\end{tabular}
\caption{Conservation analysis for the 1D linear advection Riemann problem, comparing the 6-layer neural network, 8-layer neural network, 6-layer BEACONS, and 8-layer BEACONS solutions against the formally-verified numerical solution, both for the final predicted frame, and across all frames.}
\label{tab:Table2}
\end{table}

\begin{figure}[ht]
\centering
\includegraphics[trim={1.5cm, 0cm, 1.5cm, 0cm}, clip, width=0.33\textwidth]{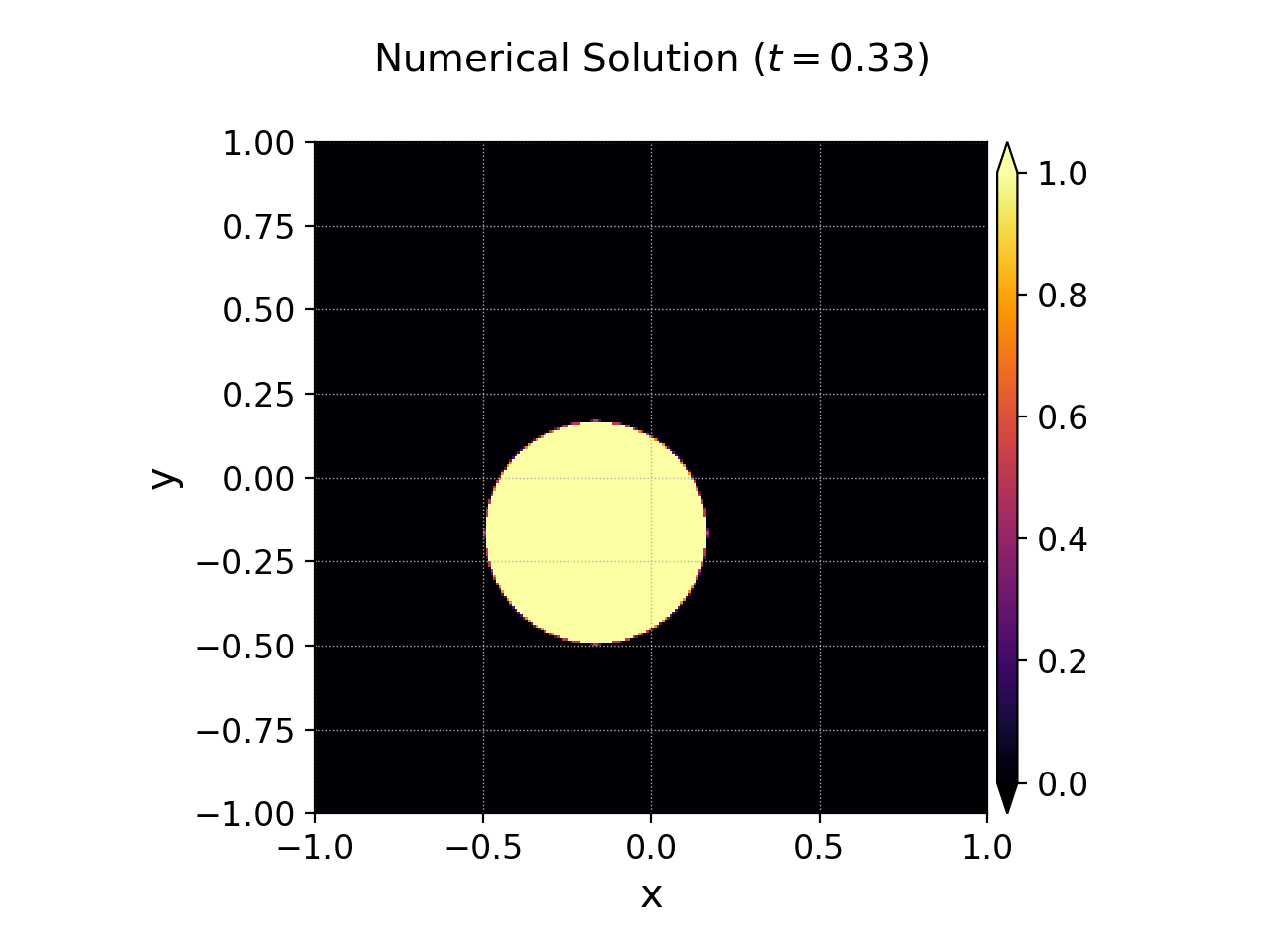}
\includegraphics[trim={1.5cm, 0cm, 1.5cm, 0cm}, clip, width=0.33\textwidth]{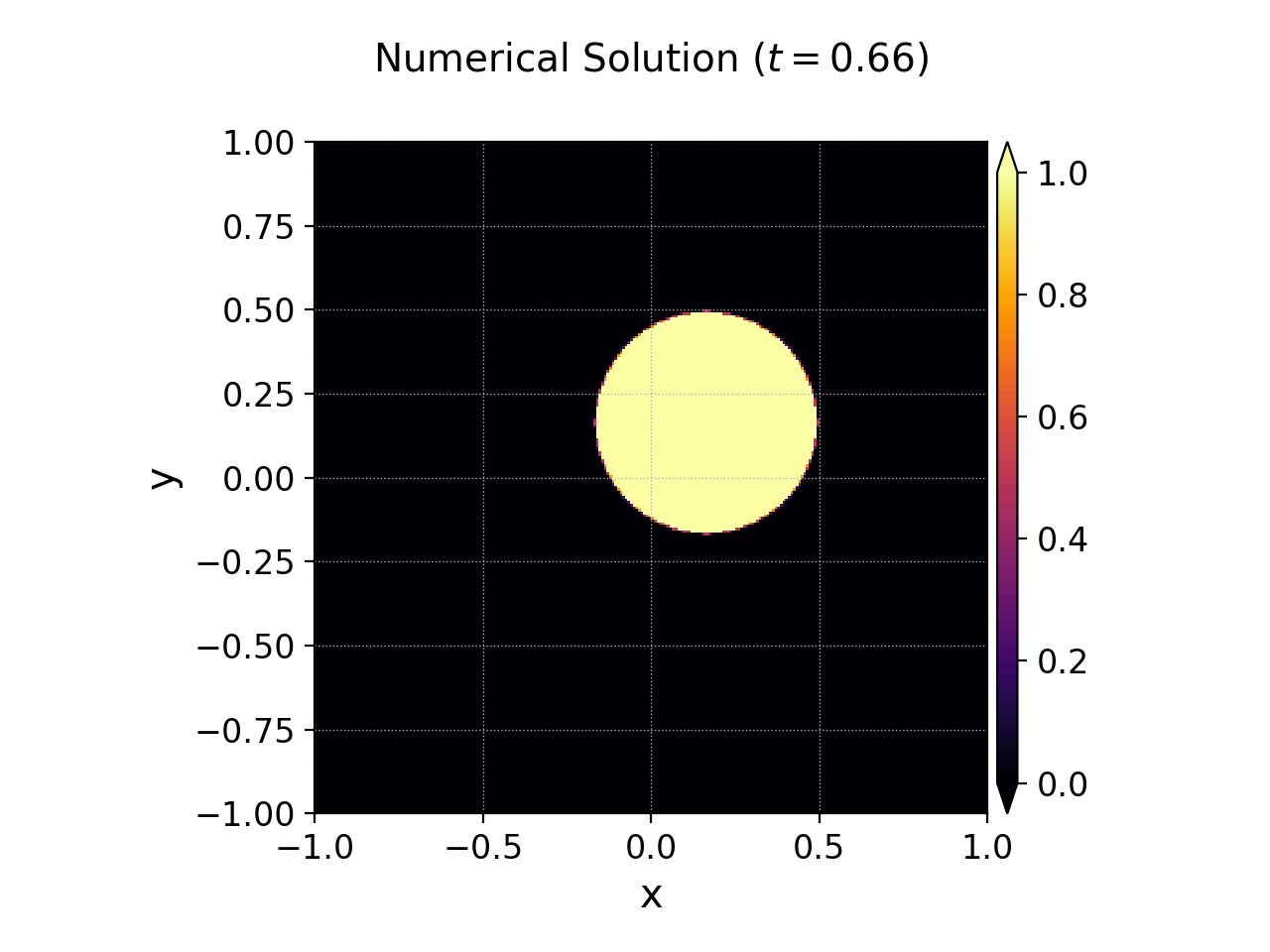}
\includegraphics[trim={1.5cm, 0cm, 1.5cm, 0cm}, clip, width=0.33\textwidth]{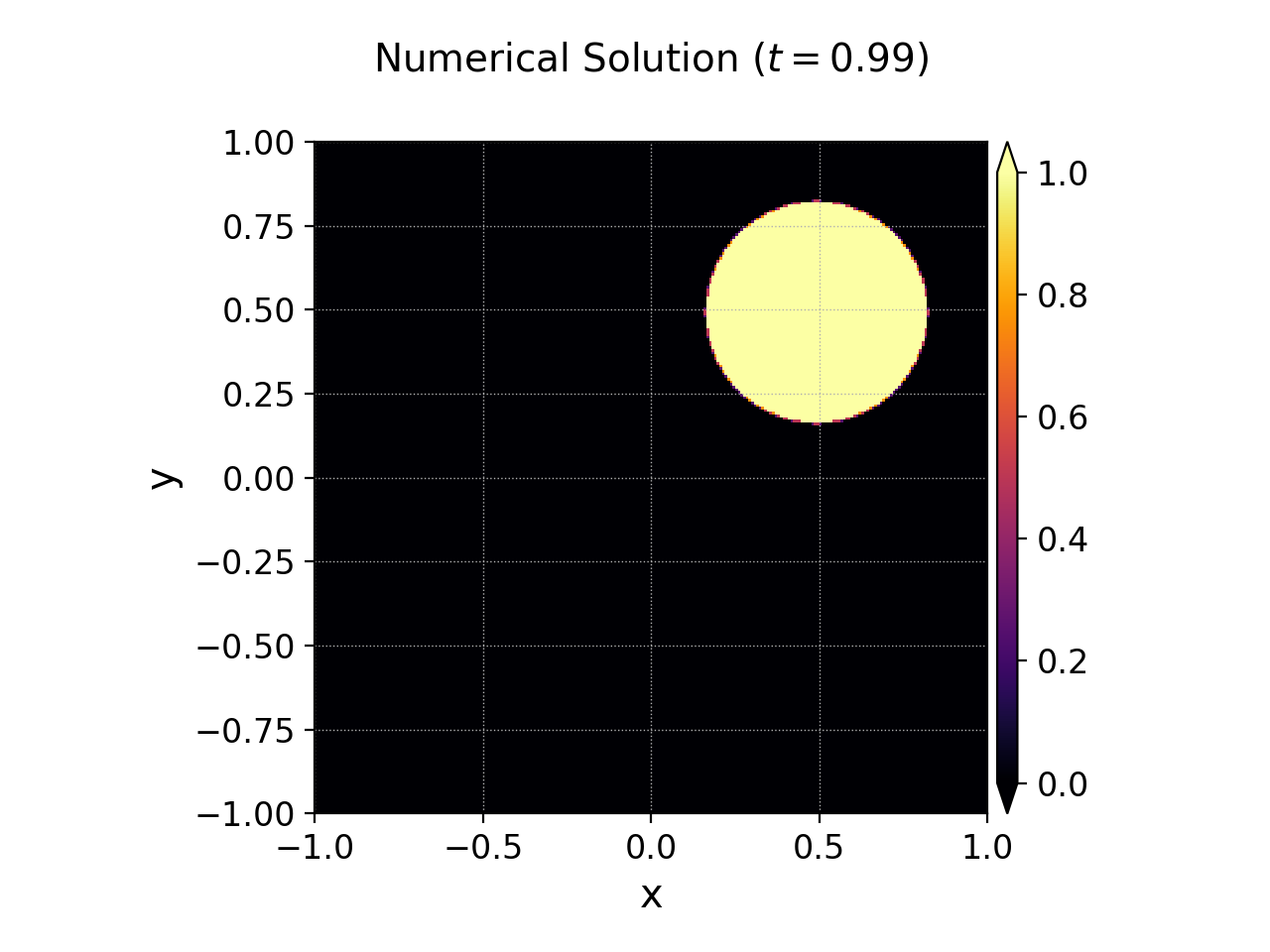}
\includegraphics[trim={1.5cm, 0cm, 1.5cm, 0cm}, clip, width=0.33\textwidth]{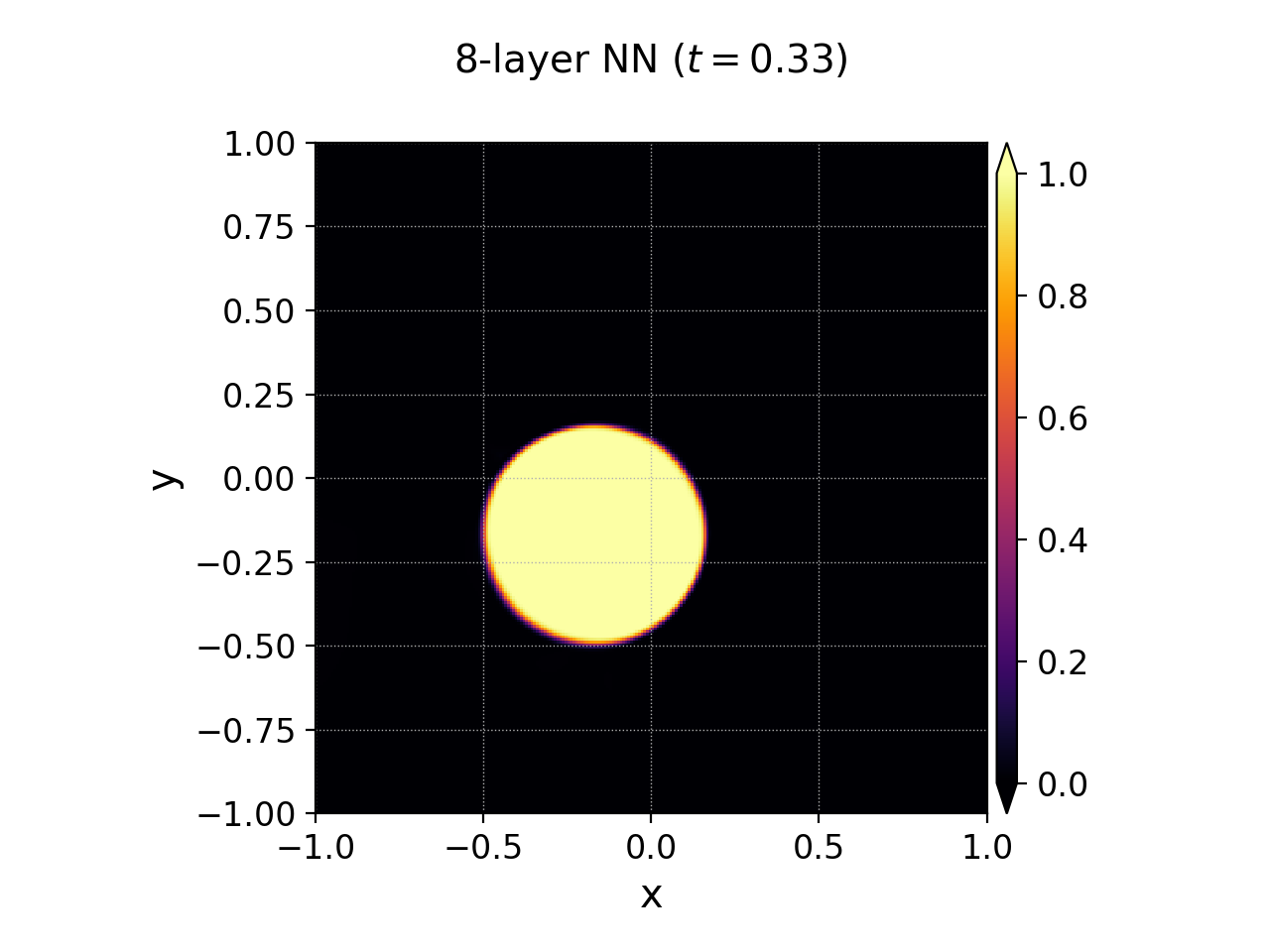}
\includegraphics[trim={1.5cm, 0cm, 1.5cm, 0cm}, clip, width=0.33\textwidth]{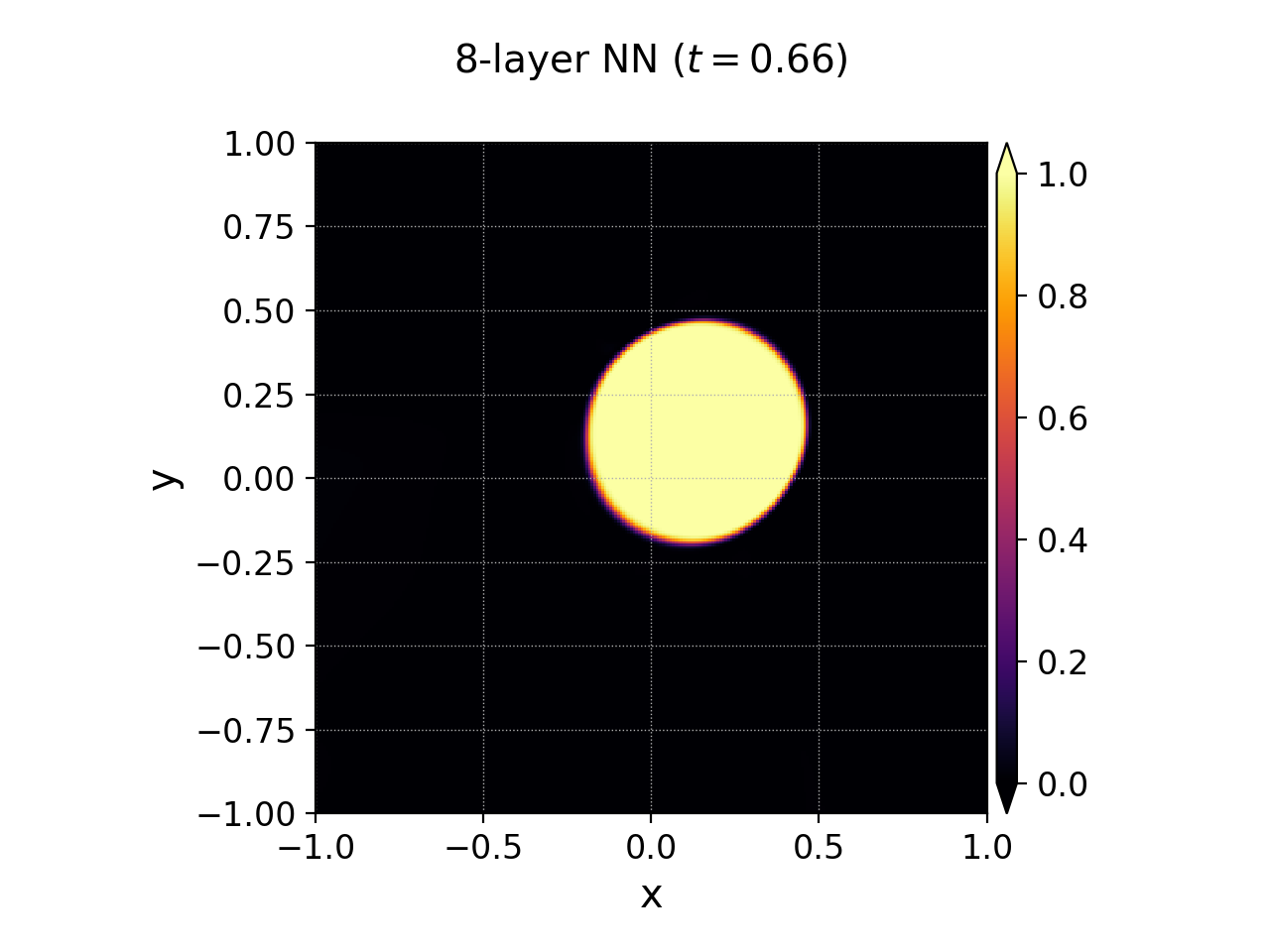}
\includegraphics[trim={1.5cm, 0cm, 1.5cm, 0cm}, clip, width=0.33\textwidth]{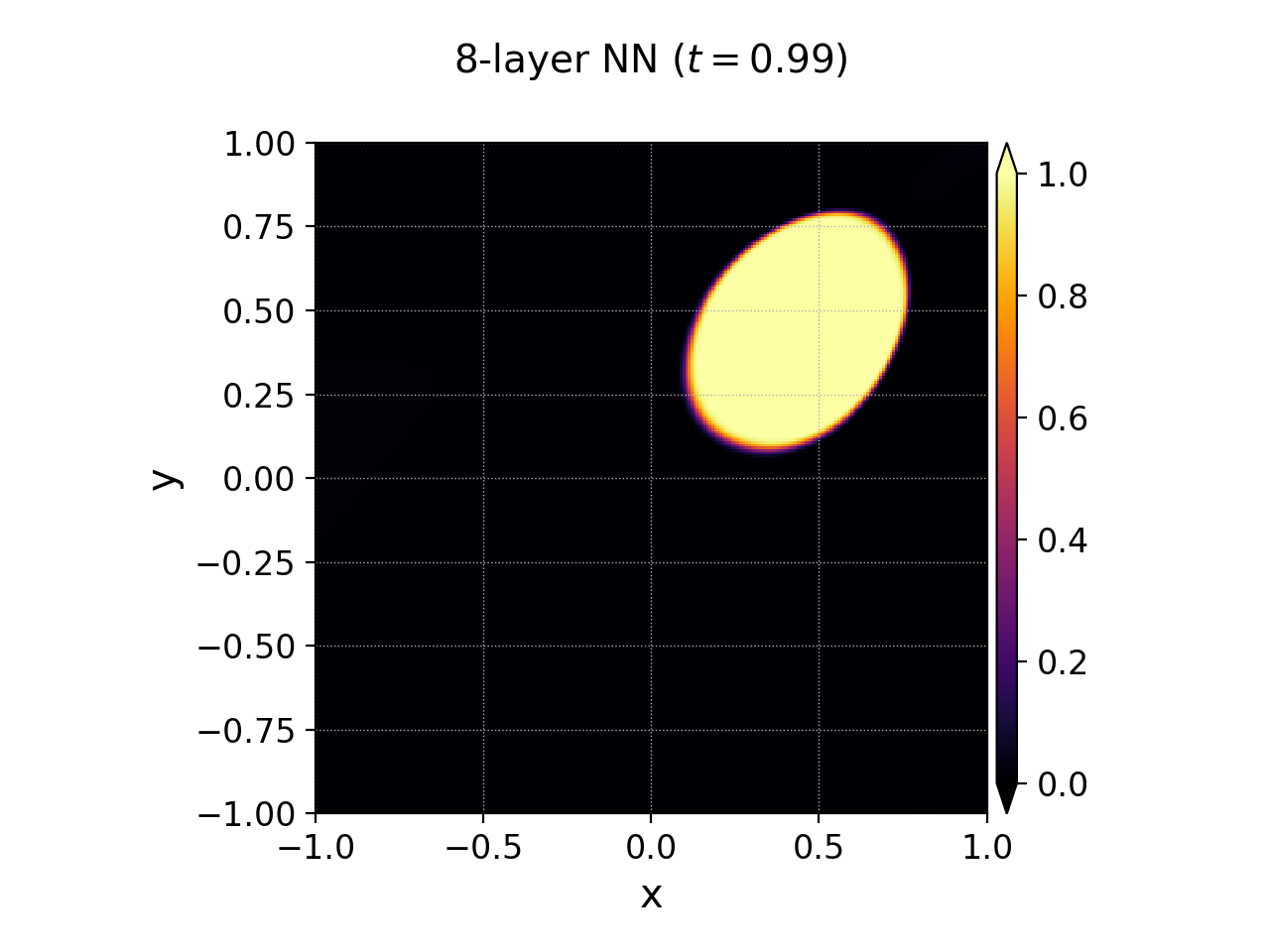}
\includegraphics[trim={1.5cm, 0cm, 1.5cm, 0cm}, clip, width=0.33\textwidth]{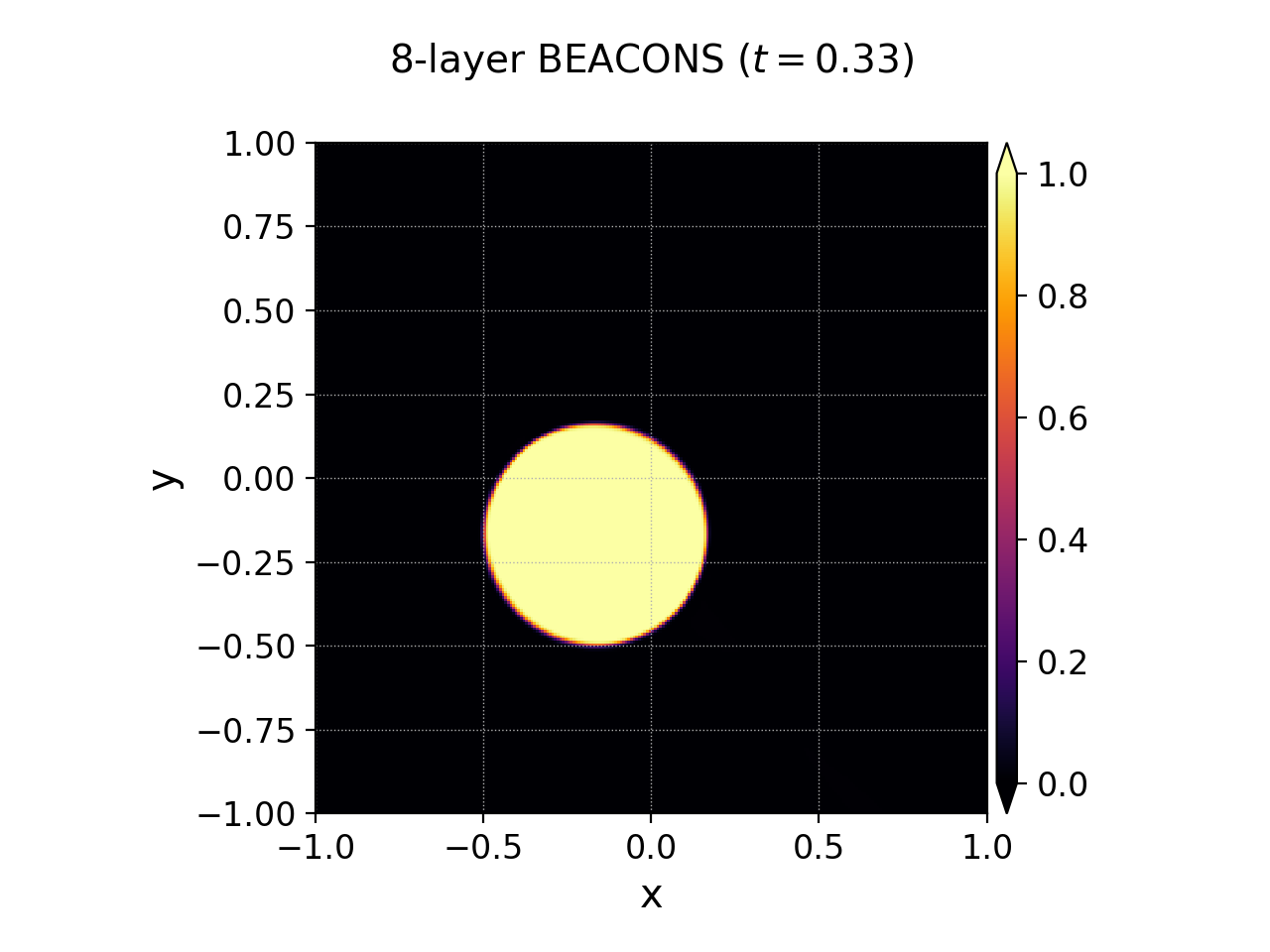}
\includegraphics[trim={1.5cm, 0cm, 1.5cm, 0cm}, clip, width=0.33\textwidth]{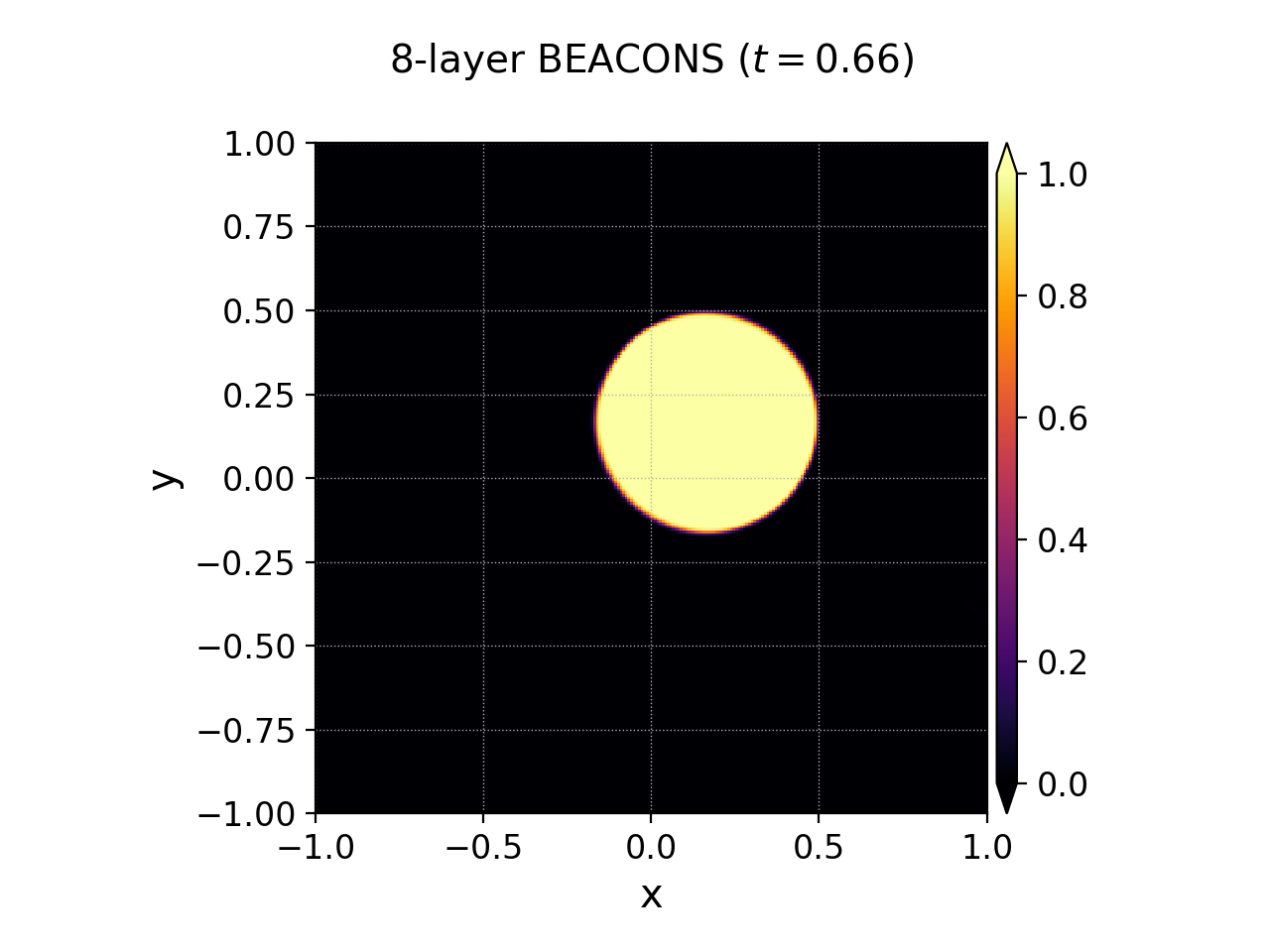}
\includegraphics[trim={1.5cm, 0cm, 1.5cm, 0cm}, clip, width=0.33\textwidth]{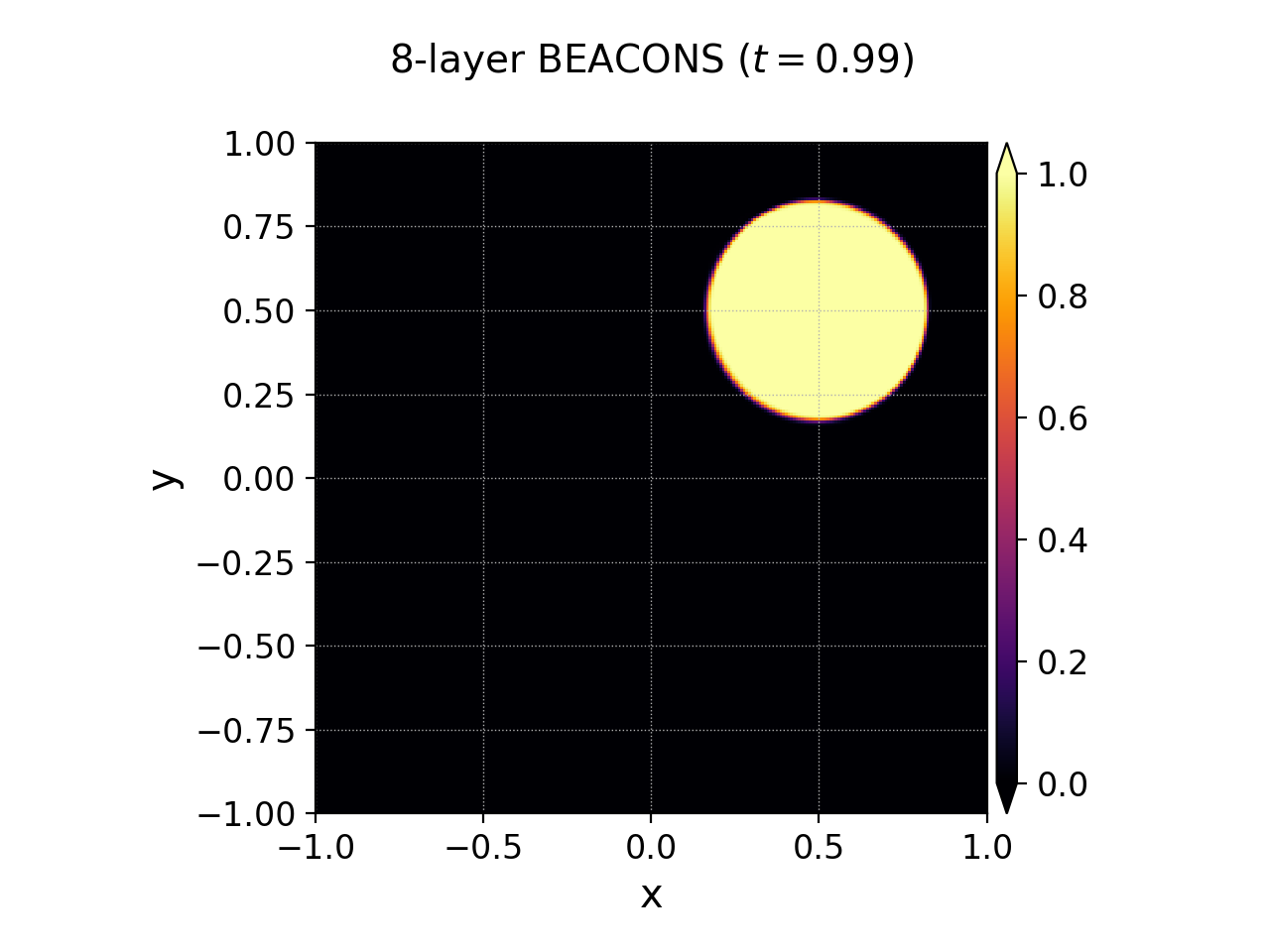}
\caption{Results for the 2D linear advection disk problem at times ${t = 0.33}$ (left), ${t = 0.66}$ (middle), and ${t = 0.99}$ (right), obtained using a formally-verified numerical solver (top), an 8-layer neural network (middle), and an 8-layer BEACONS architecture (bottom). The 8-layer neural network progressively distorts the shape of the disk as it approaches the top right of the domain, causing it to become increasingly ``egg-shaped'', while also underestimating the advection speed overall. The BEACONS architecture successfully preserves the shape of the disk, and shows near-perfect agreement with the numerical solution.}
\label{fig:Figure2}
\end{figure}

The second test problem will consider an extension of the linear advection equation to two dimensions (with the same advection speed ${a \in \mathbb{R}}$ in both directions):

\begin{equation}
\frac{\partial u}{\partial t} + \frac{\partial \left( a u \right)}{\partial x} + \frac{\partial \left( a u \right)}{\partial y} = 0,
\end{equation}
with a square domain ${\left[ -1.0, 1.0 \right] \times \left[ -1.0, 1.0 \right]}$. Our automated theorem-proving framework is again able to produce full proofs of correctness for both the Lax-Friedrichs and Roe-type finite volume solvers for this equation (including its second-order dimension splitting); hyperbolicity-preservation, CFL stability, and local Lipschitz continuity of the Lax-Friedrichs solver require a total of 65, 89, and 77 proof steps to establish, respectively, while hyperbolicity-preservation and flux conservation of the Roe-type solver require a total of 109 and 183 proof steps to establish, respectively. We advect a disk described by initial data:

\begin{equation}
u_0 \left( x, y \right) = \begin{cases}
1.0, \qquad & \text{ for }  \left( x + 0.5 \right)^2 + \left( y + 0.5 \right)^2 \leq 0.33,\\
0.0 \qquad & \text{ for } \left( x + 0.5 \right)^2 + \left( y + 0.5 \right)^2 > 0.33,
\end{cases}
\end{equation}
from the bottom left to the top right of the domain, with advection speed ${a = 1.0}$. Once again, we solve this problem using the formally-verified \textit{Roe-type} solver, a CFL coefficient ${C_{\text{CFL}} = 1.0}$, and now a spatial resolution of ${256 \times 256}$ cells. We evolve the problem numerically up until a final time of ${t = 1.0}$, again producing 100 frames of output, and now train only two different neural network architectures on the first 33 frames (up until time ${t = 0.33}$), and again ask both of them to predict the remainder of the simulation based on the learned solution function ${u \left( t, x, y \right)}$. The chosen neural network architectures are an 8-layer BEACONS architecture with 128 neurons per layer, and, for means of comparison, an 8-layer fully-connected (non-BEACONS) neural network with 128 neurons per layer. For this problem, our automated theorem-proving framework proves worst-case ${L^{\infty}}$ errors of 1.216729 (for the smooth parts of the solution) and 1.483672 (for the non-smooth parts of the solution) for a 6-layer BEACONS architecture, and 1.0 (for the smooth parts of the solution) and 1.259921 (for the non-smooth parts of the solution) for the 8-layer BEACONS architecture used here. In each case, the theorem-prover requires a total of 187 proof steps to establish the bound. The (inferred) solutions from the formally-verified numerical solver, the 8-layer neural network, and the 8-layer BEACONS architecture, at times ${t = 0.33}$, ${t = 0.66}$, and ${t = 0.99}$, are shown in Figure \ref{fig:Figure2}. We see that the 8-layer neural network solution fails to preserve the shape of the disk, which gets progressively more distorted and ``egg-shaped'' as it approaches the top right of the domain. Tracking the center of this distorted disk indicates that the 8-layer neural network solution also underestimates the advection speed on average. On the other hand, the 8-layer BEACONS solution fully preserves the shape of the disk, and the advection speed is again estimated correctly. Indeed, the 8-layer BEACONS solution is in almost perfect agreement with the numerical solution, albeit with some very slight numerical diffusion around the boundaries of the disk. Table \ref{tab:Table3} shows the normalized ${L^2}$ and ${L^{\infty}}$ errors (as compared against the formally-verified numerical solution) across the two different neural network architectures, both for the final frame only, and across all frames (using a per-frame average for the ${L^2}$ error, and normalizing across all frames for the ${L^{\infty}}$ error). We see overall that both the final-frame and all-frame ${L^2}$ and ${L^{\infty}}$ errors are significantly lower for the BEACONS architecture than for the non-BEACONS architecture. Note again that the largest ${L^{\infty}}$ error observed for the BEACONS architecture (i.e. 0.938123) remains comfortably below the worst-case bound (i.e. 1.259921).

\begin{table}[ht]
\centering
\begin{tabular}{|c||c|c|c|c|}
\hline
Architecture & ${L^{\infty}}$ Error (Final) & ${L^2}$ Error (Final) & ${L^{\infty}}$ Error (All) & ${L^2}$ Error (All)\\
\hline\hline
8-layer NN & 1.031098 & 40.481536 & 1.031098 & 22.772318\\
\hline
8-layer BEACONS & 0.864784 & 9.363477 & 0.938123 & 6.747000\\
\hline
\end{tabular}
\caption{${L^2}$ and ${L^{\infty}}$ error analysis for the 2D linear advection disk problem, comparing the 8-layer neural network and 8-layer BEACONS solutions against the formally-verified numerical solution, both for the final predicted frame, and across all predicted frames.}
\label{tab:Table3}
\end{table}

\subsection{Inviscid Burgers' Equation}
\label{sec:Section8}

\begin{figure}[ht]
\centering
\includegraphics[width=0.495\textwidth]{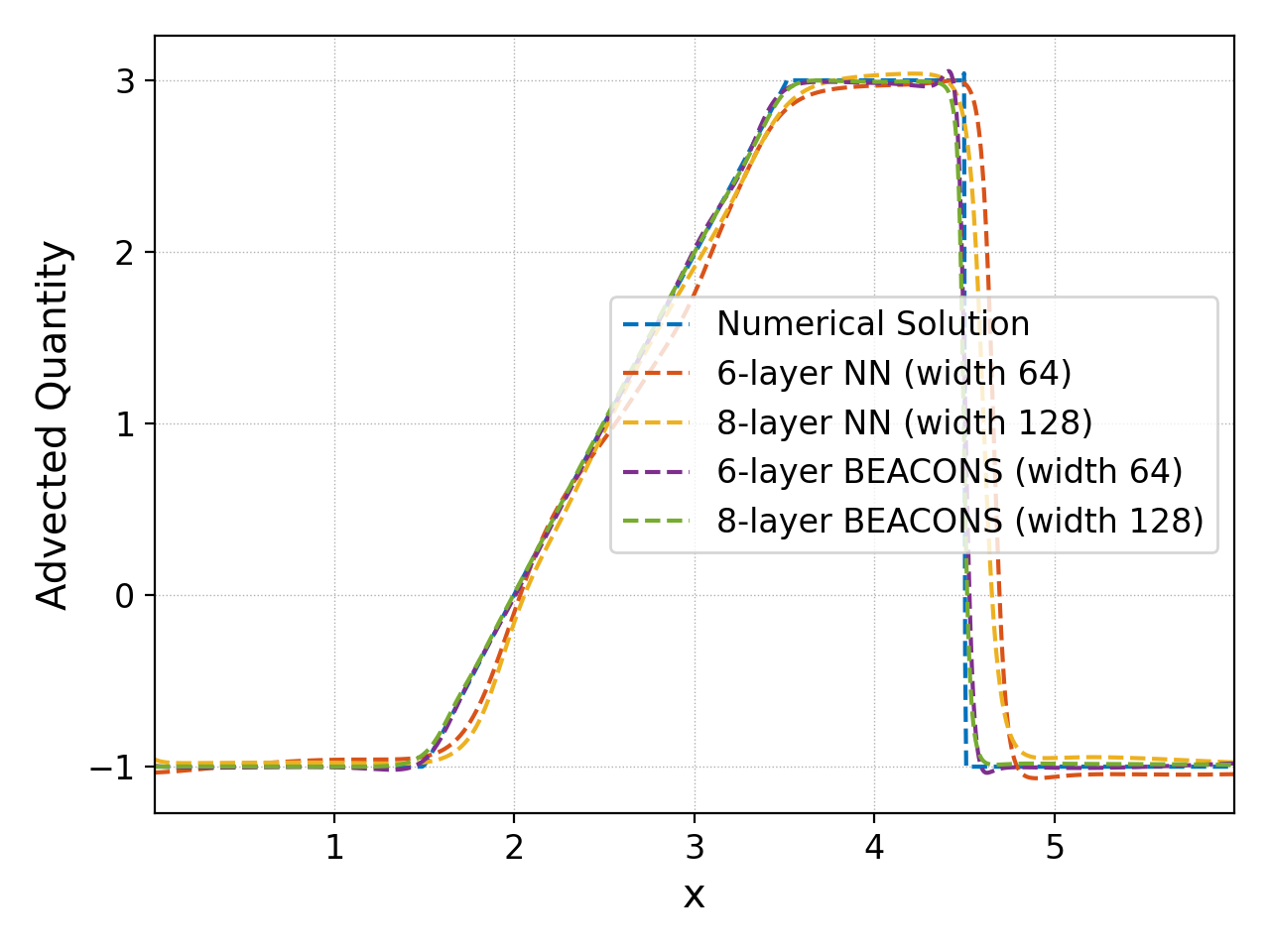}
\includegraphics[width=0.495\textwidth]{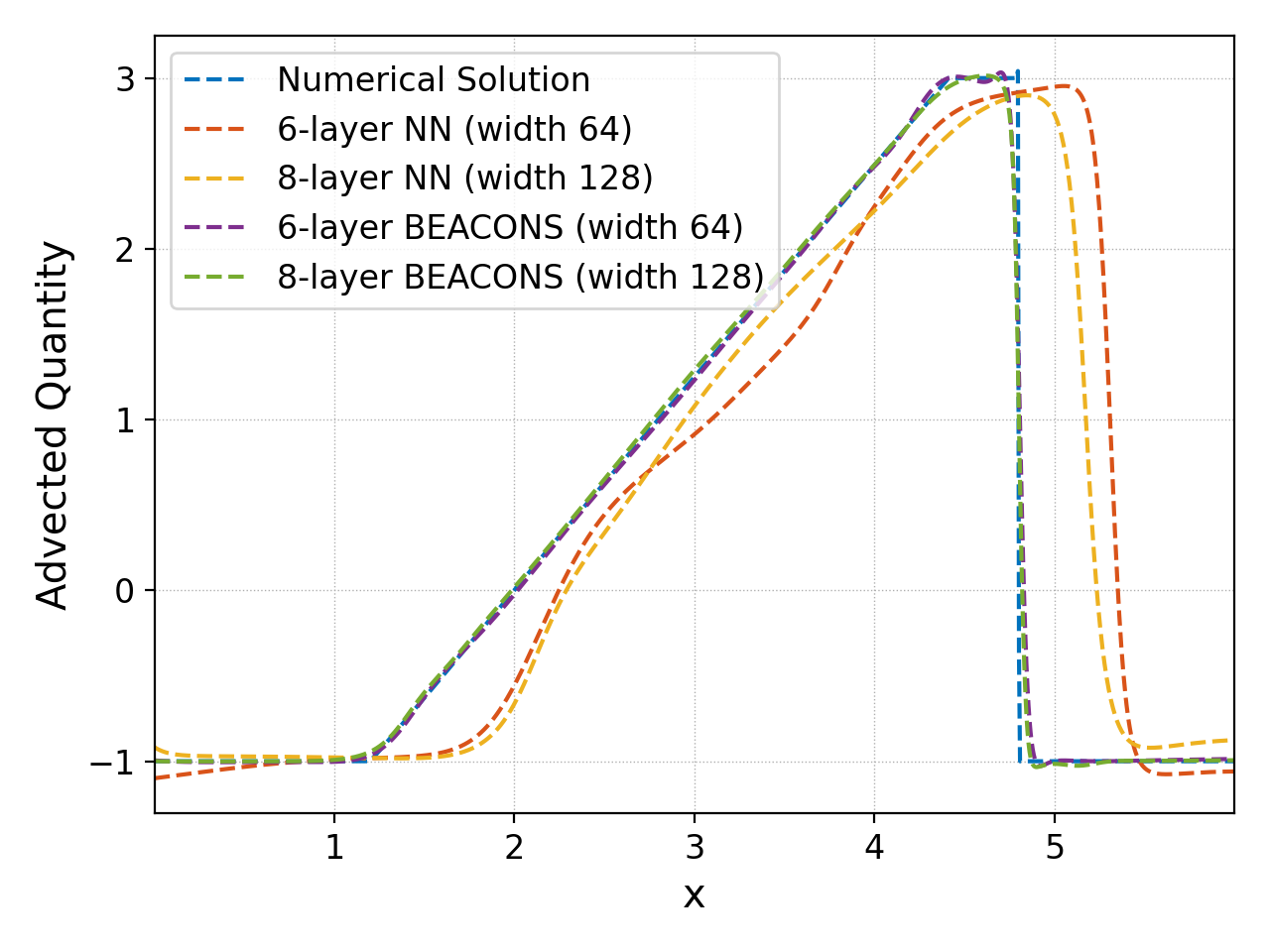}
\caption{Results for the 1D inviscid Burgers' ``top-hat'' initial value problem at times ${t = 0.5}$ (left) and ${t = 0.8}$ (right), obtained using a formally-verified numerical solver (blue), a 6-layer neural network (red), an 8-layer neural network (yellow), a 6-layer BEACONS architecture (purple), and an 8-layer BEACONS architecture (green). Both the 6-layer and 8-layer neural networks significantly overestimate the speed of the right-moving shock and underestimate the speed of the left-moving rarefaction; both neural network solutions fail to conserve the advected quantity, and exhibit significant diffusion of the overall qualitative structure of the solution. Both BEACONS solutions track the numerical solution more-or-less perfectly, without significant instability, wave-speed mis-prediction, or conservation errors.}
\label{fig:Figure3}
\end{figure}

Our second numerical test case will be to solve the scalar \textit{inviscid Burgers'} equation:

\begin{equation}
\frac{\partial u}{\partial t} + u \left( \frac{\partial u}{\partial x} \right) = \frac{\partial u}{\partial t} + \frac{\partial \left( \frac{1}{2} u^2 \right)}{\partial x} = 0,
\end{equation}
i.e. a scalar conservation equation with non-linear flux ${f \left( u \right) = \frac{1}{2} u^2}$. Due to the non-linear nature of the flux, Burgers' equation is able to exhibit \textit{shocks} (i.e. weak/distribution solutions), as well as \textit{rarefaction waves}. Our automated theorem-proving framework is able to produce full proofs of correctness for both the Lax-Friedrichs and Roe-type finite volume solvers for this equation; hyperbolicity-preservation, CFL stability, and local Lipscitz continuity of the Lax-Friedrichs solver require a total of 95, 121, and 101 proof steeps to establish, respectively, while hyperbolicity-preservation and flux conservation of the Roe-type solver require a total of 125 and 193 proof steps to establish, respectively. The first test problem will be a one-dimensional ``top-hat'' initial value problem with two initial discontinuities, defined over the spatial domain ${\left[ 0.0, 6.0 \right]}$, with initial data given by:

\begin{equation}
u_0 \left( x \right) = \begin{cases}
3.0, \qquad & \text{ for } 2.0 \leq x \leq 4.0,\\
-1.0 \qquad & \text{ for } x < 2.0 \text{ or } x > 4.0.
\end{cases}
\end{equation}
As previously, we shall solve this problem numerically using the formally-verified \textit{Roe-type} approximate Riemann solver (no longer equivalent to an exact Riemann solver, due to the non-linearity of the problem), with a CFL coefficient ${C_{\text{CFL}} = 1.0}$, and with a spatial resolution of 1024 cells. We evolve the problem numerically up until a final time of ${t = 1.0}$, producing 100 frames of output in the process. As before, we train four different neural network architectures on the first 33 frames of this output (up until time ${t = 0.33}$), and then ask each of them to predict the remainder of the simulation based on the learned solution function ${u \left( t, x \right)}$. The chosen neural network architectures, as before, are: a 6-layer BEACONS architecture with 64 neurons per layer, an 8-layer BEACONS architecture with 128 neurons per layer, and, for means of comparison, a 6-layer fully-connected (non-BEACONS) neural network with 64 neurons per layer, and an 8-layer fully-connected (non-BEACONS) neural network with 128 neurons per layer. For this problem, our automated theorem-proving framework proves worst-case ${L^{\infty}}$ errors of 1.216728 (for the smooth parts of the solution) and 1.483672 (for the non-smooth parts of the solution) for the 6-layer BEACONS architecture, and 1.0 (for the smooth parts of the solution) and 1.259921 (for the non-smooth parts of the solution) for the 8-layer BEACONS architecture\footnote{It is not a coincidence that these bounds are identical to the bounds proved previously for the 2D linear advection disk problem: it turns out that the relevant smoothness properties of the solution are identical between the two tests.}. In each case, the theorem-prover requires a total of 163 proof steps to establish the bound. The (inferred) solutions from the formally-verified numerical solver, the 6-layer neural network, the 8-layer neural network, the 6-layer BEACONS architecture, and the 8-layer BEACONS architecture, at times ${t = 0.5}$ and ${t = 0.8}$, are shown in Figure \ref{fig:Figure3}.

We see that the 6-layer neural network significantly overestimates the speed of the right-moving shock, slightly underestimates the speed of the left-moving rarefaction, and exhibits large oscillations/instabilities around the rarefaction wave. The overestimation of the right-moving shock speed is slightly reduced in the 8-layer neural network solution, but only marginally. The 8-layer neural network solution is somewhat more stable overall, but the mis-prediction of the rarefaction wave-speed is still present, and both neural network solutions fail to conserve the advected quantity and exhibit severe diffusion of the qualitative structure of the solution. Both of the BEACONS solutions remain very close to the numerical solution, with both the right-moving shock speed and the left-moving rarefaction speed estimated more-or-less exactly. The 6-layer BEACONS solution exhibits some small overshoots and undershoots in the region surrounding the right-moving shock wave, which are noticeably reduced in the 8-layer BEACONS solution, which appears slightly more diffusive in some respects but also marginally more stable overall. Both BEACONS solutions appear to conserve the advected quantity at least approximately, and do not diffuse the qualitative structure of the solution to any significant extent. Table \ref{tab:Table4} shows the normalized ${L^2}$ and ${L^{\infty}}$ errors (as compared against the formally-verified numerical solution) across the four different neural network architectures, both for the final frame only, and across all frames (using a per-frame average for the ${L^2}$ error, and normalizing across all frames for the ${L^{\infty}}$ error). We see overall that both the final-frame and all-frame ${L^2}$ and ${L^{\infty}}$ errors are significantly lower for the two BEACONS architectures than for the non-BEACONS architectures. In the non-BEACONS case, we see that increasing the number of layers has the effect of marginally decreasing the ${L^2}$ and ${L^{\infty}}$ errors, while in the BEACONS case we see the opposite effect. Table \ref{tab:Table5} shows the normalized conservation errors (as obtained by integrating over the conserved quantity, and comparing against the integral of the formally-verified numerical solution) across the four different neural network architectures, again both for the final frame only, and integrated across all frames. We see overall that both the final-frame and all-frame conservation errors are significantly lower for the two BEACONS architectures than for the non-BEACONS architectures. Again, in the non-BEACONS case, we see that increasing the number of layers has the effect of significantly decreasing both the final-frame and all-frame conservation errors, while in the BEACONS case, we observe the opposite phenomenon. These findings, wherein the performance of the 8-layer BEACONS architecture appears strictly worse than that of the 6-layer BEACONS architecture, stand in contrast to our previous findings for the linear advection equation, and suggest that the 6-layer BEACONS architecture may be in some sense optimal (or near-optimal) for solving this particular class of non-linear initial value problem. Note that the largest ${L^{\infty}}$ errors observed for the two BEACONS architectures (i.e. 1.028023 and 0.986979, respectively) still nevertheless remain comfortably below the worst-case bounds (i.e. 1.483672 and 1.259921, respectively).

\begin{table}[ht]
\centering
\begin{tabular}{|c||c|c|c|c|}
\hline
Architecture & ${L^{\infty}}$ Error (Final) & ${L^2}$ Error (Final) & ${L^{\infty}}$ Error (All) & ${L^2}$ Error (All)\\
\hline\hline
6-layer NN & 1.310069 & 15.058654 & 1.311499 & 8.144540\\
\hline
8-layer NN & 1.208962 & 12.127816 & 1.307407 & 6.691768\\
\hline
6-layer BEACONS & 0.986979 & 2.165441 & 0.973603 & 1.554421\\
\hline
8-layer BEACONS & 1.028023 & 2.404486 & 1.014091 & 1.636904\\
\hline
\end{tabular}
\caption{${L^2}$ and ${L^{\infty}}$ error analysis for the 1D inviscid Burgers' ``top-hat'' initial value problem, comparing the 6-layer neural network, 8-layer neural network, 6-layer BEACONS, and 8-layer BEACONS solutions against the formally-verified numerical solution, both for the final predicted frame, and across all frames.}
\label{tab:Table4}
\end{table}

\begin{table}[ht]
\centering
\begin{tabular}{|c||c|c|}
\hline
Architecture & Conservation Error (Final) & Conservation Error (Total)\\
\hline\hline
6-layer NN & 69.348586 & 2254.019031\\
\hline
8-layer NN & 30.707755 & 1442.490291\\
\hline
6-layer BEACONS & 4.459920 & -0.569301\\
\hline
8-layer BEACONS & 12.788327 & -35.540890\\
\hline
\end{tabular}
\caption{Conservation analysis for the 1D inviscid Burgers' ``top-hat'' initial value problem, comparing the 6-layer neural network, 8-layer neural network, 6-layer BEACONS, and 8-layer BEACONS solutions against the formally-verified numerical solution, both for the final predicted frame, and across all frames.}
\label{tab:Table5}
\end{table}

\begin{figure}[ht]
\centering
\includegraphics[trim={1.5cm, 0cm, 1.5cm, 0cm}, clip, width=0.33\textwidth]{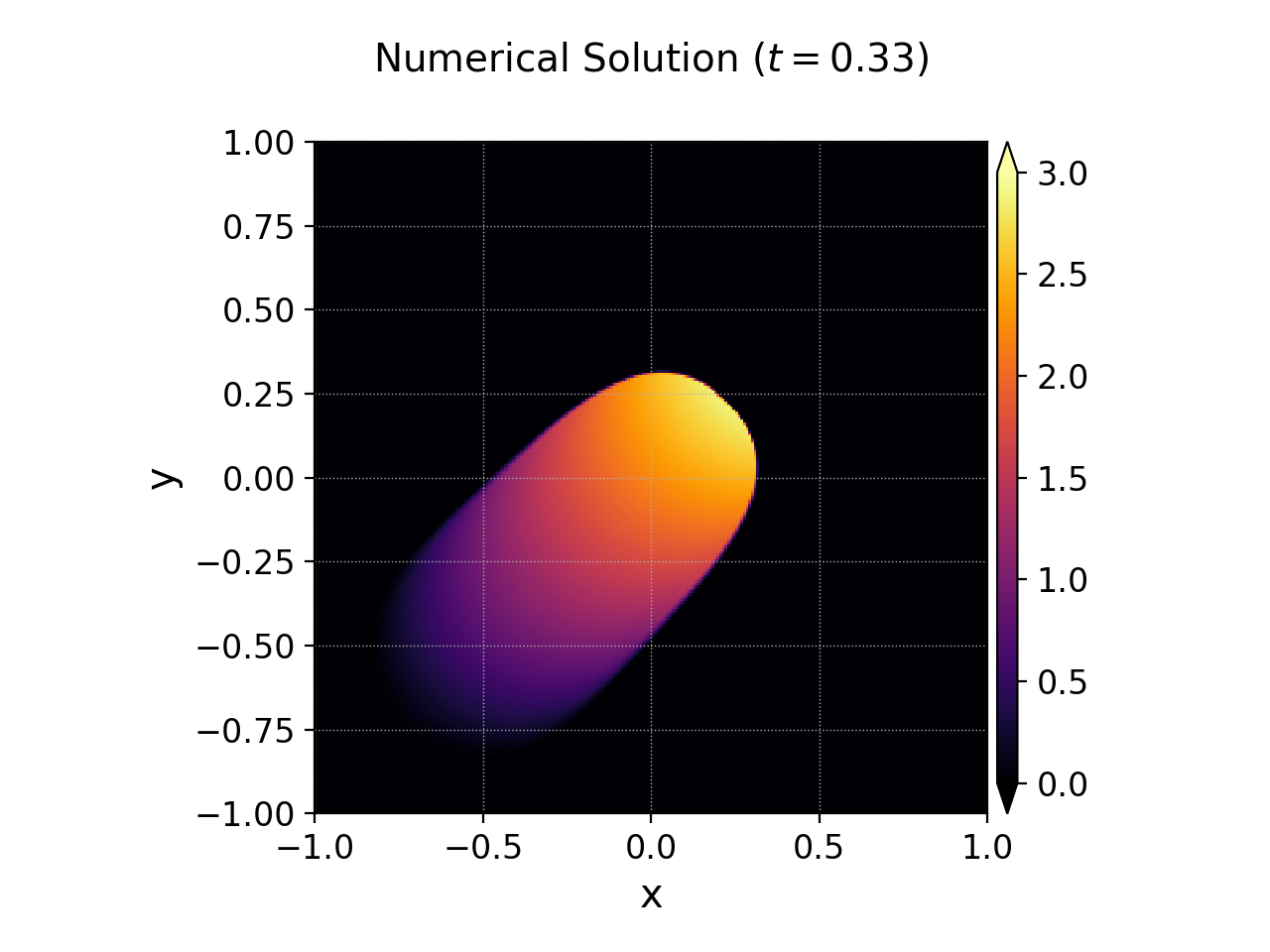}
\includegraphics[trim={1.5cm, 0cm, 1.5cm, 0cm}, clip, width=0.33\textwidth]{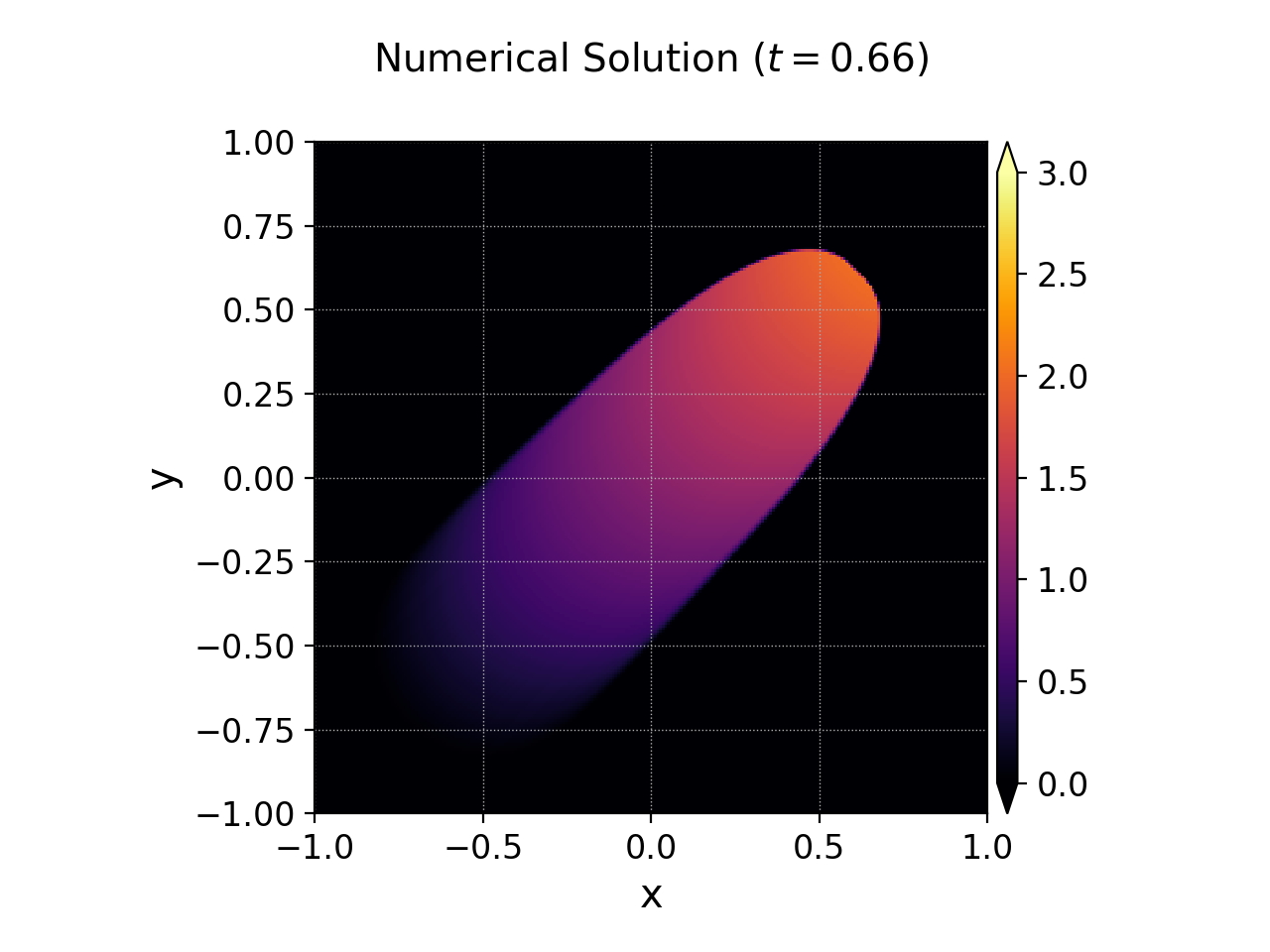}
\includegraphics[trim={1.5cm, 0cm, 1.5cm, 0cm}, clip, width=0.33\textwidth]{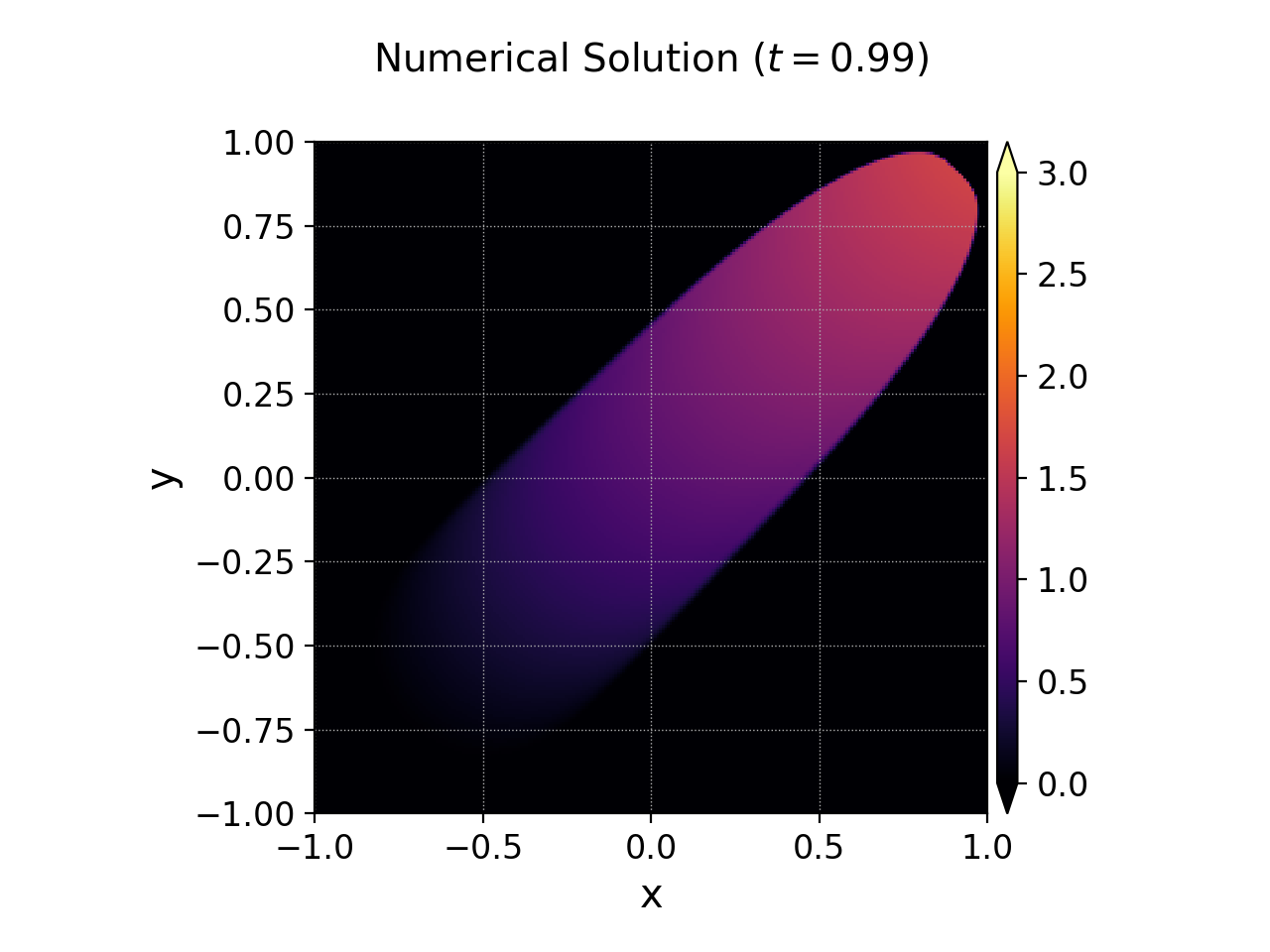}
\includegraphics[trim={1.5cm, 0cm, 1.5cm, 0cm}, clip, width=0.33\textwidth]{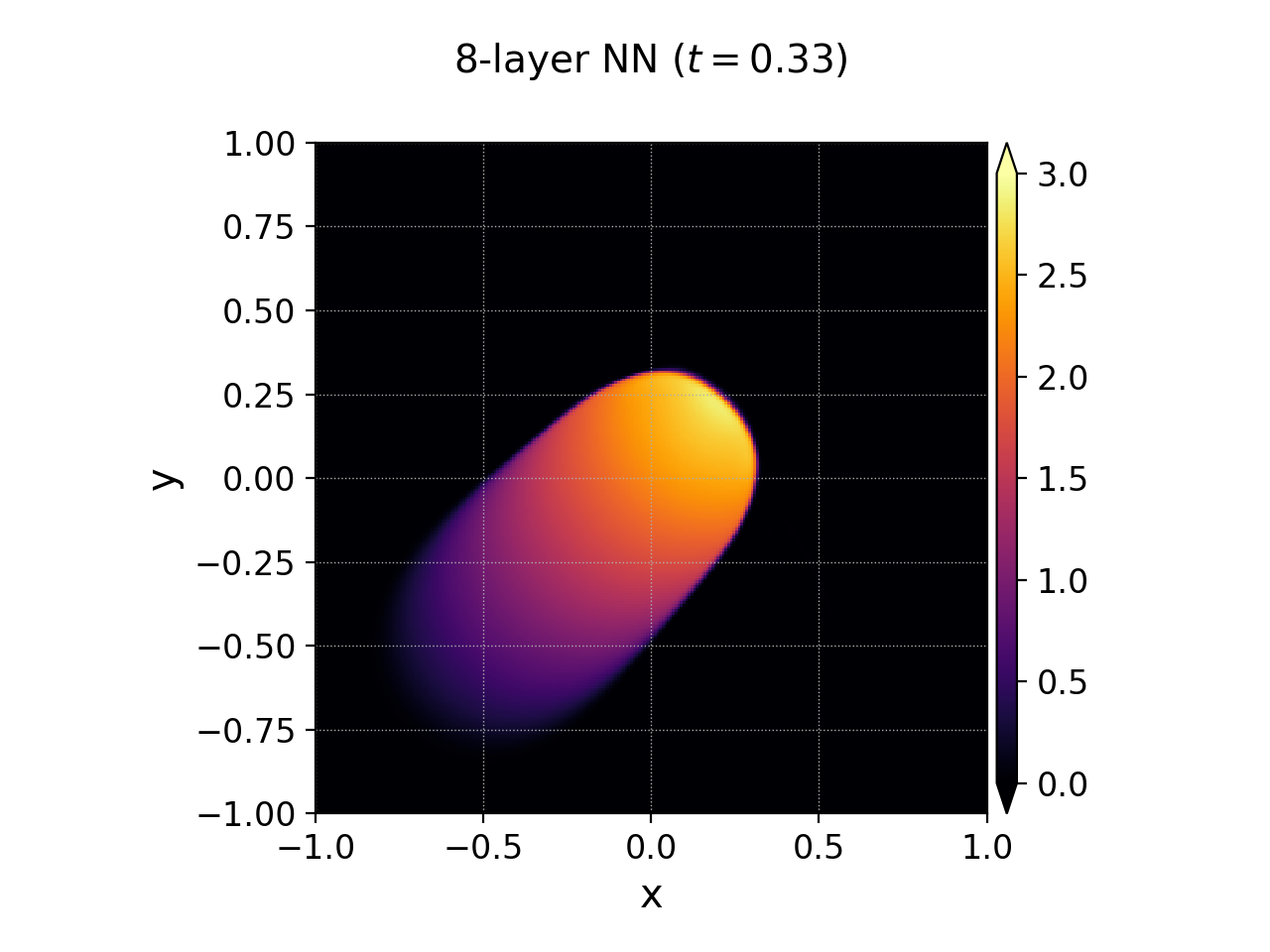}
\includegraphics[trim={1.5cm, 0cm, 1.5cm, 0cm}, clip, width=0.33\textwidth]{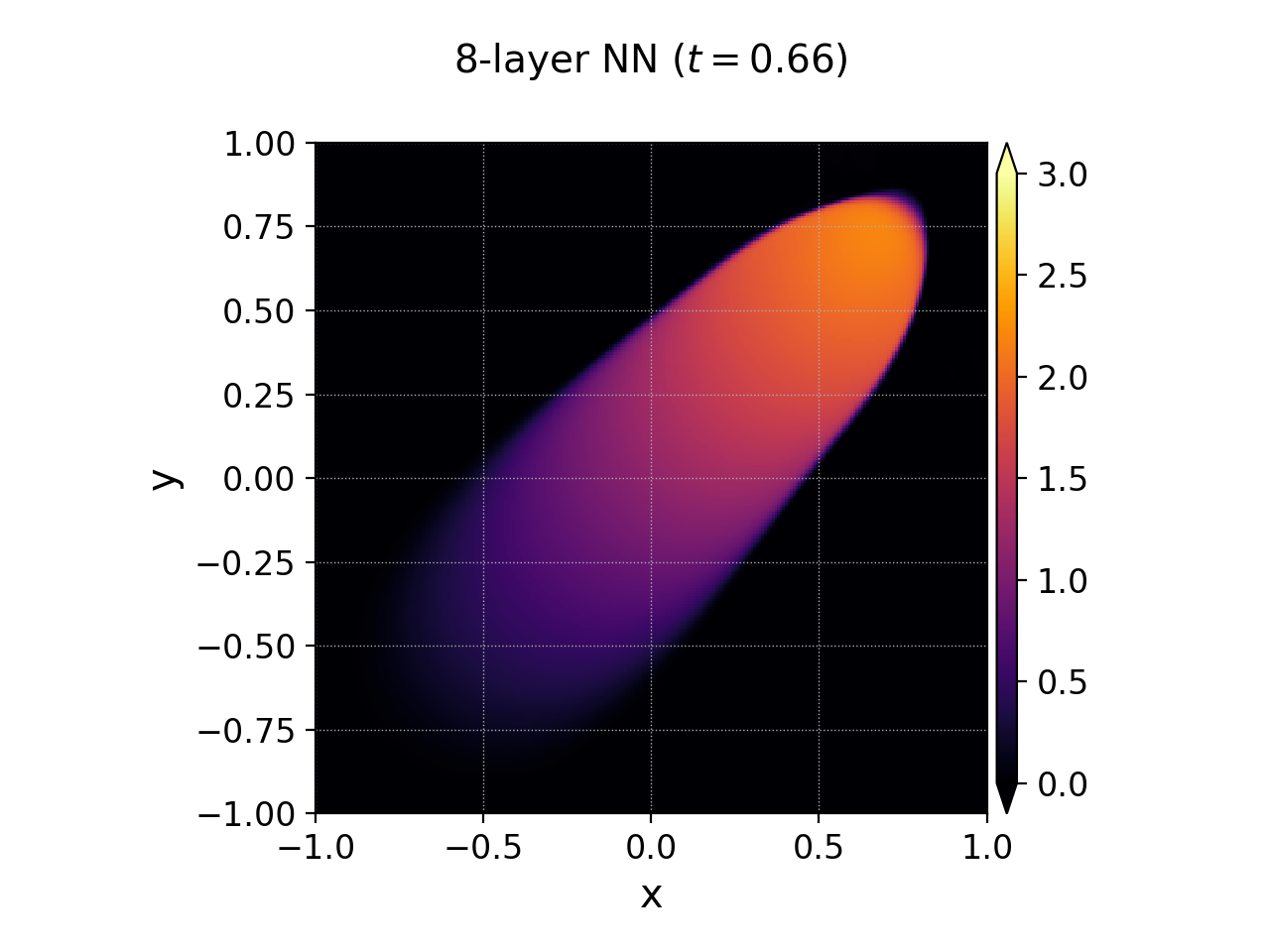}
\includegraphics[trim={1.5cm, 0cm, 1.5cm, 0cm}, clip, width=0.33\textwidth]{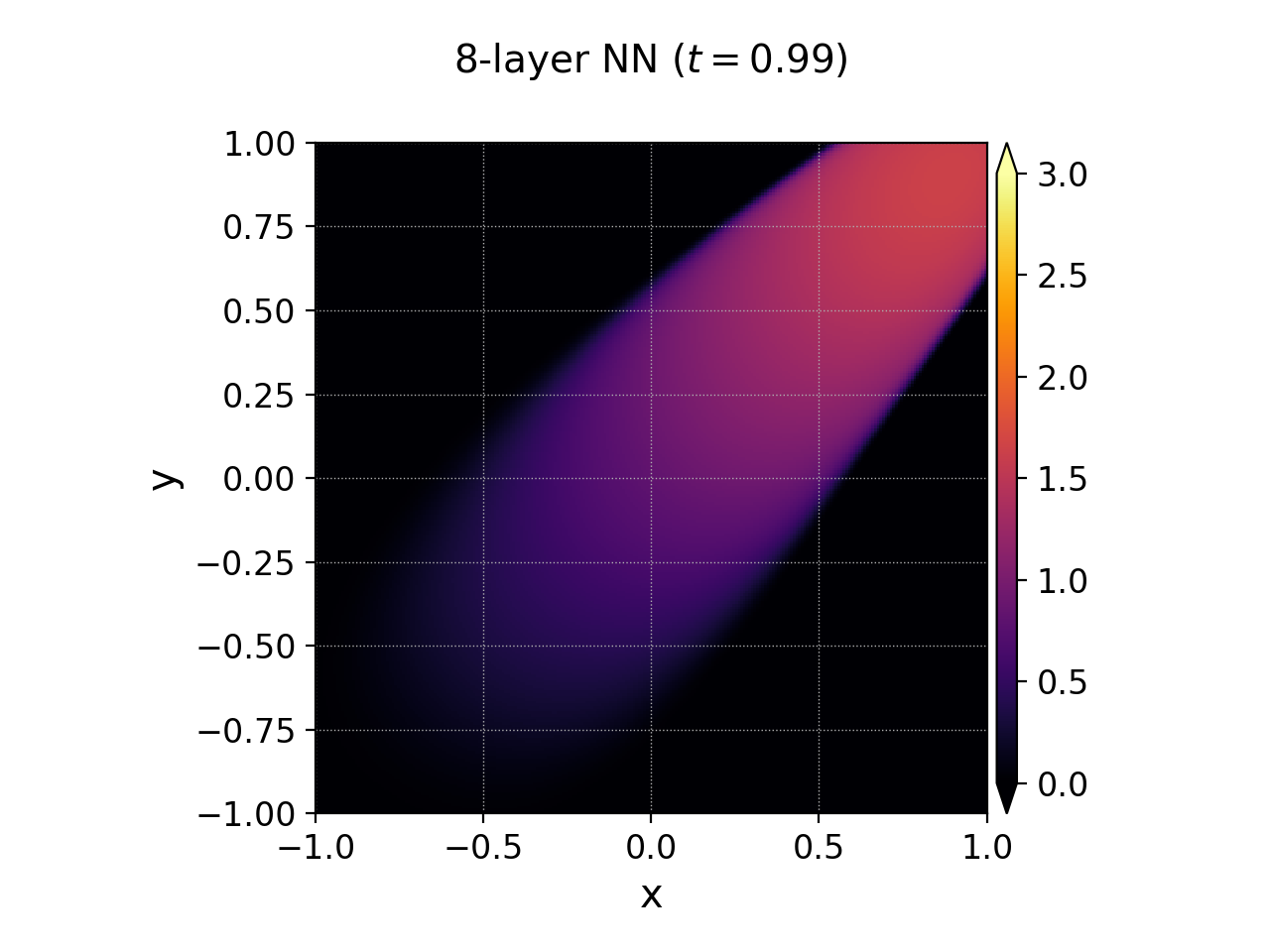}
\includegraphics[trim={1.5cm, 0cm, 1.5cm, 0cm}, clip, width=0.33\textwidth]{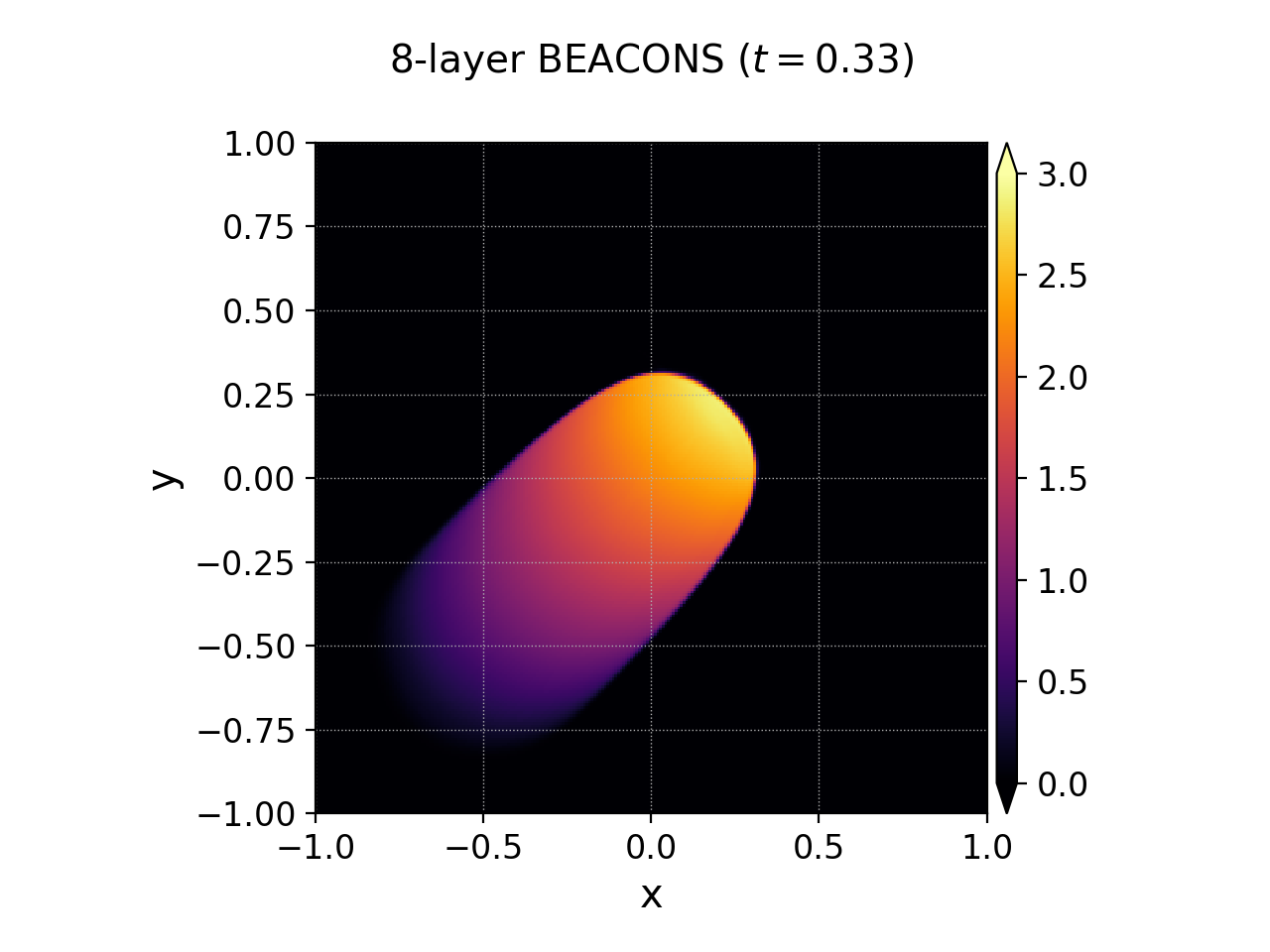}
\includegraphics[trim={1.5cm, 0cm, 1.5cm, 0cm}, clip, width=0.33\textwidth]{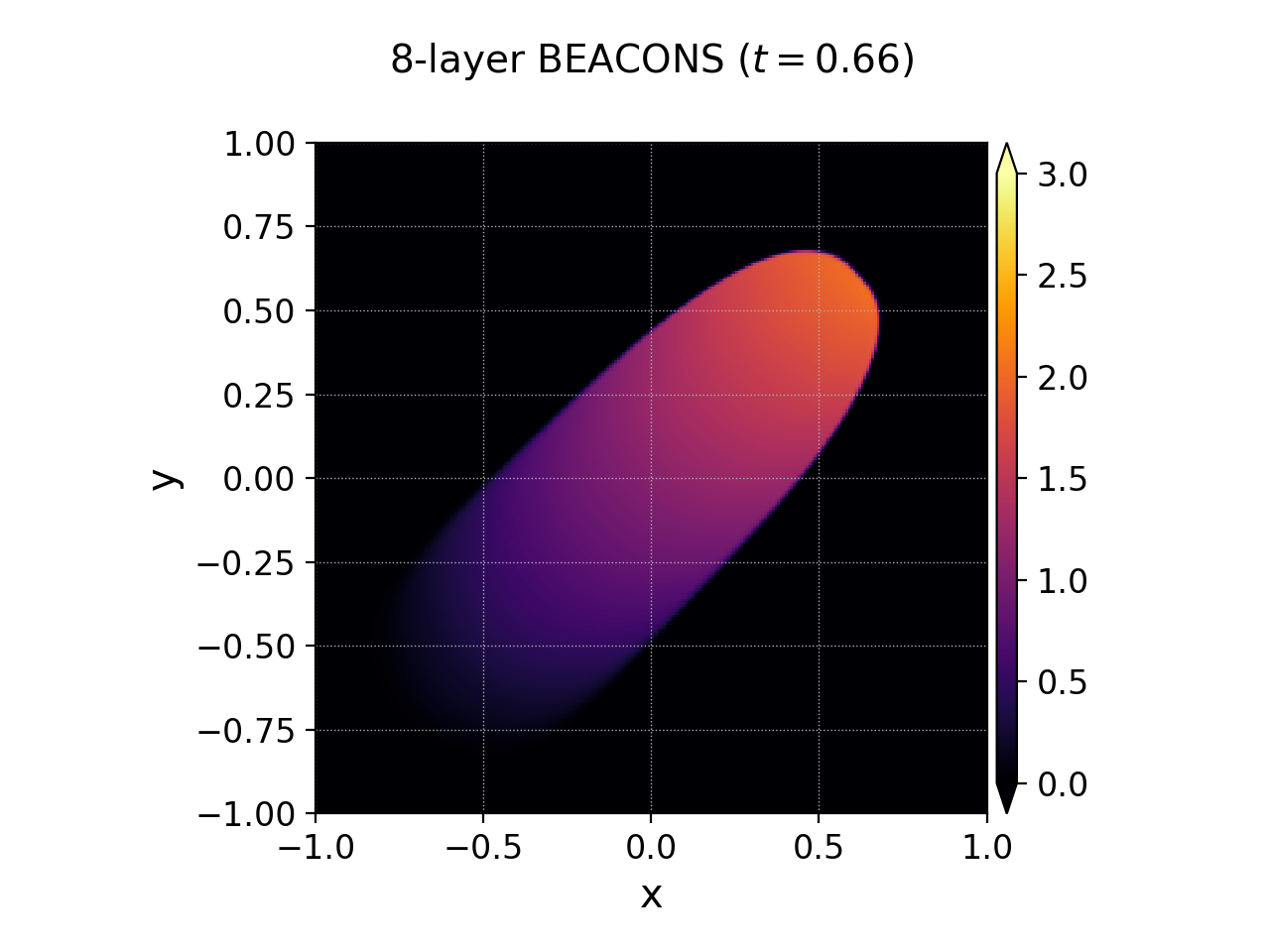}
\includegraphics[trim={1.5cm, 0cm, 1.5cm, 0cm}, clip, width=0.33\textwidth]{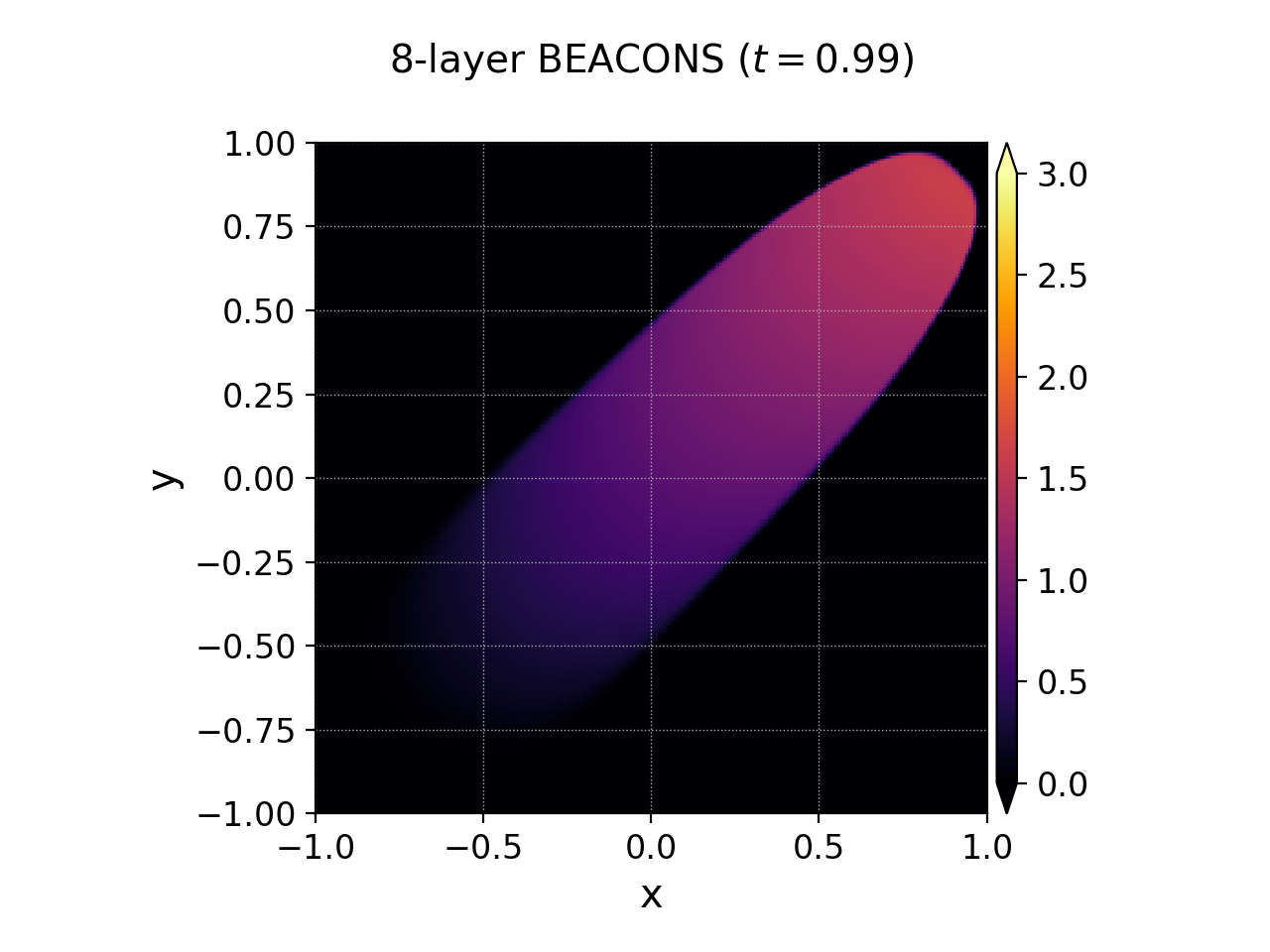}
\caption{Results for the 2D inviscid Burgers' disk problem at times ${t = 0.33}$ (left), ${t = 0.66}$ (middle), and ${t = 0.99}$ (right), obtained using a formally-verified numerical solver (top), an 8-layer neural network (middle), and an 8-layer BEACONS architecture (bottom). The 8-layer neural network mis-predicts the shape of the solution, causing it to appear much more ``pointy'' than it is, while also overestimating the propagation speed significantly, causing the advected quantity to hit the boundary of the domain before the end of the simulation. The BEACONS architecture successfully predicts both the correct shape and the correct propagation speed of the solution, matching the numerical solution more-or-less perfectly.}
\label{fig:Figure4}
\end{figure}

The second test problem will consider an extension of the inviscid Burgers' equation to two dimensions (with identical fluxes in both directions):

\begin{equation}
\frac{\partial u}{\partial t} + u \left( \frac{\partial u}{\partial x} \right) + u \left( \frac{\partial u}{\partial y} \right) = \frac{\partial u}{\partial t} + \frac{\partial \left( \frac{1}{2} u^2 \right)}{\partial x} + \frac{\partial \left( \frac{1}{2} u^2 \right)}{\partial y} = 0,
\end{equation}
with a square domain ${\left[ -1.0, 1.0 \right] \times \left[ -1.0, 1.0 \right]}$. Our automated theorem-proving framework is again able to produce full proofs of correctness for both the Lax-Friedrichs and Roe-type finite volume solvers for this equation (including its second-order dimension splitting); hyperbolicity-preservation, CFL stability, and local Lipschitz continuity of the Lax-Friedrichs solver require a total of 179, 231, and 191 proof steps to establish, respectively, while hyperbolicity-preservation and flux conservation of the Roe-type solver require a total of 239 and 375 proof steps to establish, respectively. We evolve a disk described by initial data (equivalent to the previous 2D linear advection case):

\begin{equation}
u_0 \left( x, y \right) = \begin{cases}
1.0, \qquad & \text{ for } \left( x + 0.5 \right)^2 + \left( y + 0.5 \right)^2 \leq 0.33,\\
0.0, \qquad & \text{ for } \left( x + 0.5 \right)^2 + \left( y + 0.5 \right)^2 > 0.33,
\end{cases}
\end{equation}
starting from the bottom left of the domain, and then smearing out in a ``comet'' pattern as it moves towards the top right. Once again, we solve this problem using the formally-verified \textit{Roe-type} solver, a CFL coefficient ${C_{\text{CFL}} = 1.0}$, and a spatial resolution of ${256 \times 256}$ cells. We evolve the problem numerically up until a final time of ${t = 1.0}$ again, producing 100 frames of output, and now train only two different neural network architectures on the first 33 frames (up until time ${t = 0.33}$), and again ask both of them to predict the remainder of the simulation based on the learned solution function ${u \left( t, x, y \right)}$. The chosen neural network architectures are an 8-layer BEACONS architecture with 128 neurons per layer, and, for means of comparison, an 8-layer fully-connected (non-BEACONS) neural network with 128 neurons per layer. For this problem, our automated theorem-proving framework proves a worst-case ${L^{\infty}}$ error of 1.483672 for the smooth parts of the solution, but is unable to prove any error bound on the non-smooth parts of the solution\footnote{The reason for this is that the theorem-prover concludes, correctly, that in the infinite-time limit of this 2D inviscid Burgers' disk problem, the shock discontinuity eventually damps to zero, and the solution becomes asymptotically smooth everywhere. Therefore it is unable to prove an error bound for non-smooth parts of the solution, because no non-smooth parts exist asymptotically.}, for a 6-layer BEACONS architecture. Likewise, the framework proves a worst-case ${L^{\infty}}$ error of 1.259921 for the smooth parts of the solution, but is unable to prove any error bound on the non-smooth parts of the solution, for the 8-layer BEACONS architecture used here. For both architectures, the theorem-prover requires a total of 315 proof steps to establish a bound in the smooth case, and completes 221 proof steps before concluding that no bound can be proven (using the available methods) in the non-smooth case. The (inferred) solutions from the formally-verified numerical solver, the 8-layer neural network, and the 8-layer BEACONS architecture, at times ${t = 0.33}$, ${t = 0.66}$, and ${t = 0.99}$, are shown in Figure \ref{fig:Figure4}. We see that the 8-layer neural network solution fails to predict the correct ``comet'' shape of the solution, causing it to appear ``pointy'' rather than rounded. The 8-layer neural network solution also overestimates the propagation speed of the advected quantity substantially, causing the ``comet'' to hit the boundary and exit the domain prematurely, prior to the end of the simulation. On the other hand, the 8-layer BEACONS solution correctly predicts both the correct qualitative shape and the correct quantitative propagation speed of the ``comet'' structure, exhibiting only very slight numerical diffusion surrounding the propagation front as compared to the numerical solution. Table \ref{tab:Table6} shows the normalized ${L^2}$ and ${L^{\infty}}$ errors (as compared against the formally-verified numerical solution) across the two different neural network architectures, both for the final frame only, and across all frames (using a per-frame average for the ${L^2}$ error, and normalizing across all frames for the ${L^{\infty}}$ error). We see overall that both the final-frame and all-frame ${L^2}$ and ${L^{\infty}}$ errors are significantly lower for the BEACONS architecture than for the non-BEACONS architecture. Note again that the largest ${L^{\infty}}$ error observed for the BEACONS architecture (i.e. 0.595719) remains comfortably below the \textit{provable} worst-case bound (i.e. 1.259921).

\begin{table}[ht]
\centering
\begin{tabular}{|c||c|c|c|c|}
\hline
Architecture & ${L^{\infty}}$ Error (Final) & ${L^2}$ Error (Final) & ${L^{\infty}}$ Error (All) & ${L^2}$ Error (All)\\
\hline\hline
8-layer NN & 0.976685 & 37.415877 & 0.653738 & 17.396335\\
\hline
8-layer BEACONS & 0.319521 & 2.688830 & 0.595719 & 1.888049\\
\hline
\end{tabular}
\caption{${L^2}$ and ${L^{\infty}}$ error analysis for the 2D inviscid Burgers' disk problem, comparing the 8-layer neural network and 8-layer BEACONS solutions against the formally-verified numerical solution, both for the final predicted frame, and across all predicted frames.}
\label{tab:Table6}
\end{table}

\subsection{Compressible Euler Equations}
\label{sec:Section9}

\begin{figure}[ht]
\centering
\includegraphics[width=0.495\textwidth]{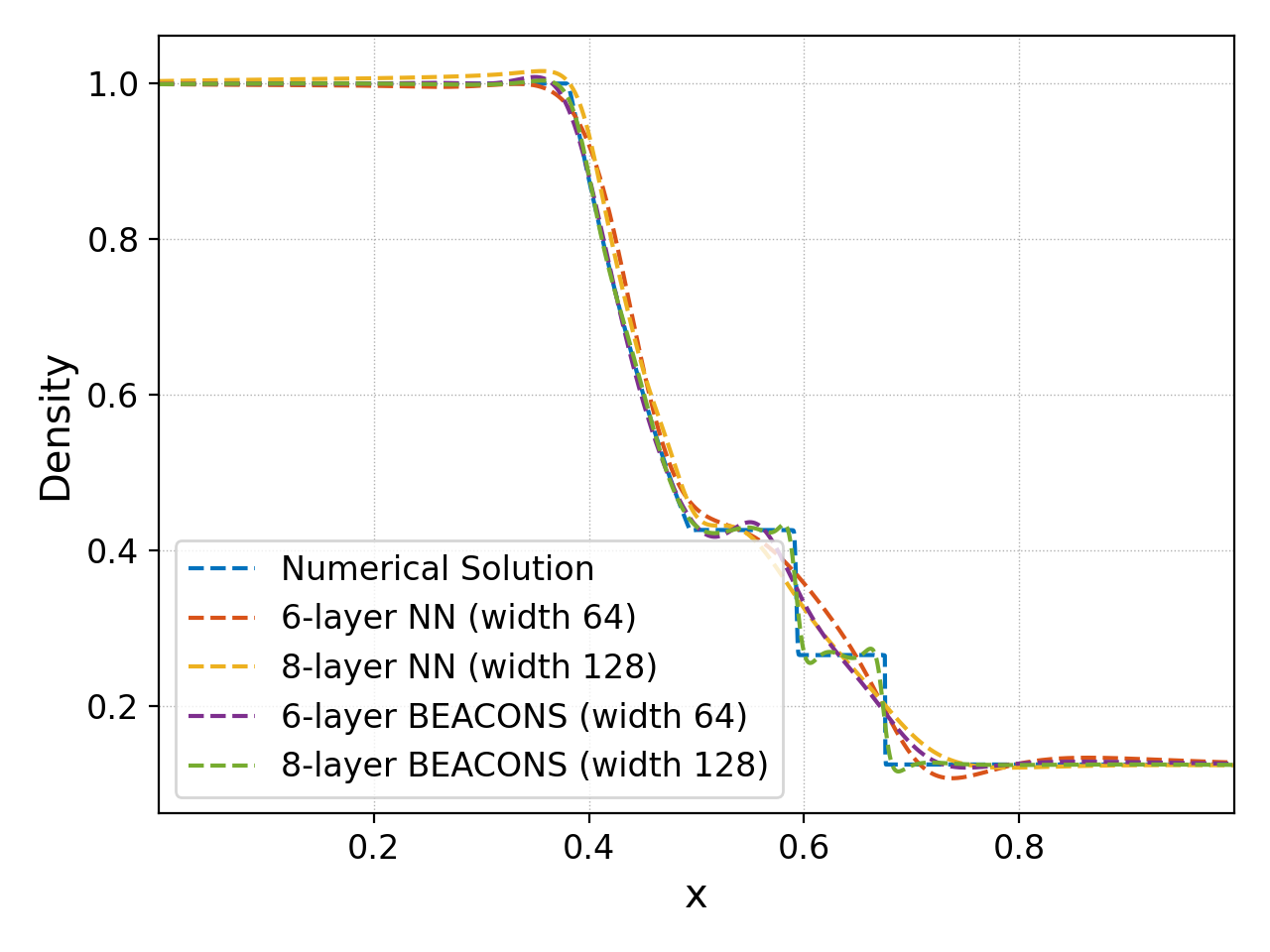}
\includegraphics[width=0.495\textwidth]{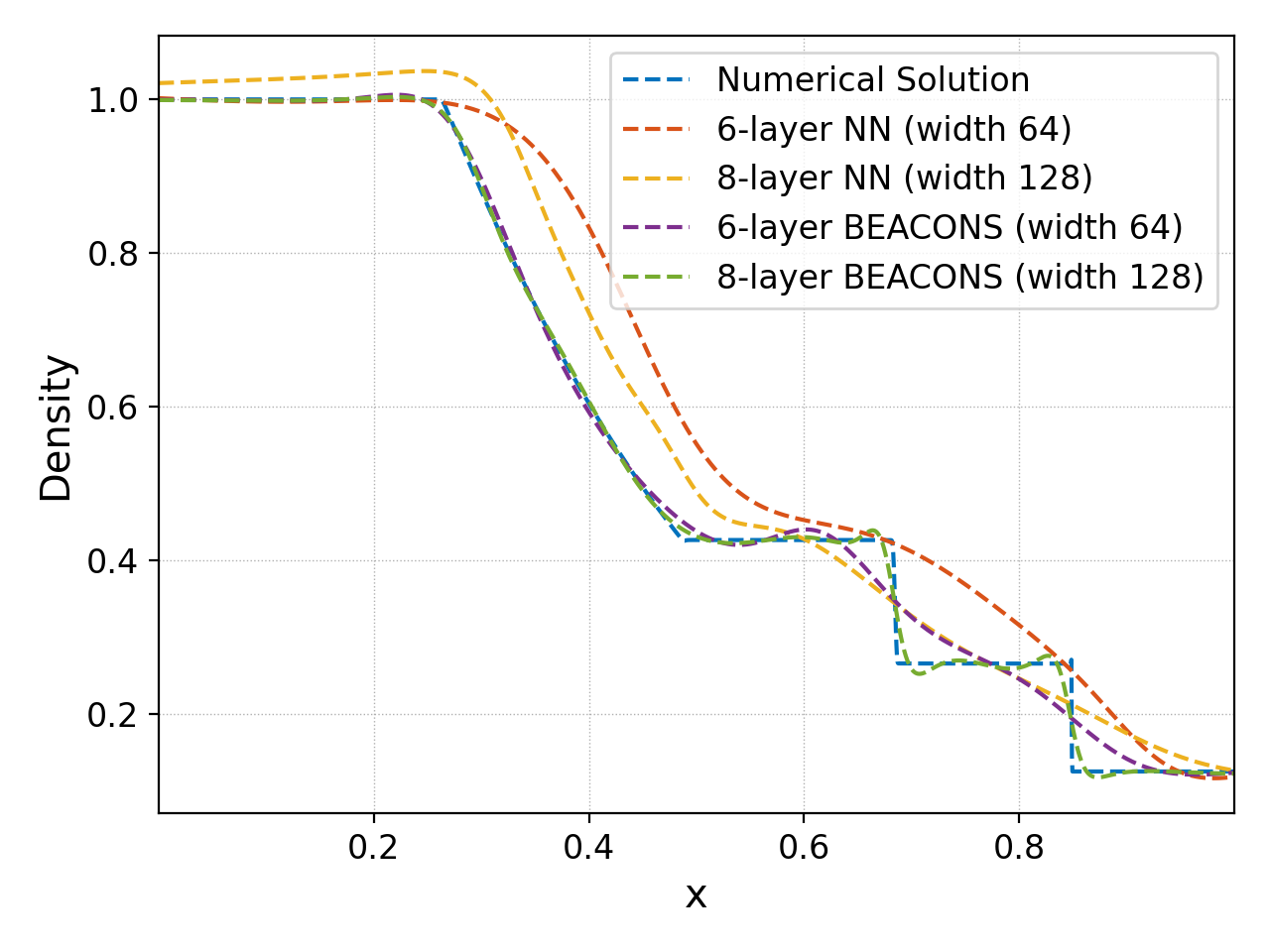}
\caption{Results for the 1D compressible Euler Sod-type shock tube problem at times ${t = 0.1}$ (left) and ${t = 0.2}$ (right), obtained using a high-resolution numerical solver (blue), a 6-layer neural network (red), an 8-layer neural network (yellow), a 6-layer BEACONS architecture (purple), and an 8-layer BEACONS architecture (green). Both the 6-layer and 8-layer neural networks completely diffuse the qualitative structure of the contact discontinuity and the right-moving shock, and significantly underestimate the speed of the left-moving rarefaction. Both BEACONS solutions successfully predict the speed of the left-moving rarefaction; in the 6-layer BEACONS solution again the qualitative structure of the contact discontinuity and the right-moving shock is diffused, but the 8-layer BEACONS solution tracks to numerical solution more-or-less perfectly.}
\label{fig:Figure5}
\end{figure}

Our third and final numerical test case will be to solve the system of \textit{compressible Euler equations}:

\begin{equation}
\frac{\partial}{\partial t} \begin{bmatrix}
\rho\\
\rho u\\
\rho E
\end{bmatrix} + \frac{\partial}{\partial x} \begin{bmatrix}
\rho u\\
\rho u^2 + P\\
u \left( \rho E + P \right)
\end{bmatrix} = \mathbf{0},
\end{equation}
i.e. a system of conservation laws for mass, momentum, and total energy ${\mathbf{U} = \left[ \rho, \rho u, \rho E \right]^{\intercal}}$, with non-linear flux ${\mathbf{F} \left( \mathbf{U} \right) = \left[ \rho u, \rho u^2 + P, u \left( \rho E + P \right) \right]^{\intercal}}$, where we assume an \textit{ideal gas} equation of state, relating the total energy $E$ and pressure $P$ by:

\begin{equation}
E = \left( \frac{P}{\gamma - 1} \right) + \frac{1}{2} \rho u^2, \qquad \Leftrightarrow \qquad P = \left( E - \frac{1}{2} \rho u^2 \right) \left( \gamma - 1 \right),
\end{equation}
where ${\gamma \in \mathbb{R}}$ is an arbitrary (constant) \textit{adiabatic index}. In contrast to the previous cases, we do not attempt any formal verification of the underlying numerical solvers for these equations; instead, we prove the bounds on the BEACONS architectures \textit{conditionally} (i.e. subject to the \textit{hypothesis} that the underlying numerical solvers are correct). The first test problem will be a one-dimensional Riemann problem known as the \textit{Sod shock tube}, defined over the spatial domain ${\left[ 0.0, 1.0 \right]}$, with adiabatic index ${\gamma = 1.4}$, and initial data given by:

\begin{equation}
\mathbf{U}_0 \left( x \right) = \begin{cases}
\left[ 1.0, 0.0, 2.5 \right]^{\intercal}, \qquad & \text{ for } x \leq 0.5,\\
\left[ 0.125, 0.0, 0.25 \right]^{\intercal}, \qquad & \text{ for } x > 0.5.
\end{cases}
\end{equation}
As previously, we shall solve this problem numerically using a high-resolution (but now unverified) \textit{Roe-type} approximate Riemann solver with a CFL coefficient ${C_{\text{CFL}} = 0.95}$, and with a spatial resolution of 2048 cells. We evolve the problem numerically up until a final time of ${t = 0.2}$, producing 200 frames of output in the process. Similar to before, we train four different neural network architectures on the first 66 frames of this output (up until time ${t = 0.067}$), and then ask each of them to predict the remainder of the simulation based on the learned solution function ${\mathbf{U} \left( t, x \right)}$. The chosen neural network architectures, as before, are: a 6-layer BEACONS architecture with 64 neurons per layer, an 8-layer BEACONS architecture with 128 neurons per layer, and, for means of comparison, a 6-layer fully-connected (non-BEACONS) neural network with 64 neurons per layer, and an 8-layer fully-connected (non-BEACONS) neural network with 128 neurons per layer. For this problem, our automated theorem-proving framework proves a worst-case ${L^{\infty}}$ error of 0.903602 (for density ${\rho}$, across both the smooth and non-smooth parts of the solution) for the 6-layer BEACONS architecture, and 0.707106 (for density ${\rho}$, across both the smooth and non-smooth parts of the solution) for the 8-layer BEACONS architecture\footnote{Note that these bounds are identical to the 1D linear advection case described previously, which stems from the fact that the governing equation for density ${\rho}$ is equivalent to the linear advection equation when the fluid velocity $u$ is held constant.}. In each case, the theorem-prover requires a total of 98,532 steps to establish the bound. The (inferred) solutions for ${\rho}$ (density) from the formally-verified numerical solver, the 6-layer neural network, the 8-layer neural network, the 6-layer BEACONS architecture, and the 8-layer BEACONS architecture, at times ${t = 0.1}$ and ${t = 0.2}$, are shown in Figure \ref{fig:Figure5}.

We see that the 6-layer neural network significantly underestimates the speed of the left-moving rarefaction, and completely diffuses out the qualitative structure of the contact discontinuity and the right-moving shock, which are not captured at all. The contact and shock waves also fail to be captured by the 8-layer neural network, although the mis-prediction of the left-moving rarefaction speed is now marginally less severe. Both neural networks visibly fail to conserve the mass density of the fluid. Both of the BEACONS solutions correctly estimate the left-moving rarefaction speed more-or-less perfectly, although the 6-layer BEACONS architecture still diffuses out the qualitative structure of the rest of the solution, and in particular fails to capture the contact discontinuity or the right-moving shock. However, the 8-layer BEACONS solution remains very close to the numerical solution, with all three waves (and their corresponding wave-speeds) being captured correctly, with only slight density overshoots and undershoots in the region surrounding the contact discontinuity and the right-moving shock. Both BEACONS solutions appear to conserve the mass density of the fluid at least approximately, with no significant diffusion of the qualitative structure of the solution in the 8-layer BEACONS case. Table \ref{tab:Table7} shows the normalized ${L^2}$ and ${L^{\infty}}$ errors (as compared against the high-resolution numerical solution) across the four different neural network architectures, both for the final frame only, and across all frames (using a per-frame average for the ${L^2}$ error, and normalizing across all frames for the ${L^{\infty}}$ error). We see that the final-frame ${L^2}$ and ${L^{\infty}}$ errors, along with the the all-frame ${L^2}$ errors, are significantly lower for the two BEACONS architectures than for the non-BEACONS architectures, although the all-frame ${L^{\infty}}$ errors remain broadly comparable between the BEACONS and non-BEACONS cases. In the non-BEACONS case, we see that increasing the number of layers has the effect of marginally decreasing the final-frame ${L^2}$ and ${L^{\infty}}$ errors, along with the all-frame ${L^2}$ errors, but has negligible effect on the all-frame ${L^{\infty}}$ errors. In the BEACONS case, we see that increasing the number of layers has the effect of significantly decreasing both the final-frame and all-frame ${L^2}$ errors, but has a negligible (or even marginally deleterious) effect on each of the final-frame or all-frame ${L^{\infty}}$ errors. Table \ref{tab:Table8} shows the normalized conservation errors (as obtained by integrating over the conserved quantity, and comparing against the integral of the high-resolution numerical solution) across the four different neural network architectures, again both for the final frame only, and integrated across all frames. We see overall that both the final-frame and all-frame conservation errors are significantly lower for the two BEACONS architectures than for the non-BEACONS architectures. In both the BEACONS and non-BEACONS cases, we see that increasing the number of layers has the effect of decreasing both the final-frame and all-frame conservation errors, although these decreases are altogether more substantial in the BEACONS case. Note that the largest ${L^{\infty}}$ errors observed for the two BEACONS architectures (i.e. 0.422681 and 0.523623, respectively) remain comfortably below the worst-case bounds (i.e. 0.903602 and 0.707106, respectively).

\begin{table}[ht]
\centering
\begin{tabular}{|c||c|c|c|c|}
\hline
Architecture & ${L^{\infty}}$ Error (Final) & ${L^2}$ Error (Final) & ${L^{\infty}}$ Error (All) & ${L^2}$ Error (All)\\
\hline\hline
6-layer NN & 0.230453 & 4.522916 & 0.444256 & 2.075029\\
\hline
8-layer NN & 0.159231 & 3.002094 & 0.452337 & 1.546402\\
\hline
6-layer BEACONS & 0.076894 & 1.004591 & 0.422681 & 0.758668\\
\hline
8-layer BEACONS & 0.080181 & 0.411615 & 0.523623 & 0.328464\\
\hline
\end{tabular}
\caption{${L^2}$ and ${L^{\infty}}$ error analysis for the 1D compressible Euler Sod-type shock tube problem, comparing the 6-layer neural network, 8-layer neural network, 6-layer BEACONS, and 8-layer BEACONS solutions against the high-resolution numerical solution, both for the final predicted frame, and across all frames.}
\label{tab:Table7}
\end{table}

\begin{table}[ht]
\centering
\begin{tabular}{|c||c|c|}
\hline
Architecture & Conservation Error (Final) & Conservation Error (Total)\\
\hline\hline
6-layer NN & 133.878340 & 6219.332405\\
\hline
8-layer NN & 83.996340 & 5127.036345\\
\hline
6-layer BEACONS & 3.343495 & 281.148077\\
\hline
8-layer BEACONS & -0.991722 & -211.287987\\
\hline
\end{tabular}
\caption{Conservation analysis for the 1D compressible Euler Sod-type shock tube problem, comparing the 6-layer neural network, 8-layer neural network, 6-layer BEACONS, and 8-layer BEACONS solutions against the high-resolution numerical solution, both for the final predicted frame, and across all frames.}
\label{tab:Table8}
\end{table}

\begin{figure}[ht]
\centering
\includegraphics[trim={1.5cm, 0cm, 1.5cm, 0cm}, clip, width=0.33\textwidth]{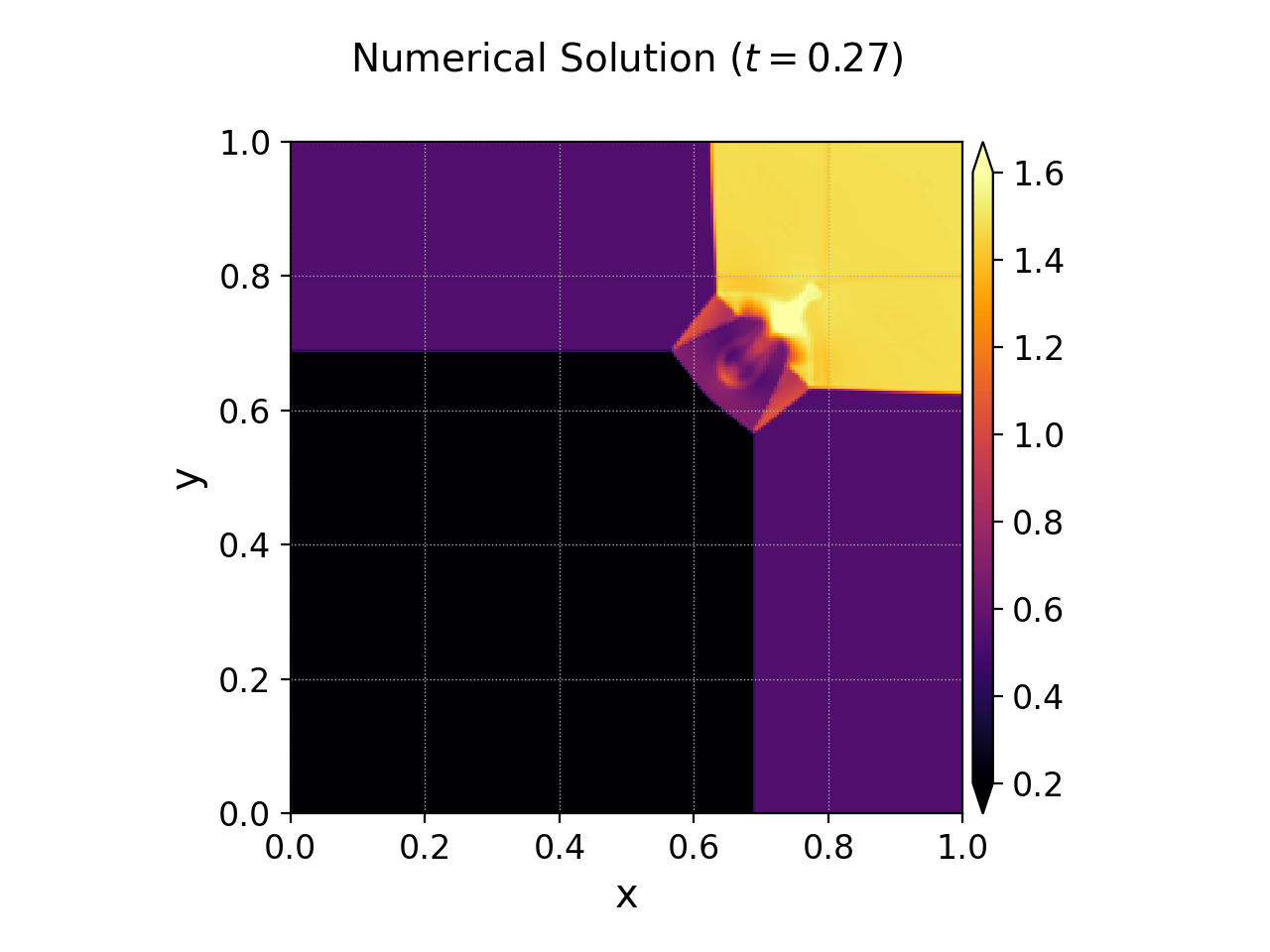}
\includegraphics[trim={1.5cm, 0cm, 1.5cm, 0cm}, clip, width=0.33\textwidth]{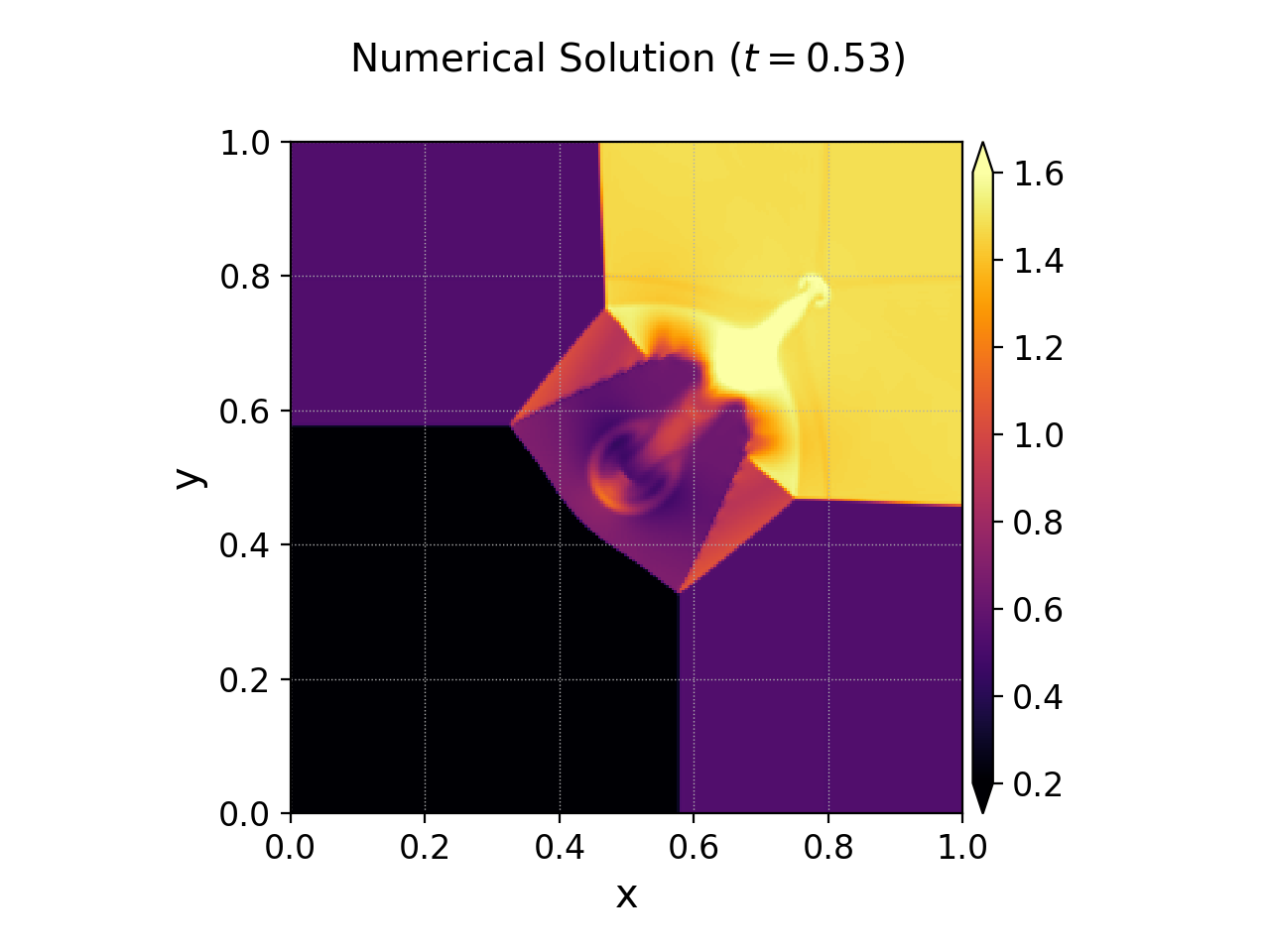}
\includegraphics[trim={1.5cm, 0cm, 1.5cm, 0cm}, clip, width=0.33\textwidth]{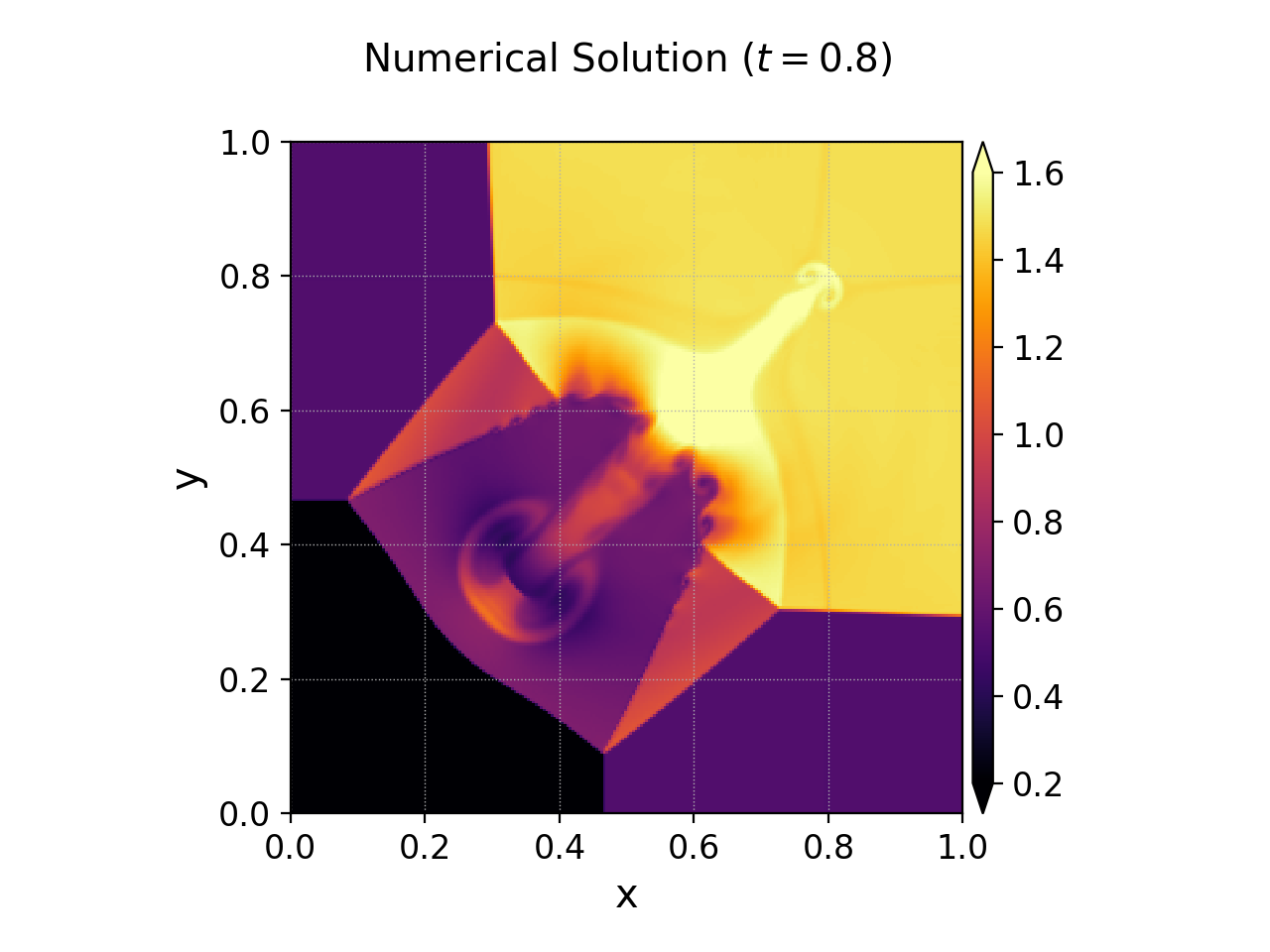}
\includegraphics[trim={1.5cm, 0cm, 1.5cm, 0cm}, clip, width=0.33\textwidth]{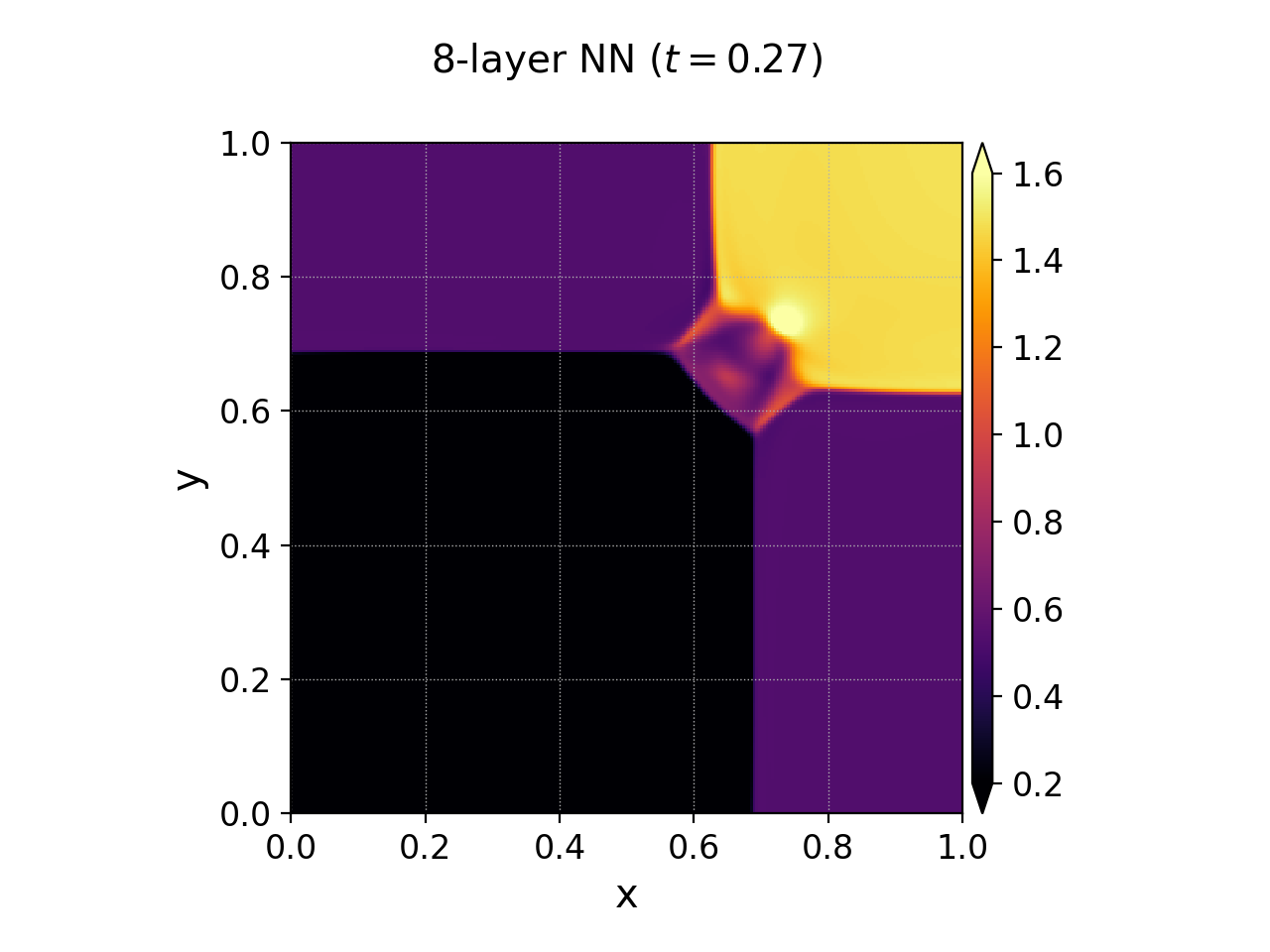}
\includegraphics[trim={1.5cm, 0cm, 1.5cm, 0cm}, clip, width=0.33\textwidth]{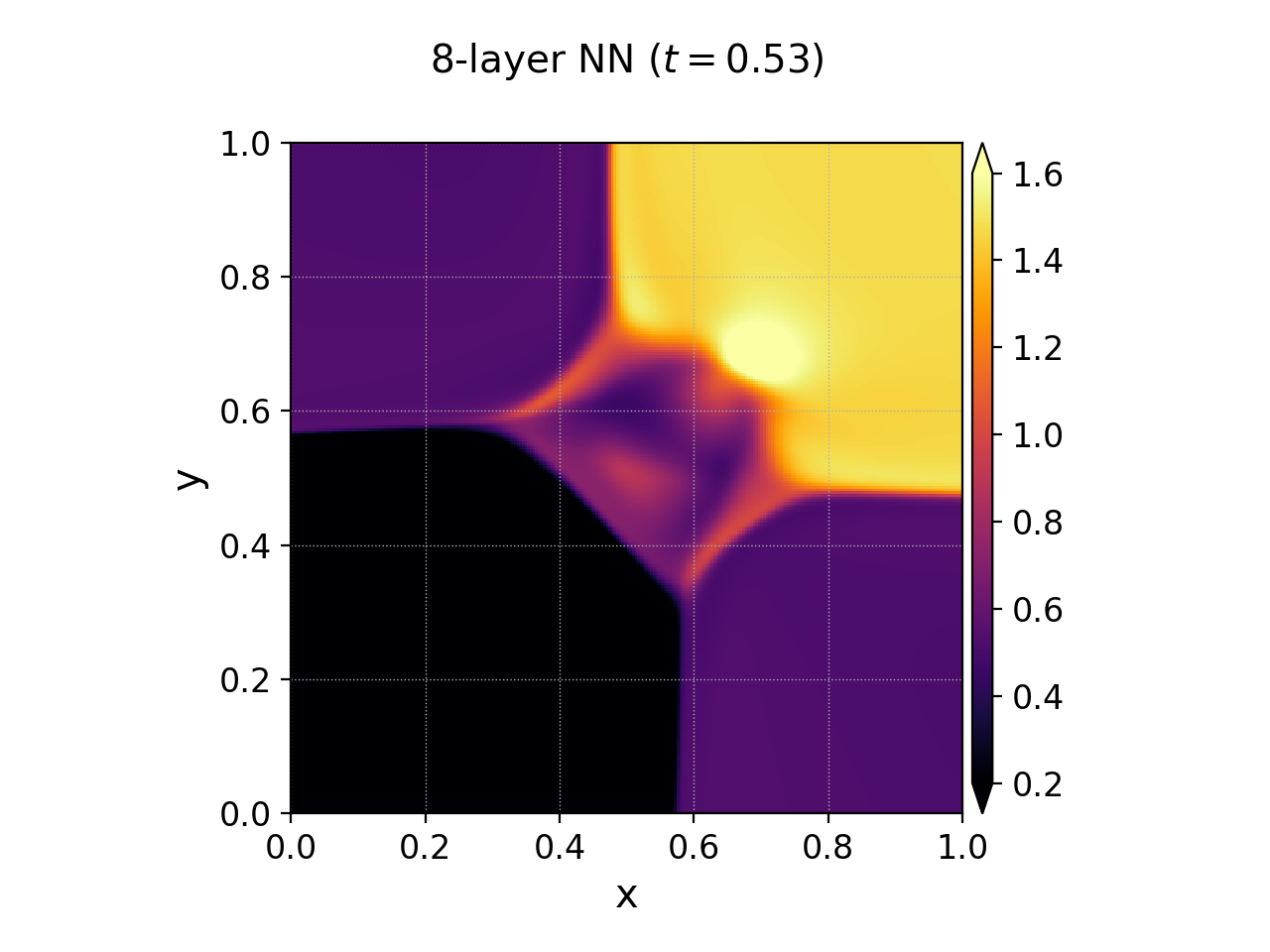}
\includegraphics[trim={1.5cm, 0cm, 1.5cm, 0cm}, clip, width=0.33\textwidth]{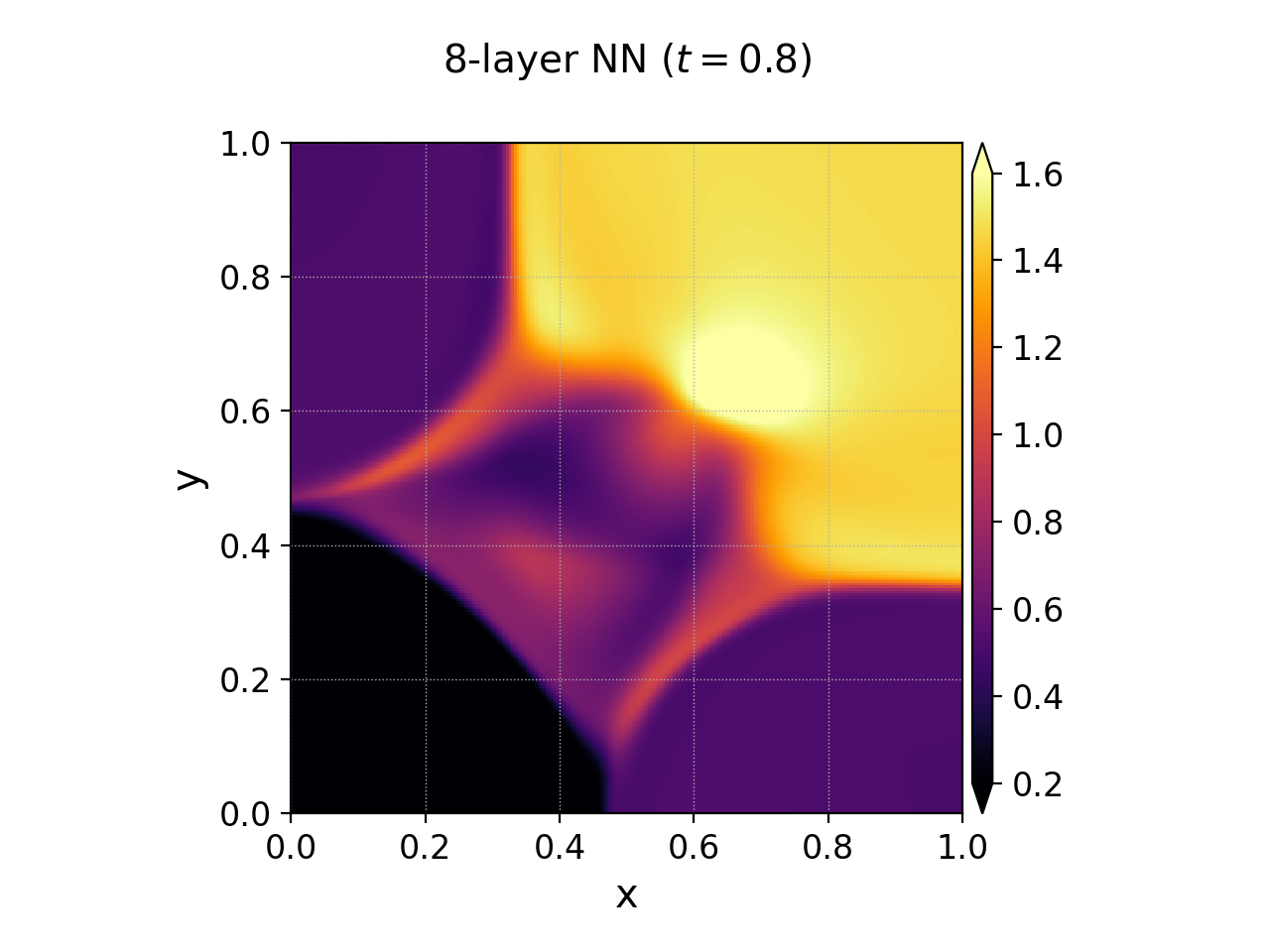}
\includegraphics[trim={1.5cm, 0cm, 1.5cm, 0cm}, clip, width=0.33\textwidth]{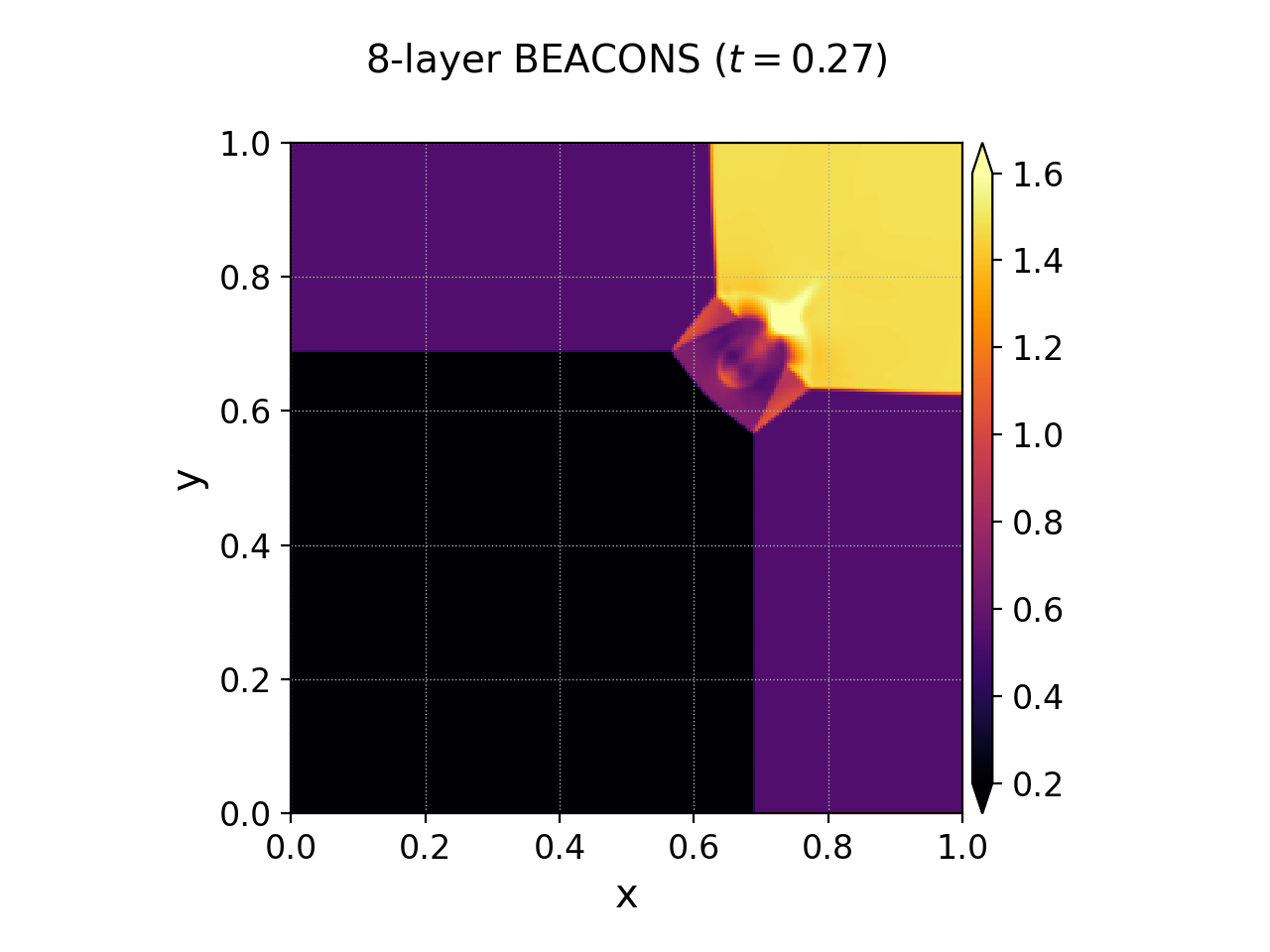}
\includegraphics[trim={1.5cm, 0cm, 1.5cm, 0cm}, clip, width=0.33\textwidth]{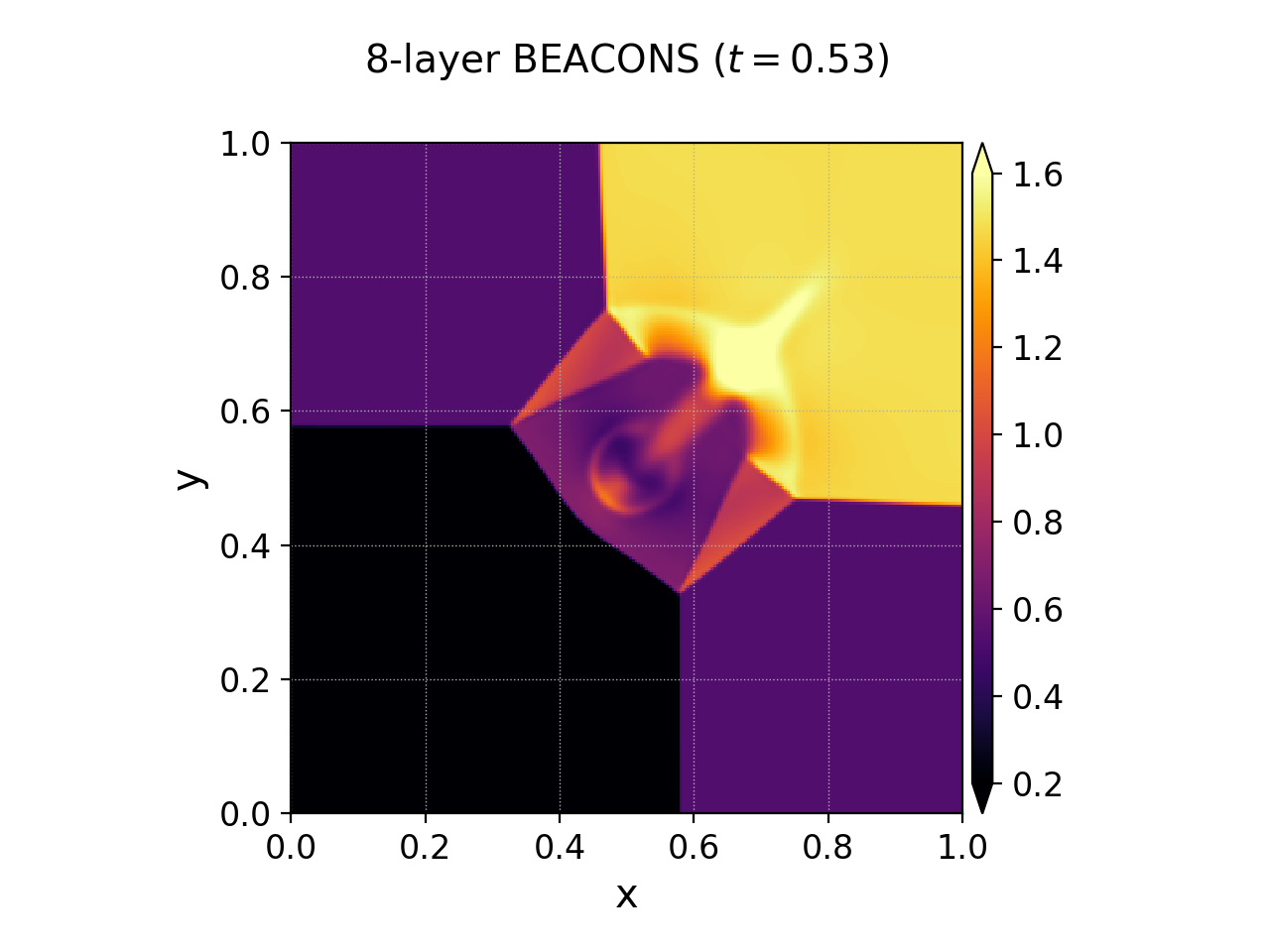}
\includegraphics[trim={1.5cm, 0cm, 1.5cm, 0cm}, clip, width=0.33\textwidth]{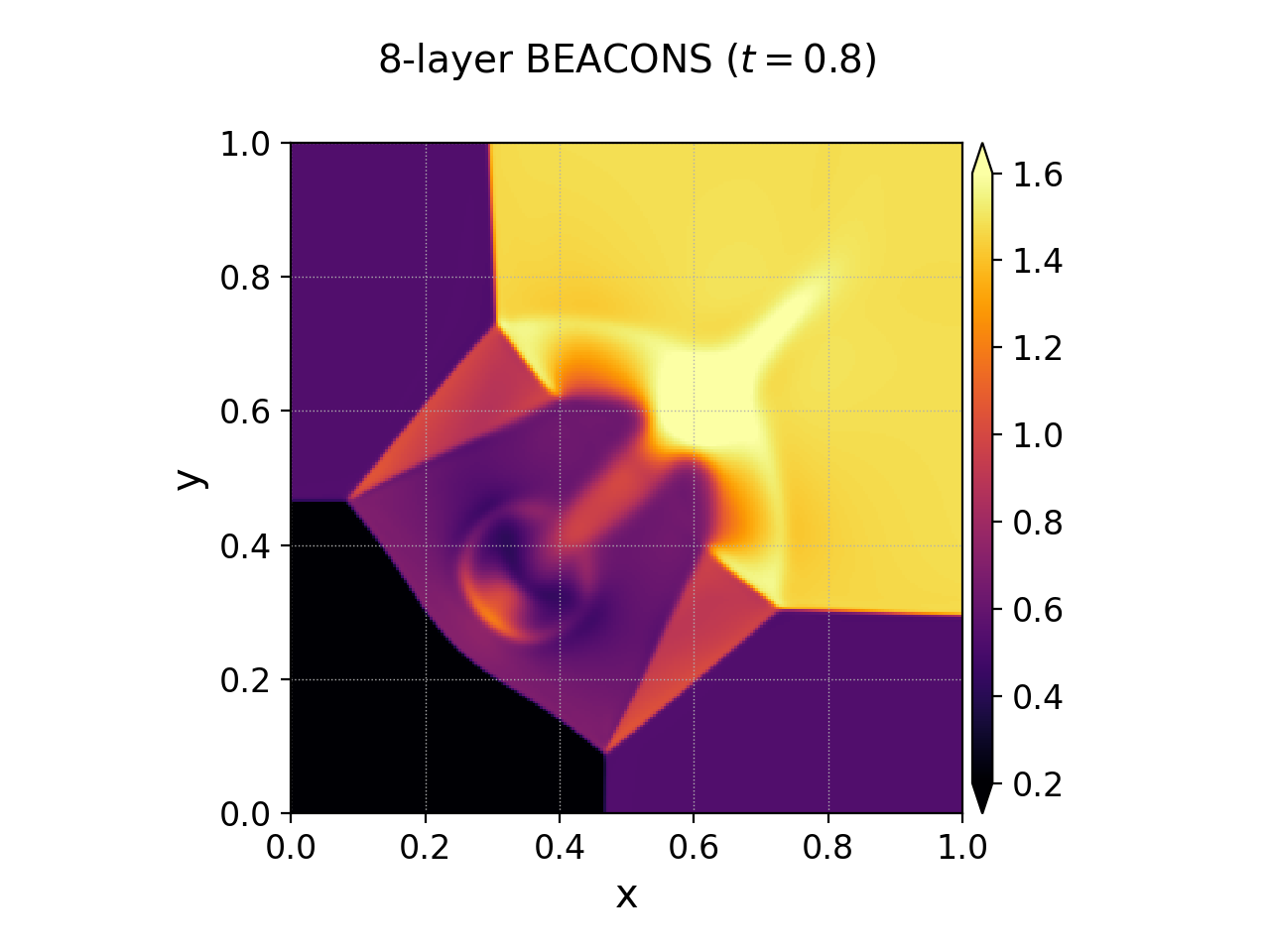}
\caption{Results for the 2D compressible Euler quadrants problem at times ${t = 0.27}$ (left), ${t = 0.53}$ (middle), and ${t = 0.8}$ (right), obtained using a high-resolution numerical solver (top), an 8-layer neural network (middle), and an 8-layer BEACONS architecture (bottom). The 8-layer neural network completely diffuses the qualitative structure of the solution and mis-predicts many of the relevant wave-speeds; the protruding ``ray'' and surrounding structure on the top right of the domain is completely absent, the propagating wave front on the bottom left of the domain is incorrectly predicted as being concave rather than convex, etc. The BEACONS architecture predicts both the correct overall shape and the correct propagation speeds of the constituent waves of the solution, matching the numerical solution more-or-less perfectly (albeit with an absence of certain fine structures and instabilities due to increased numerical diffusion).}
\label{fig:Figure6}
\end{figure}

The second test problem will consider an extension of the compressible Euler equations to two dimensions:

\begin{equation}
\frac{\partial}{\partial t} \begin{bmatrix}
\rho\\
\rho u\\
\rho v\\
\rho E
\end{bmatrix} + \frac{\partial}{\partial x} \begin{bmatrix}
\rho u\\
\rho u^2 + P\\
\rho u v\\
u \left( \rho E + P \right)
\end{bmatrix} + \frac{\partial}{\partial y} \begin{bmatrix}
\rho v\\
\rho v u\\
\rho v^2 + P\\
v \left( \rho E + P \right)
\end{bmatrix} = \mathbf{0},
\end{equation}
i.e. a system of conservation laws for mass, $x$-momentum, $y$-momentum, and total energy ${\mathbf{U} = \left[ \rho, \rho u, \rho v, \rho E \right]^{\intercal}}$, with non-linear directional fluxes ${\mathbf{F}_x \left( \mathbf{U} \right) = \left[ \rho u, \rho u^2 + P, \rho u v, u \left( \rho E + P \right) \right]^{\intercal}}$ and ${\mathbf{F}_y \left( \mathbf{U} \right) = \left[ \rho v, \rho v u, \rho v^2 + P, v \left( \rho E + P \right) \right]^{\intercal}}$, with the natural extension to the ideal gas equation of state (for adiabatic index ${\gamma \in \mathbb{R}}$):

\begin{equation}
E = \left( \frac{P}{\gamma - 1} \right) + \frac{1}{2} \rho \left( u^2 + v^2 \right), \qquad \Leftrightarrow \qquad P + \left( E - \frac{1}{2} \rho \left( u^2 + v^2 \right) \right) \left( \gamma - 1 \right).
\end{equation}
As in the 1D case, we do not attempt any formal verification of the underlying numerical solvers for these equations. We initialize a two-dimensional \textit{quadrants} problem\cite{schulz-rinne_numerical_1993}\cite{leveque_wave_1997} on a square domain ${\left[ 0.0, 1.0 \right] \times \left[ 0.0, 1.0 \right]}$, with four initial discontinuities:

\begin{equation}
\mathbf{U}_0 \left( x, y \right) = \begin{cases}
\left[ 1.5, 0.0, 0.0, 3.75 \right]^{\intercal}, \qquad & \text{ for } x \geq 0.8 \text{ and } y \geq 0.8\\
\left[ 0.5323, 0.641954, 0.0, 1.1371 \right]^{\intercal}, \qquad & \text{ for } x < 0.8 \text{ and } y \geq 0.8,\\
\left[ 0.5323, 0.0, 0.641954, 1.1371 \right]^{\intercal}, \qquad & \text{ for } x \geq 0.8 \text{ and } y < 0.8\\
\left[ 0.138, 0.166428, 0.166428, 0.273212 \right]^{\intercal}, \qquad & \text{ for } x < 0.8 \text{ and } y < 0.8,
\end{cases}
\end{equation}
and with adiabatic index ${\gamma = 1.4}$, yielding a highly complex and intricate solution structure consisting of multiple interacting waves. Once again, we solve this problem using a high-resolution (but unverified) \textit{Roe-type} approximate Riemann solver, with a CFL coefficient ${C_{\text{CFL}} = 0.95}$, and a spatial resolution of ${256 \times 256}$ cells. We evolve the problem numerically up until a final time of ${t = 0.8}$, producing 100 frames of output, and now train only two different neural network architectures on the first 33 frames (up until time ${t = 0.27}$), and again ask both of them to predict the remainder of the simulation based on the learned solution function ${\mathbf{U} \left( t, x, y \right)}$. The chosen neural network architectures are an 8-layer BEACONS architecture with 128 neurons per layer, and, for means of comparison, an 8-layer fully-connected (non-BEACONS) neural network with 128 neurons per layer. For this problem, our automated theorem-proving framework proves worst-case ${L^{\infty}}$ errors of 1.216729 (for density ${\rho}$ in the smooth parts of the solution) and 1.483672 (for density ${\rho}$ in the non-smooth parts of the solution) for a 6-layer BEACONS architecture, and 1.0 (for density ${\rho}$ in the smooth parts of the solution) and 1.259921 (for density ${\rho}$ in the non-smooth parts of the solution) for the 8-layer BEACONS architecture used here\footnote{Again, these bounds are identical to the 2D linear advection case, due to the nature of the governing equation for density ${\rho}$ when the fluid velocity $u$ remains constant.}. In each case, the theorem-prover requires a total of 142,104 proof steps to establish the bound. The (inferred) solutions from the high-resolution numerical solver, the 8-layer neural network, and the 8-layer BEACONS architecture, at times ${t = 0.27}$, ${t = 0.53}$, and ${t = 0.8}$, are shown in Figure \ref{fig:Figure6}. We see that the 8-layer neural network completely fails to capture the qualitative structure of the solution, which is diffused beyond reasonable recognition, and also fails to predict the correct speeds and interactions for many of the relevant waves. In particular, the protruding ``ray'' and surrounding shock front structure on the top right of the domain, along with most of the internal shock and rarefaction structure in the middle and bottom left of the domain, are completely absent from the 8-layer neural network solution, which also mis-predicts the propagating wave front at the bottom left of the domain as being concave rather than convex in shape. On the other hand, the 8-layer BEACONS solution successfully predicts both the correct qualitative shape and the correct quantitative propagation speeds for all of the relevant waves and their non-linear interactions, exhibiting only an absence of certain fine structures (such as the ``finger-like'' instabilities present in the numerical solution at late times) due to increased numerical diffusion around the wave fronts. Table \ref{tab:Table9} shows the normalized ${L^2}$ and ${L^{\infty}}$ errors (as compared against the high-resolution numerical solution) across the two different neural network architectures, both for the final frame only, and across all frames (using a per-frame average for the ${L^2}$, and normalizing across all frames for the ${L^{\infty}}$ error). We see overall that both the final-frame and all-frame ${L^2}$ and ${L^{\infty}}$ errors are significantly lower for the BEACONS architecture than for the non-BEACONS architecture. Note again that the largest ${L^{\infty}}$ error observed for the BEACONS architecture (i.e. 0.310002) remains comfortably below the worst-case bound (i.e. 1.259921).

\begin{table}[ht]
\centering
\begin{tabular}{|c||c|c|c|c|}
\hline
Architecture & ${L^{\infty}}$ Error (Final) & ${L^2}$ Error (Final) & ${L^{\infty}}$ Error (All) & ${L^2}$ Error (All)\\
\hline\hline
8-layer NN & 0.546258 & 25.843028 & 0.536434 & 12.315916\\
\hline
8-layer BEACONS & 0.198093 & 2.958570 & 0.310002 & 1.539846\\
\hline
\end{tabular}
\caption{${L^2}$ and ${L^{\infty}}$ error analysis for the 2D compressible Euler quadrants problem, comparing the 8-layer neural network and 8-layer BEACONS solutions against the high-resolution numerical solution, both for the final predicted frame, and across all predicted frames.}
\label{tab:Table9}
\end{table}

\section{Concluding Remarks}
\label{sec:Section10}

One of the overarching philosophical themes of this paper has been that neural networks represent a large and highly general class of fundamental numerical methods, and that they can consequently be rigorously analyzed in much the same way as classical numerical algorithms (such as finite element, finite difference, finite volume, etc. methods), in many cases by using appropriate generalizations of the same mathematical techniques. One reason why such a research program seems not to have been pursued systematically in the past may be a consequence of Sutton's so-called ``bitter lesson''\cite{sutton2019bitter}: incremental improvements in neural network-based methods due to developments in underlying mathematical theory or breakthroughs in architectural design are often dwarfed by much larger improvements resulting simply from increases in computational scale, for instance by training on more data, incorporating larger numbers of parameters, or using deeper architectures. This stands in stark contrast to the case of computational physics, where increases in computational scale allow one to run larger and higher-resolution simulations more quickly, but do not lead to intrinsic improvements in the quality of the numerical methods themselves. Thus, developments in classical numerical methods have historically been bottlenecked by the rates of progress in their mathematical foundations and algorithmic design, yet neural network-based methods have hitherto been remarkably successful in evading their analogous limitations through the application of brute computational power.

However, if neural network-based methods are to be truly successful in accelerating, supplementing, extending, or perhaps even wholesale replacing conventional numerical simulation codes for the solution of PDEs, then this requirement for a rigorous underlying mathematical theory guaranteeing the correctness (or at least bounding the errors) of their results can no longer be ignored. The success of computational physics derives, to a very great extent, from the reliability of its numerical simulation codes, and the guarantees of convergence, stability, consistency, and correctness that they are able to afford. Until or unless neural network-based methods are able to reach an equivalent level of reliability, the need for classical numerical codes will persist. A neurosymbolic approach, with a neural network-based solver augmented by an automated theorem-prover that is able to verify its correctness properties formally for each individual problem, seems potentially optimal for bridging this gap. We believe that the BEACONS framework, with its machine-checkable proof certificates and its ability to scale to arbitrarily deep architectures via algebraic composability, represents an important first step along this exciting path, as well as the beginning of a large-scale research program in its own right.

In the short term, we anticipate extending the BEACONS paradigm to systems of ordinary differential equations (ODEs), as well as to systems of elliptic and parabolic PDEs, for instance by exploiting elliptic regularity theorems\cite{fernandez-real_regularity_2022} in place of the method of characteristics in order to make a priori predictions regarding the smoothness of solutions. Although in this paper we have focused exclusively on the \textit{forward problem} of inferring PDE solutions from initial data, we also expect many future applications of BEACONS-based approaches to PDE-constrained (non-convex) optimization\cite{jain_non-convex_2017} and dimensionality reduction\cite{gorard2025improveddimensionalityreductioninverse}, thereby enabling the construction of formally-verified surrogates for \textit{inverse problems} too. Over a longer time horizon, just as multiple numerical solvers encompassing distinct laws of physics may be coupled together to form more complex \textit{multi-physics} solvers, one could equally imagine coupling together multiple BEACONS architectures modeling distinct laws of physics to form more complex BEACONS-based \textit{foundation models}. In particular, one could envision \textit{pretraining} a large collection of specialized BEACONS architectures for simulating individual laws of physics, and then coupling those architectures together to form a ``supernetwork'' by means of a \textit{mixture-of-experts} (MoE) approach\cite{mu2026comprehensivesurveymixtureofexpertsalgorithms}. The \textit{finetuning} process for such a BEACONS-based foundation model would then consist of training this ``supernetwork'' to assemble a solution to a complex, multi-physics problem as a non-linear combination of these individual BEACONS solutions, each solving for a separate component of the overall multi-physics system, and with the ``supernetwork'' then learning how to couple these components together correctly in a fully formally-verified way. Initial experimentation with such an approach has already yielded highly promising results.

One of the core principles of the BEACONS philosophy is that the development of classical numerical solvers and neural network-based solvers must necessarily go hand in hand: the formally-verified numerical solver is what enables the \textit{bootstrapping} of the neural network-based solver via the generation of arbitrary amounts of (certifiably-correct) training data, while the boostrapped neural network-based solver effectively extends the classical numerical solver by extrapolating its solutions into regimes which have not yet been explicitly simulated. Indeed, perhaps the most exciting possibility is the ability of the neural network-based solver to extrapolate into regimes which \textit{cannot}, even in principle, be explicitly simulated. For example, there may be regions of parameter space which cannot be modeled using any explicit time integration method, because the characteristic time-scales become so short that it would force the stable time-step of any explicit numerical method to ``crash'' to zero, yet the solution still remains well-defined in a mathematical sense. A BEACONS architecture, having been trained on explicit numerical solutions obtained from the unproblematic regions of the parameter space (in which the stable time-step remains finite), would thereby give one a systematic means of answering the following counterfactual question: what \textit{would} the simulation code have predicted the solution to be, \textit{if} it had been able to run in this regime without crashing? Such an ability to answer counterfactual questions regarding otherwise unsimulable parameter regimes, in a mathematically principled way, would be of obvious and enduring importance for computational physics.

Since neural networks are, in essence, arbitrarily efficient compressors of algorithmic information, the aforementioned process of bootstrapping a compressed neural network-based solver from an existing numerical simulation code may be regarded as a highly aggressive and non-deterministic form of compiler optimization. Modest amounts of non-determinism and small floating-point inaccuracies are already tolerated as part of standard compiler optimization routines in scientific computing, such as GCC's \texttt{-ffast-math} flag; conventional neural network architectures merely represent a particularly extreme limit case of the same idea. The BEACONS paradigm may ultimately be thought of as an effort to promote the neural network ``compiler'' from a lossy, heuristic, and unprincipled one into a lossless, verified, and predictable one.

\section*{Acknowledgements}

J.G. was partially funded by the Princeton University Research Computing group. J.G., A.H., and J.J. were partially funded by the U.S. Department of Energy under Contract No. DE-AC02-09CH1146 via an LDRD grant. The development of \textsc{Gkeyll} was partially funded, besides the grants mentioned above, by the NSF-CSSI program, Award Number 2209471. No generative artificial intelligence was used in the production of this manuscript, nor in any part of the research described therein.

\bibliographystyle{acm}
\bibliography{BEACONSPaper}

\end{document}